\documentclass[10pt]{article}
\usepackage{latexsym,amsmath,amssymb,graphics,amscd, empheq}
\usepackage{multirow}
\usepackage{soul}
\usepackage{booktabs}
\usepackage{tabularx}
\usepackage[hang,footnotesize]{caption}
\usepackage[pdftex]{graphicx}
\usepackage{graphicx}
\usepackage{overpic}
\usepackage{color}
\usepackage[small,nohug,heads=littlevee]{diagrams}
\diagramstyle[labelstyle=\scriptstyle]
\usepackage{cancel}
\usepackage[pdftex,breaklinks,colorlinks,
  citecolor=blue,
  urlcolor=blue]{hyperref}
\usepackage{scalerel,stackengine}
\stackMath
\newcommand{\vertiii}[1]{{\left\vert\kern-0.25ex\left\vert\kern-0.25ex\left\vert #1 
    \right\vert\kern-0.25ex\right\vert\kern-0.25ex\right\vert}}
\newcommand\reallywidehat[1]{%
\savestack{\tmpbox}{\stretchto{%
  \scaleto{%
    \scalerel*[\widthof{\ensuremath{#1}}]{\kern-.6pt\bigwedge\kern-.6pt}%
    {\rule[-\textheight/2]{1ex}{\textheight}}%WIDTH-LIMITED BIG WEDGE
  }{\textheight}% 
}{0.5ex}}%
\stackon[1pt]{#1}{\tmpbox}%
}
\usepackage{nomencl}
\makenomenclature

\addtocounter{footnote}{1}
\makeatletter
\@addtoreset{table}{bsection}
\def\thetable{\thesection.\@arabic\c@table}
\def\fps@table{h, t}
\@addtoreset{equation}{section}

\makeatother

\newtheorem{theorem}{Theorem}[section]
\newtheorem{definition}[theorem]{Definition}
\newtheorem{lemma}[theorem]{Lemma}
\newtheorem{remark}[theorem]{Remark}
\newtheorem{proposition}[theorem]{Proposition}

\newtheorem{corollary}[theorem]{Corollary}

\newcommand{\bfi}{\bfseries\itshape}
\newsavebox{\savepar}

\makeatletter

\begin{document}

\title{\textbf{Universal discrete-time reservoir computers with stochastic inputs and linear readouts using non-homogeneous state-affine systems}}
\author{Lyudmila Grigoryeva$^{1}$ and Juan-Pablo Ortega$^{2, 3}$}
\date{}
\maketitle

\begin{abstract}
A new class of non-homogeneous state-affine systems is introduced for use in reservoir computing. Sufficient conditions are identified that guarantee first, that the associated reservoir computers with linear readouts are causal, time-invariant, and satisfy the fading memory property and second, that a subset of this class is universal in the category of fading memory filters with stochastic almost surely uniformly bounded inputs. This means that any discrete-time filter that satisfies the fading memory property with random inputs of that type can be uniformly approximated by elements in the non-homogeneous state-affine family.
\end{abstract}

\bigskip


\makeatletter
\addtocounter{footnote}{1} \footnotetext{%
Department of Mathematics and Statistics. Universit\"at Konstanz. Box 146. D-78457 Konstanz. Germany. {\texttt{Lyudmila.Grigoryeva@uni-konstanz.de} }}
\addtocounter{footnote}{1} \footnotetext{%
Universit\"at Sankt Gallen. Faculty of Mathematics and Statistics. Bodanstrasse 6.
CH-9000 Sankt Gallen. Switzerland. {\texttt{Juan-Pablo.Ortega@unisg.ch}}}
\addtocounter{footnote}{1} \footnotetext{%
Centre National de la Recherche Scientifique (CNRS). France. }
\makeatother

\medskip

\nomenclature{$F: \mathbb{R} ^N\times \mathbb{R} ^n\longrightarrow  \mathbb{R} ^N$}{Reservoir map}
\nomenclature{$h: \mathbb{R}^N \rightarrow \mathbb{R}$}{Generic readout map}
\nomenclature{$H _U: (\mathbb{R}^n)^{\Bbb Z _-} \longrightarrow \mathbb{R} $}{Functional associated to the causal and time-invariant filter $U: (\mathbb{R}^n) ^{\Bbb Z} \longrightarrow \mathbb{R} ^{\Bbb Z} $}
\nomenclature{$K _M $}{Space of semi-infinite sequences that are uniformly bounded by $M$}
\nomenclature{$K^{L^{\infty}}_{M} $}{Space of semi-infinite processes that are almost surely uniformly bounded by $M$}
\nomenclature{$\mathbb{M}  _n $}{Space of square matrices of order $n \in \mathbb{N}$}
\nomenclature{$\mathbb{M}_{n ,  m }$}{Space of real $n\times m$ matrices with $m, n \in \mathbb{N} $}
\nomenclature{$\mathbb{M}_{m,n}[z] $}{$\mathbb{M}_{m,n} $-valued polynomials  on $z$ with coefficients in $\mathbb{M}_{m,n} $}
\nomenclature{$n$}{Dimension of the elements of the input signal}
\nomenclature{$N$}{Number of virtual neurons. Dimension of the reservoir state vectors}
\nomenclature{$U: (\mathbb{R}^n) ^{\Bbb Z} \longrightarrow \mathbb{R} ^{\Bbb Z} $}{Filter with inputs in ${\Bbb R}^n $ and outputs in $\mathbb{R}$}
\nomenclature{$U ^F: (\mathbb{R}^n)^{\Bbb Z} \longrightarrow(\mathbb{R}^N)^{\Bbb Z} $}{Filter determined by the reservoir map $F$}
\nomenclature{$U ^F _h: (\mathbb{R}^n)^{\Bbb Z} \longrightarrow\mathbb{R}^{\Bbb Z} $}{Reservoir filter}
\nomenclature{$U_H: (\mathbb{R}^n) ^{\Bbb Z} \longrightarrow \mathbb{R} ^{\Bbb Z} $}{Causal and time-invariant filter associated to the functional $H : (\mathbb{R}^n)^{\Bbb Z _-} \longrightarrow \mathbb{R}  $}
\nomenclature{$w : \mathbb{N} \longrightarrow (0,1] $}{Weighting sequence}
\nomenclature{${\bf x} $}{(Semi)-infinite sequence containing the reservoir states.  The elements of this  sequence are denoted by ${\bf x} _t \in {\Bbb R}^N$}
\nomenclature{${\bf y} $}{(Semi)-infinite output signal. The elements of this  sequence are denoted by $y _t \in \mathbb{R}$}
\nomenclature{${\bf z} $}{(Semi)-infinite input signal. The elements of this  sequence are denoted by ${\bf z} _t \in {\Bbb R}^n$}
\nomenclature{$\ell ^{\infty}_w({\Bbb R}^n)$}{Banach space of semi-infinite sequences with finite weighted norm}

\section{Introduction}

A {\bfi reservoir computer (RC)}~\cite{jaeger2001, Jaeger04, maass1, maass2, Crook2007, verstraeten, lukosevicius} or a {\bfi  RC system} is a specific type of recurrent neural network  determined by two maps, namely a {\bfi  reservoir} $F: \mathbb{R} ^N\times \mathbb{R} ^n\longrightarrow  \mathbb{R} ^N$, $n,N \in \mathbb{N} $,  and a {\bfi  readout} map $h: \mathbb{R}^N \rightarrow \mathbb{R}$ that under certain hypotheses transform (or filter) an infinite discrete-time input  ${\bf z}=(\ldots, {\bf z} _{-1}, {\bf z} _0, {\bf z} _1, \ldots) \in (\mathbb{R}^n) ^{\Bbb Z } $ into an output signal ${\bf y} \in \mathbb{R} ^{\Bbb Z } $ of the same type using  the state-space transformation given by:
\begin{empheq}[left={\empheqlbrace}]{align}
\mathbf{x} _t &=F(\mathbf{x}_{t-1}, {\bf z} _t),\label{reservoir equation}\\
y _t &= h (\mathbf{x} _t), \label{readout}
\end{empheq}
where $t \in \Bbb Z $ and the dimension $N \in \mathbb{N} $ of the state vectors $\mathbf{x} _t \in \mathbb{R} ^N $ will be referred to as the number of virtual {\bfi  neurons} of the system. The expressions \eqref{reservoir equation}-\eqref{readout} determine a nonlinear state-space system and many of its dynamical properties (stability, controlability) have been studied for decades  in the literature from that point of view. 

This notion of reservoir computer (also known as {\bfi  liquid state machine}) is a significant generalization of the definitions found in the literature, where the readout map $h$ is consistently taken to be linear. In many supervised machine learning applications, the reservoir map is randomly generated (see, for instance, the echo state networks in~\cite{jaeger2001, Jaeger04}) and the memoryless readout is trained so that the output matches a given {\bfi  teaching signal} that we denote by $\mathbf{d} \in \mathbb{R} ^{\Bbb Z } $. Two important advantages of this approach lay on the fact that they reduce the training of a dynamic task to a static problem and, moreover, if the reservoir map is {\it rich} enough, good performances can be indeed attained with just linear readouts that are trained via a (eventually regularized) linear regression that minimizes the Euclidean distance between the output ${\bf y}$ and the teaching signal $ \mathbf{d} $. These features circumvent well-known difficulties in the training of generic recurrent neural networks having to do with bifurcation phenomena~\cite{Doya92} and that, despite recent progress in the regularization and training of deep RNN structures (see, for instance \cite{Graves2013, pascanu:rnn, zaremba}, and references therein), render classical gradient descent methods non-convergent.

The interest for reservoir computing in both the machine learning and the signal processing communities has strongly increased in the last years. One of the main reasons for this fact is that some RC implementations are based on the computational capacities of certain non-neural dynamical systems~\cite{Crutchfield2010}, which opens the door to physical (optical or optoelectronic) realizations  that have already been built using dedicated hardware (see, for instance,~\cite{jaeger2, Atiya2000, Appeltant2011, Rodan2011, SOASforRC, Larger2012, Paquot2012, photonicReservoir2013, swirl:paper, Vinckier2015}) and that have shown unprecedented information processing speeds. 
         
There are two central questions that need to be addressed when designing a machine learning paradigm, namely, the {\bfi  capacity} and the {\bfi  universality} problems. The capacity problem concerns generically the estimation of the error that is going to be committed in the execution of a specific task. In statistical learning and in the approximation theoretical treatment of static neural networks, this estimation has taken the form of generic bounds that incorporate various architecture parameters of the system like in~\cite{maurey:result, Jones, Barron1993, kurkova:sanguineti:curse}. In the specific context of reservoir computing, and in dynamic learning in general, one is interested in various notions of memory capacity that have been the subject of much research~\cite{Jaeger:2002, White2004, Ganguli2008, Hermans2010, dambre2012, GHLO2014_capacity, linearESN, RC3}.

The universality problem consists in showing that the set of input/output functionals that can be generated with a specific architecture is dense in a sufficiently rich class, like the one containing, for example, all continuous or even all measurable functionals. For classical machine learning paradigms like neural networks, this question has given rise to well-known results~\cite{komogorovnn, arnoldnn, sprecherthesis, sprecherI, sprecherII, cybenko, hornik, rueschendorf:thomsen} that show that they can be considered as universal approximators in a static and deterministic setup.

There is no general recipe that allows one to conclude the universality of a given  machine learning approach. The proof strategy depends much on the specific paradigm and, more importantly, on the nature of the inputs and the outputs. In the context of reservoir computing there are several situations for which universality has been established when the inputs/outputs are deterministic. There are two features that influence significantly the level of mathematical sophistication that is needed to conclude universality: first, the compactness of the time domain under consideration and second, if one works in continuous or discrete time. In the following paragraphs we briefly review the results that have already been obtained and, in passing, we present and put in context the contributions contained in this paper.

The compactness of the time domain is crucial because, as we will see later on, universality can be obtained as a consequence of various versions of the Stone-Weierstrass Theorem, which are invariably formulated for functions defined on a compact metric space. When the time domain is compact, this property is naturally inherited by the spaces relevant in the proofs. However, when it is not, it can still be secured using functionals that satisfy a condition introduced in~\cite{Boyd1985} known as the {\bfi  fading memory property}. The distinction between continuous and discrete time inputs is justified by the availability in the continuous setup of different tools coming from functional analysis that do not exist for discrete time.

Reservoir computing universality for compact time domains is a corollary of classical results in systems theory. Indeed, in the continuous time setup, it can be established~\cite{fliess:bilinear, sussmann:bilinear:systems} for linear systems using polynomial readouts and for bilinear systems using linear readouts. In the discrete-time setup, the situation is more convoluted when the readout is linear and required the introduction in~\cite{FliessNormand1980} of the so-called (homogeneous) {\bfi  state-affine systems (SAS)} (see also~\cite{Sontag1979, sontag:polynomial:1979}). The extension of these results to continuous non-compact time intervals was carried out in~\cite{Boyd1985} for fading memory filters using exponentially stable linear RCs with polynomial readouts and their bilinear counterparts with linear readouts (see also~\cite{Maass2000, maass1, corticalMaass, MaassUniversality}). An extension to the non-compact discrete-time setup based on the Stone-Weierstrass theorem is, to our knowledge, not available in the literature and it is one of the main contributions of this paper. This problem has only been tackled from an {internal approximation} point of view, which consists in uniformly approximating  the reservoir and readout maps in \eqref{reservoir equation}-\eqref{readout} in order to obtain an approximation of the resulting filter; this strategy has been introduced in \cite{Matthews:thesis, Matthews1993} for absolutely summable systems. The proofs in those works were unfortunately based on an invalid compactness assumption. Even though  corrections were proposed in \cite{perryman:thesis, Stubberuda}, this approach yields, in the best of cases, universality only within the reservoir filter category, while we aim at proving that statement in the much larger category of fading memory filters. 

The paper is structured in three sections:

\begin{itemize}
\item All the notation and main definitions which are used later on  in the paper are provided in Section~\ref{Notation, definitions, and preliminary discussions}. Important concepts like filters, reservoir filters, and the fading memory property are discussed.
\item Section~\ref{Universality results in the deterministic setup} contains  two different universality results. The first one in Subsection \ref{Universality for fading memory RCs with non-linear readouts} shows that the entire family of fading memory RCs itself is universal, as well as the much smaller one containing all the linear reservoirs with polynomial readouts, when certain spectral restrictions are imposed on the reservoir matrices (see below for details).  The second universality result is contained in Subsection \ref{State-affine systems and universality for fading memory RCs with linear readouts} and is one of the main contributions of the paper. Here we restrict ourselves to reservoir computers with linear readouts which are closer to the type of RCs used in applications. We introduce a non-homogeneous  variant of the state-affine systems in~\cite{FliessNormand1980} and identify sufficient conditions  that guarantee that the associated reservoir computers with linear readouts are causal, time-invariant, and satisfy the echo state and the fading memory properties. Finally, we state a universality result for a  subset of this class which  is shown to be universal in the category of fading memory filters with uniformly bounded inputs.  
\item These universality statements are generalized to the stochastic setup for almost surely uniformly bounded inputs in Section~\ref{Reservoir universality results in the stochastic setup}. In particular, it is  shown that any discrete-time filter that has the fading memory property with almost surely uniformly bounded stochastic inputs can be uniformly approximated by elements in the non-homogeneous state-affine family. 
\end{itemize}

Despite some preexisting work on the uniform approximation in probability of stochastic systems with finite memory \cite{perryman:thesis, Perryman, Stubberud}, the universality result in the stochastic setup is, to our knowledge, the first of its type in the literature and opens the door to new developments in the learning of stochastic processes and their obvious applications to forecasting~\cite{galtier:sto}. In the deterministic setup, RC has been very successful (see for instance~\cite{Jaeger04, pathak:chaos, Pathak:PRL}) in the learning of the attractors of various dynamical systems. This approach is used for forecasting by  path continuation of synthetically learnt proxies, which has led to several orders of magnitude accuracy improvements with respect to most standard dynamical systems forecasting techniques based on Takens' Theorem~\cite{takensembedding}. We expect that the results in this paper should lead to comparable improvements in the density forecasting of stochastic processes. 
\section{Notation, definitions, and preliminary discussions}
\label{Notation, definitions, and preliminary discussions}

\paragraph{Vector and matrix notations. Polynomials.}
A column vector is denoted by a bold lower case  symbol like $\mathbf{r}$ and $\mathbf{r} ^\top $ indicates its transpose. Given a vector $\mathbf{v} \in \mathbb{R}  ^n $, we denote its entries by $v_i$ or $v ^i $, depending on the context, with $i \in \left\{ 1, \dots, n
\right\} $; we also write $\mathbf{v}=(v _i)_{i \in \left\{ 1, \dots, n\right\} }$.  
We denote by $\mathbb{M}_{n ,  m }$ the space of real $n\times m$ matrices with $m, n \in \mathbb{N} $. When $n=m$, we use the symbol $\mathbb{M}  _n $ to refer to the space of square matrices of order 
$n$. $\mathbb{D} _n\subset \mathbb{M}  _n $ is the set of diagonal matrices of order $n$ and $\mathbb{D} $ denotes the set of diagonal matrices of any order.
\nomenclature{$\mathbb{D} _n $}{Space of diagonal matrices of order $n \in \mathbb{N}$}
\nomenclature{$\mathbb{D} $}{Space of diagonal matrices of any order}
Given a vector $\mathbf{v} \in \mathbb{R}  ^n $, we denote by ${\rm diag} (\mathbf{v})$ the diagonal matrix in $\mathbb{M}  _n $ with the elements of $\mathbf{v}  $  as diagonal entries. 
$\mathbb{N}{\rm il} _n^k \subset \mathbb{M} _n  $ is the set of nilpotent matrices in $\mathbb{M} _n  $ of index $k\leq n$ , that is, $A \in \mathbb{N}{\rm il} _n^k $ if and only if $A \in \mathbb{M} _n  $, $A^k=0 $, and $A^l\neq 0 $ for any $l<k $. $\mathbb{N}{\rm il} $ denotes the set of nilpotent matrices of any order and any index.
\nomenclature{$\mathbb{N}{\rm il} _n^k $}{Space of nilpotent matrices of index $k \in \mathbb{N}$ in $\mathbb{M} _n  $}
\nomenclature{$\mathbb{N}{\rm il} $}{Space of nilpotent matrices of any order and any index}
Given a matrix $A \in \mathbb{M}  _{n , m} $, we denote its components by $A _{ij} $ and we write $A=(A_{ij})$, with $i \in \left\{ 1, \dots, n\right\} $, $j \in \left\{ 1, \dots m\right\} $. Given a vector $\mathbf{v} \in \mathbb{R}  ^n $, the symbol $\| \mathbf{v}\|  $ stands for its Euclidean norm. For any $A \in \mathbb{M}  _{n , m} $,  $\|A\| _2  $ denotes its matrix norm induced by the Euclidean norms in $\mathbb{R}^m $ and $\mathbb{R} ^n $,  and satisfies~\cite[Example 5.6.6]{horn:matrix:analysis} that $\|A\| _2=\sigma_{{\rm max}}(A)$, with $\sigma_{{\rm max}}(A)$  the largest singular value of $A$. $\|A\| _2  $ is sometimes referred to as the spectral norm of $A$~\cite{horn:matrix:analysis}.

Let $V _1, V _2, W _1, W _2  $  be vector spaces. The symbols $V _1\oplus V _2 $ and $V _1\otimes V _2 $ denote the corresponding direct sum and tensor product vector spaces \cite{hungerford:algebra}, respectively, of $V _1  $ and $V _2$. Given any ${\bf v} _1 \in V _1 $ and ${\bf v} _2 \in V _2 $, the vectors ${\bf v} _1\oplus {\bf v} _2 \in V _1\oplus V _2 $ and ${\bf v} _1\otimes {\bf v} _2 \in V _1\otimes V _2 $ are the direct sum and the (pure) tensor product of ${\bf v} _1 $ and ${\bf v} _2$, respectively. Given two linear maps $A _1: V _1 \longrightarrow W _1 $ and $A _2: V _2 \longrightarrow W _2 $, we  denote by $A _1\oplus A _2: V _1\oplus V _2 \longrightarrow W _1\oplus W _2 $ and $A _1\otimes A _2: V _1\otimes V _2 \longrightarrow W _1\otimes W _2 $ the associated direct sum and tensor product maps, respectively, defined by $A _1\oplus A _2 \left({\bf v} _1\oplus {\bf v} _2\right):= A _1 \left({\bf v} _1\right)\oplus A _2 \left({\bf v} _2\right)$ and $A _1\otimes A _2 \left({\bf v} _1\otimes {\bf v} _2\right):= A _1 \left({\bf v} _1\right)\otimes A _2 \left({\bf v} _2\right)$. The matrix representation of $A _1 \oplus A _2 $ is obtained by concatenating in a block diagonal matrix the matrix representations of $A _1 $ and $A _2$. As to the matrix representation of $A _1 \otimes A _2 $ it is obtained via the Kronecker product of the matrix representations of $A _1 $ and $A _2$ \cite{horn:matrix:analysis}.

Given an element ${\bf z} \in \mathbb{R} ^n $, we denote by $\mathbb{R}[{\bf z}] $ the real-valued multivariate polynomials on ${\bf z} $ with real coefficients. Analogously, ${\rm Pol}(\mathbb{R} ^n , \mathbb{R})$ will denote the set of real-valued polynomials on  $\mathbb{R} ^n $. When $z \in \mathbb{R} $ and $m,n \in \mathbb{N} $, we define the set $\mathbb{M}_{m,n}[z] $ of $\mathbb{M}_{m,n} $-valued polynomials  on $z$ with coefficients in $\mathbb{M}_{m,n} $ as
\begin{equation}
\label{polynomial matrices}
\mathbb{M}_{m,n}[z] :=\{A _0+zA _1+z ^2A _2+ \cdots + z ^r A _r\mid r \in \mathbb{N}, A _0,A _1,A _2, \ldots , A _r \in \mathbb{M}_{m,n}\}.
\end{equation}
$\mathbb{N}{\rm il} _n^k[z] \subset \mathbb{M} _n[z]  $ is the set of nilpotent $\mathbb{M}_{n} $-valued polynomials  on $z$ of index $k $, that is, $p (z) \in \mathbb{N}{\rm il} _n^k[z] $ whenever $k $ is the smallest natural number for which $p (z) ^k= {\bf 0} $, for all $z \in \mathbb{R} $. $\mathbb{N}{\rm il} [z] $ is the set of matrix-valued nilpotent polynomials  on $z$ of any order and any index.
\nomenclature{$\mathbb{N}{\rm il} _n^k[z]$}{Space of nilpotent $\mathbb{M}_{n} $-valued polynomials  on $z$ with coefficients in $\mathbb{M}_{n} $ of index $k $}
\nomenclature{$\mathbb{N}{\rm il} [z]$}{Space of matrix-valued nilpotent polynomials  on $z$ of any order and any index}

\paragraph{Sequence spaces.}
 
$\mathbb{N}$ denotes the set of natural numbers with the zero element included. $\Bbb Z $ (respectively, $\Bbb Z _+ $ and $\Bbb Z _- $) are the integers (respectively, the positive and the negative integers).
The symbol $(\mathbb{R}^n) ^{\Bbb Z } $ denotes the set of infinite real sequences of the form ${\bf z}=(\ldots, {\bf z} _{-1}, {\bf z} _0, {\bf z} _1, \ldots) $, $ {\bf z} _i \in \mathbb{R}^n $, $i \in \Bbb Z $; $(\mathbb{R}^n) ^{\Bbb Z _ -} $ and $(\mathbb{R}^n) ^{\Bbb Z _ +} $ are the subspaces consisting of, respectively, left and right infinite sequences: $(\mathbb{R}^n) ^{\Bbb Z _ -}=\{{\bf z}=(\ldots, {\bf z} _{-2}, {\bf z} _{-1}, {\bf z} _0) \mid {\bf z} _i \in \mathbb{R}^n, i \in \mathbb{Z}_{-}\}$, $(\mathbb{R}^n) ^{\Bbb Z _ +}=\{{\bf z}=({\bf z} _0, {\bf z} _1, {\bf z} _2, \ldots) \mid {\bf z} _i \in \mathbb{R}^n, i \in \mathbb{Z}_{+}\}$.  Analogously, $(D_n) ^{\Bbb Z } $, $(D_n) ^{\Bbb Z _ -} $, and $(D_n) ^{\Bbb Z _ +} $ stand for (semi-)infinite sequences with elements in the subset $D_n\subset \mathbb{R}^n $.
In most cases we shall use in these infinite product spaces either the product topology (see~\cite[Chapter 2]{Munkres:topology}) or the topology induced by the supremum norm $\| {\bf z}\| _{\infty}:= {\rm sup}_{ t \in \Bbb Z} \left\{\| {\bf z} _t
\|\right\}$. The symbols $\ell ^{\infty}(\mathbb{R}^n) $ and $\ell_{\pm} ^{\infty}(\mathbb{R}^n) $ will be used to denote the Banach spaces formed by the elements in  those infinite product spaces that have a finite supremum norm $\| \cdot \| _{\infty} $. The symbol $B_n({\bf v}, M) \subset \mathbb{R} ^n $, 
denotes the open ball of radius $M >0$ and center ${\bf v} \in {\Bbb R}^n$ with respect to the Euclidean norm. The bars over sets stand for  topological closures, in particular, $\overline{B_n({\bf v}, M)} $ is the closed ball.

\paragraph{Filters.}

We will refer to the maps of the type $U: (D_n) ^{\Bbb Z} \longrightarrow \mathbb{R} ^{\Bbb Z} $ as {\bfi  filters} or {\bfi  operators} and to those like $H: (D_n) ^{\Bbb Z} \longrightarrow \mathbb{R} $ (or $H: (D_n) ^{\Bbb Z_\pm} \longrightarrow \mathbb{R} $) as {\bfi  functionals}. A filter $U$ is called {\bfi  causal} when for any two elements ${\bf z} , \mathbf{w} \in (D_n) ^{\Bbb Z}  $  that satisfy that ${\bf z} _\tau = \mathbf{w} _\tau$ for all $\tau \leq t  $, for any given  $t \in \Bbb Z $, we have that $U ({\bf z}) _t= U ({\bf w}) _t $. Let $U _\tau:(D_n) ^{\Bbb Z} \longrightarrow(D_n) ^{\Bbb Z} $, $\tau\in \Bbb Z $, be the time delay operator defined by $U _\tau ({\bf z}) _t:= {\bf z}_{t- \tau}$.
The filter $U$ is called {\bfi  time-invariant} when it commutes with the time delay operator, that  is, $U _\tau \circ U=U  \circ U_\tau $ (in this expression, the two time delay operators $U _\tau $ have  to be understood as defined in the appropriate sequence spaces). We recall (see for instance~\cite{Boyd1985}) that there is a bijection between causal time-invariant filters and functionals on $(D_n)^{\Bbb Z _-} $. Indeed, given a time-invariant filter $U$, we can associate to it a functional $H _U: (D_n)^{\Bbb Z _-} \longrightarrow \mathbb{R} $ via the assignment $H _U ({\bf z}):= U({\bf z} ^e) _0 $, where ${\bf z} ^e \in (D_n)^{\Bbb Z } $ is an arbitrary extension of ${\bf z} \in (D_n)^{\Bbb Z _-} $ to $ (D_n)^{\Bbb Z } $. Conversely, for any functional  $H: (D_n)^{\Bbb Z _-} \longrightarrow \mathbb{R} $, we can define a time-invariant causal filter $U_H: (D_n) ^{\Bbb Z} \longrightarrow \mathbb{R}^{\Bbb Z} $ by $U_H({\bf z}) _t:= H((\mathbb{P}_{\Bbb Z_-} \circ U _{-t}) ({\bf z})) $, where $U _{-t} $ is the $(-t)$-time delay operator and $\mathbb{P}_{\Bbb Z_-}: (D_n)^{\Bbb Z} \longrightarrow (D_n)^{\Bbb Z _-} $ is the natural projection. It is easy to verify that:
\begin{eqnarray*}
H_{U _H}&=& H,\quad \mbox{for any functional} \quad H: (D_n)^{\Bbb Z _-} \longrightarrow \mathbb{R},\\
U_{H _U} &= &U, \quad \mbox{for any causal time-invariant filter} \quad U: (D_n) ^{\Bbb Z} \longrightarrow \mathbb{R} ^{\Bbb Z}.
\end{eqnarray*}
Additionally, let $H _1, H _2:(D_n)^{\Bbb Z_-} \longrightarrow \mathbb{R}$ and $\lambda \in \mathbb{R} $, then $U_{H _1+ \lambda H _2} ({\bf z}) = U_{H _1}( {\bf z})+ \lambda U_{H _2} ({\bf z}) $, for any ${\bf z} \in (\mathbb{R} ^n) ^{\Bbb Z} $. In the following pages and when the discussion will take place in a causal and time-invariant setup, we will use the term filter to denote exchangeably the associated functional and the filter itself.

\paragraph{Reservoir filters.}

Consider now the RC system determined by~\eqref{reservoir equation}--\eqref{readout}. It is worth mentioning that, unlike in those expressions, the reservoir and the readout maps are in general defined only on subsets $D _N, D' _N \subset \mathbb{R}^N $ and $D_n\subset \mathbb{R}^n $ and not on the entire Euclidean spaces ${\Bbb R}^N $ and ${\Bbb R}^n$, that is, $F: D _N\times D_n\longrightarrow  D' _N$ and $h: D'_N \rightarrow \mathbb{R}$. Reservoir systems determine a filter when the following existence and uniqueness property holds: for each ${\bf z} \in \left(D_n\right)^{\mathbb{Z}} $  there exists a unique ${\bf x} \in \left(D_N\right)^{\mathbb{Z}} $ such that for each $t \in \Bbb Z $, the relation~\eqref{reservoir equation} holds. This condition is known in the literature as the {\bfi  echo state property}~\cite{jaeger2001, Yildiz2012} and has deserved much attention in the context of echo state networks~\cite{Jaeger04, Buehner:ESN, zhang:echo, Wainrib2016, Manjunath:Jaeger}. The echo state property formulated for infinite (or semi-infinite) inputs guarantees that the output of the filter at any given time does not depend on initial conditions.
We emphasize that this is a genuine condition that is not automatically satisfied by all RC systems.

We will denote by $U ^F: (D_n)^{\Bbb Z} \longrightarrow(D_N)^{\Bbb Z} $ the filter determined by the reservoir map via~\eqref{reservoir equation}, that is, $U ^F ({\bf z}) _t := \mathbf{x} _t \in D_N $,  and by $U ^F _h: (D_n)^{\Bbb Z} \longrightarrow\mathbb{R}^{\Bbb Z} $ the one determined by the entire reservoir system, that is,  $U ^F_h ({\bf z}) _t := h \left(U ^F ({\bf z}) _t\right)=y _t $.  $ U ^F_h $ will be called the {\bfi  reservoir filter} associated to the RC system~\eqref{reservoir equation}--\eqref{readout}. The filters $U ^F $ and $U ^F _h $ are causal by construction and it can also be shown that they are necessarily time-invariant \cite{RC7}. We can hence associate  to $U ^F _h $ a {\bfi  reservoir functional} $H^F _h : (D _n)^{\Bbb Z _-} \longrightarrow \mathbb{R}$ determined by $H^F _h:=H_{U ^F _h} $. 

\paragraph{Weighted norms and the fading memory property (FMP).}

Let $w : \mathbb{N} \longrightarrow (0,1] $ be a decreasing sequence with zero limit. We define the associated {\bfi  weighted norm } $\| \cdot \| _w $ on $(\mathbb{R}^n)^{\Bbb Z _{-}}$ associated to the {\bfi  weighting sequence} $w$ as the map:
\begin{eqnarray*}
\begin{array}{cccc}
\| \cdot \| _w :& (\mathbb{R}^n)^{\Bbb Z _{-}} & \longrightarrow & \overline{\mathbb{R}^+}\\
	&{\bf z} &\longmapsto &\| {\bf z} \| _w:= \sup_{t \in \Bbb Z_-}\{\| {\bf z}_t w_{-t}\|\},
\end{array}
\end{eqnarray*} 
where $\| \cdot \| $ denotes the Euclidean norm in $\mathbb{R} ^n $. It is worth noting that the space
\begin{equation}
\label{lww space}
\ell ^{\infty}_w({\Bbb R}^n):= \left\{{\bf z}\in \left(\mathbb{R}^n\right)^{\mathbb{Z}_{-}}\mid \| {\bf z}\| _w< \infty\right\},
\end{equation}
endowed with weighted norm  $\| \cdot \| _w $ forms a Banach space \cite{RC7}.

All along the paper, we will work with {\bfi  uniformly bounded}  families of sequences, both in the deterministic and the stochastic setups. The two main properties of these subspaces in relation with the weighted norms are spelled out in the following two lemmas. 

\begin{lemma}
\label{uniformly bounded}
Let $M >0 $ and let $K_{M} $ be the set of elements in $\left(\mathbb{R}^n\right)^{\mathbb{Z}_{-}} $ which are uniformly bounded  by $M$, that is,
\begin{equation}
\label{Kset}
K_{M}:=\left\{ {\bf z} \in \left(\mathbb{R}^n\right)^{\mathbb{Z}_{-}} \mid \| {\bf z}_t\| \leq M \quad \mbox{for all} \quad t \in \Bbb Z _{-} \right\}= \overline{B_n({\bf 0}, M)}^{\mathbb{Z}_{-}},
\end{equation}
with $\overline{B_n({\bf 0}, M)} \subset \mathbb{R} ^n $  
\nomenclature{$B_n({\bf 0}, M) $}{Ball of radius $M $ and center ${\bf 0} $ in ${\Bbb R}^n $ with respect to the Euclidean norm}
the closed ball of radius $M $ and center ${\bf 0} $ in ${\Bbb R}^n $ with respect to the Euclidean norm. Then, for any weighting sequence $w $ and ${\bf z} \in K_{M} $, we have that $\| {\bf z}\|_w < \infty $. 

Additionally, let $\lambda, \rho \in (0,1) $  and let $ w , w ^\rho, w ^{1- \rho} $ be the weighting sequences given by $w _t:= \lambda ^t $, $w_t ^\rho:= \lambda ^{ \rho t}$, $w_t^{1-\rho}:= \lambda^{(1- \rho) t} $, $t \in \mathbb{N}  $. Then, the  series $\sum_{t=0}^{\infty}\| {\bf z} _{-t}\| w _t $ is absolutely convergent and satisfies the inequalities:
\begin{eqnarray}
\sum_{t=0}^{\infty}\| {\bf z} _{-t}\| w _t &= &\sum_{t=0}^{\infty}\| {\bf z} _{-t}\| \lambda ^t \leq \| {\bf z}\|_{w^{1- \rho}}\frac{1}{1- \lambda ^\rho},\label{ineq1rho}\\
\sum_{t=0}^{\infty}\| {\bf z} _{-t}\| w _t &= &\sum_{t=0}^{\infty}\| {\bf z} _{-t}\| \lambda ^t \leq \| {\bf z}\|_{w^{ \rho}}\frac{1}{1- \lambda ^{1-\rho}}.\label{ineq2rho}
\end{eqnarray}
\end{lemma}

The following result is a discrete-time version of Lemma 1 in~\cite{Boyd1985} that is easily obtained by noticing that in the discrete-time setup all functions are trivially continuous if we consider the discrete topology for their domains and, moreover, all families of functions are equicontinuous. A proof is given in the appendices for the sake of completeness.

\begin{lemma}
\label{uniformly bounded and equicontinuous}
Let $M>0 $ and let $K_{M} $ be as in ~\eqref{Kset}.
Let $w : \mathbb{N} \longrightarrow (0,1] $ be a weighting sequence. Then $K_{M }$ is a compact topological space when endowed with the relative topology inherited from the norm topology in the Banach space $(\ell ^{\infty}_w({\Bbb R}^n), \left\|\cdot \right\|_w) $.
\end{lemma}

\begin{definition}
\label{fmp definition}
Let $D_n\subset \mathbb{R} ^n$ and let $H_U: (D_n)^{\Bbb Z _{-}}\longrightarrow \mathbb{R}  $ be the functional associated to the causal and time-invariant filter $U: (D_n)^{\Bbb Z } \longrightarrow\mathbb{R}^{\Bbb Z }  $. We say that $U$ has the {\bfi  fading memory property (FMP)} whenever there exists a weighting sequence $w: \mathbb{N} \longrightarrow (0,1] $ such that  the map
$H_U: ((D_n)^{\Bbb Z _{-}}, \| \cdot \| _w)\longrightarrow \mathbb{R}  $ is continuous. This means that for any ${\bf z} \in (D_n)^{\Bbb Z _{-}} $ and any $\epsilon>0  $, there exists a $\delta(\epsilon)> 0 $ such that for any ${\bf s} \in (D_n)^{\Bbb Z _{-}}$ that satisfies that
\begin{equation*}
\| {\bf z} - {\bf s}\|_w=\sup_{t \in \Bbb Z_-}\{\| ({\bf z}_t-{\bf s}_t) w_{-t}\|\}< \delta(\epsilon), \quad \mbox{then} \quad |H _U({\bf z})-H _U({\bf s})|< \epsilon.
\end{equation*}
If the weighting sequence $w$ is such that $w _t= \lambda ^t $, for some $\lambda\in (0,1) $ and all $t \in \mathbb{N}  $,  then $U$ is said to have the $\lambda $-{\bfi exponential fading memory property}.
\end{definition}

\begin{remark}
\normalfont
This formulation of the fading memory property is due to Boyd and Chua \cite{Boyd1985} and it is the key concept that allowed these authors to extend to non-compact time intervals the first filter universality results formulated in the classical works of Fr\'echet \cite{frechet:volterra_series}, Volterra, and Wiener \cite{wiener:book, brilliant:volterra, george:volterra}, always under compactness assumptions on the input space and the time interval in which inputs are defined.
\end{remark}

\begin{remark}
\normalfont
In the context of reservoir filters, the fading memory property is in some occasions related to the {\bfi  Lyapunov stability} of the autonomous system associated to the reservoir map by setting the input sequence equal to zero. This connection has been made explicit, for example, for discrete-time nonlinear state-space models that are affine in their inputs, and have direct feed-through term in the output ~\cite{Zang:iglesias} or for time delay reservoirs ~\cite{RC4pv}. 
\end{remark}

\begin{remark}
\normalfont
Time-invariant fading memory filters always have the {\bfi bounded input, bounded output (BIBO)} property. Indeed, if for simplicity we consider functionals $H _U $ that map the zero input  to zero, that is $H _U({\bf 0})=0 $, and we want bounded outputs such that $|H _U ({\bf z})|<k$, for a given constant $k>0$, by Definition ~\ref{fmp definition} it suffices to consider inputs ${\bf z} \in (\mathbb{R} ^N)^{\Bbb Z _-} $ such that  $\| {\bf z} \|_{\infty}:= \sup_{t \in \Bbb Z_-}\{\| {\bf z}_t\|\}<\delta(k) $. Indeed, if $H _U $ has the FMP with respect to a weighting sequence $w $, then $\| {\bf z}\| _w \leq \| {\bf z} \|_{\infty}<\delta(k) $ and hence $|H _U ({\bf z})|<k$, as required. Another important dynamical implication of the fading memory property is the {\bfi uniqueness of steady states} or, equivalently, the asymptotic independence of the dynamics on the initial conditions. See Theorem 6 in \cite{Boyd1985} for details about this fact.
\end{remark}

The following lemma, which will be used later on in the paper, spells out how the FMP depends on the weighting sequence used to define it.

\begin{lemma}
\label{weighting sequence dependence}
Let $D_n\subset \mathbb{R} ^n$ and let $H_U: (D_n)^{\Bbb Z _{-}}\longrightarrow \mathbb{R}  $ be the functional associated to the causal and time-invariant filter $U: (D_n)^{\Bbb Z } \longrightarrow(\mathbb{R})^{\Bbb Z }  $. If $H_U$ has the FMP with respect to a given weighting sequence $w $, then it also has it with respect to any other weighting sequence $w ' $ which satisfies
\begin{equation*}
\frac{w _t}{w' _t}<\lambda, \quad \mbox{for a fixed} \quad \lambda>0 \quad \mbox{and for all} \quad t \in \mathbb{N}.
\end{equation*}
In particular, the thesis of the lemma holds when $w' $ dominates $w $, that is when $\lambda=1$.
\end{lemma}

It can be shown \cite{RC7} that when in this lemma the set $(D_n)^{\Bbb Z _{-}}$ is made of uniformly bounded sequences, that is, $(D_n)^{\Bbb Z _{-}}  = K _M $, with $K _M $ as in \eqref{Kset} then, if a filter has the FMP with respect to a given weighting sequence, it necessarily has the same property with respect to any other weighting sequence.

\section{Universality results in the deterministic setup}
\label{Universality results in the deterministic setup}

The goal of this section is identifying families of reservoir filters that are able to uniformly approximate any time-invariant, causal, and fading memory  filter with deterministic inputs with any desired degree of accuracy. Such families of reservoir computers are said to be {\bfi  universal}.

The main mathematical tool that we use is the Stone-Weierstrass theorem  for polynomial subalgebras of real-valued functions defined on compact metric spaces. This approach provides us with universal families of filters as long as we can prove that, roughly speaking, their elements form polynomial algebras using a product defined in the space of functionals. More specifically, if $D_n\subset \mathbb{R} ^n$ and $H_{U_1}, H_{U_2}: (D_n)^{\Bbb Z _{-}}\longrightarrow \mathbb{R}  $ are the functionals associated to the causal and time-invariant filters $U_1, U_2: (D_n)^{\Bbb Z } \longrightarrow\mathbb{R}^{\Bbb Z }  $, we readily define their product $H_{U_1} \cdot H_{U_2}: (D_n)^{\Bbb Z _{-}}\longrightarrow \mathbb{R}  $ and linear combination $H_{U_1}+ \lambda H_{U_2}: (D_n)^{\Bbb Z _{-}}\longrightarrow \mathbb{R}  $, $\lambda \in \mathbb{R}  $, as
\begin{equation}
\label{polynomial definition}
(H_{U_1} \cdot H_{U_2}) \left({\bf z}\right):= H_{U_1}\left({\bf z}\right) \cdot H_{U_2} \left({\bf z}\right), \quad (H_{U_1} + \lambda H_{U_2}) \left({\bf z}\right):= H_{U_1}\left({\bf z}\right) + \lambda H_{U_2} \left({\bf z}\right), \quad {\bf z}\in (D_n)^{\Bbb Z _{-}}.
\end{equation} 

This section contains two different universality results. The first one shows that polynomial algebras of filters generated by reservoir systems using the operations in \eqref{polynomial definition} that have the fading memory property and that separate points, are able to approximate any fading memory filter.  Two important consequences of this result are that the entire family of fading memory RCs itself is universal, as well as the one containing all the linear reservoirs with polynomial readouts, when certain spectral restrictions are imposed on the reservoir matrices (see below for details).
In the second result, we restrict ourselves to reservoir computers with linear readouts and introduce the non-homogeneous state-affine family in order to be able to obtain a similar universality statement. The linearity restriction on the readouts makes this universality statement closer to the type of RCs used in applications and to the standard notion of reservoir system that one commonly finds in the literature \cite{lukosevicius}.

The first  result can be seen as a discrete-time version of the one in~\cite{Boyd1985} for continuous-time filters, while the second one is an extension to infinite time intervals of the main approximation result in~\cite{FliessNormand1980}, which was originally formulated for compact time intervals using homogeneous state-affine systems.

\subsection{Universality for fading memory RCs with non-linear readouts}
\label{Universality for fading memory RCs with non-linear readouts}

The following statement is a direct consequence of the compactness result in Lemma~\ref{Kset} and the Stone-Weierstrass, as formulated in Theorem 7.3.1 in~\cite{Dieudonne:analysis}. See Appendix~\ref{proof of universality deterministic general} for a detailed proof.

All along this subsection, we work with reservoir filters with uniformly bounded inputs in a set  $K_M \subset (\mathbb{R}^n) ^{\Bbb Z _-} $, as  defined in~\eqref{Kset}. These filters are generated by reservoir systems $F: D _N\times \overline{B _n({\bf 0}, M)}\longrightarrow  D _N$ and $h: D_N \rightarrow \mathbb{R}$, for some $n, N \in \mathbb{N} $, $M>0 $,   and $D_N \subset \mathbb{R} ^N $.

\begin{theorem}
\label{universality deterministic general}
Let $K_M \subset (\mathbb{R}^n) ^{\Bbb Z _-} $ be a subset of the type defined in~\eqref{Kset}, $I $ an index set, and let 
\begin{equation}
\label{generating family reservoirs}
\mathcal{R}:=\{H_{h _i}^{F _i}:K_M \longrightarrow \mathbb{R}\mid h _i \in  C^{\infty}(D_{N _i}), F _i: D_{N _i} \times \overline{B _n({\bf 0}, M)}  \longrightarrow D_{N _i}, i \in I, N _i\in  \mathbb{N}\}
\end{equation} 
\nomenclature{$\mathcal{R} $}{Set of reservoir filters defined on $K_M$}
be a set of reservoir filters defined on $K_M$ that have the FMP with respect to a given weighted norm $\| \cdot \| _w  $. Let $\mathcal{A}(\mathcal{R}) $ 
\nomenclature{$\mathcal{A}(\mathcal{R}) $}{Polynomial algebra generated by the set $\mathcal{R} $ of reservoir filters defined on $K_M$}
be the polynomial algebra generated by $\mathcal{R}  $, that is, the set formed by finite products and linear combinations of elements in $\mathcal{R} $ according to the operations defined in \eqref{polynomial definition}. If the algebra $\mathcal{A}(\mathcal{R})$ contains the constant functionals and separates the points in $K_M$, then any causal, time-invariant fading memory filter $H:K_M \longrightarrow \mathbb{R} $ can be uniformly approximated  by elements in $\mathcal{A}(\mathcal{R})$, that is, $\mathcal{A}(\mathcal{R}) $ is dense in the set $(C ^0(K_M),\|\cdot \| _w)$ of real-valued continuous functions on $(K_M, \| \cdot \| _w)$. More explicitly, this implies that for any fading memory filter $H$ and any $\epsilon>0 $, there exist a finite set of indices $\{i _1, \ldots, i _r\} \subset I $ and a polynomial $p: \mathbb{R} ^r \longrightarrow \mathbb{R} $ such that 
\begin{equation*}
\| H - H _h ^F\| _{\infty} := \sup_{{\bf z} \in K_M} \{|H ({\bf z})-H_h^F ({\bf z}) |\}< \epsilon \quad \mbox{with} \quad h:=p(h_{i _1}, \ldots, h_{i _r})\quad \mbox{and} \quad F:=(F_{i _1}, \ldots, F_{i _r}).
\end{equation*}
\end{theorem}

An important fact is that {\it the polynomial algebra $\mathcal{A}(\mathcal{R})$ generated by a set formed by fading memory reservoir filters  is made of fading memory reservoir filters}.  Indeed, let $h _i \in  C^{\infty}(D_{N _i}) $, $F _i: D_{N _i} \times \overline{B _n({\bf 0}, M)}  \longrightarrow D_{N _i}$, $i \in \{1,2\} $, and $\lambda \in \mathbb{R} $. Then, the product $H_{h _1}^{F _1}\cdot H_{h _2}^{F _2} $ and the linear combination $H_{h _1}^{F _1}+ \lambda H_{h _2}^{F _2} $ filters, as they were defined in \eqref{polynomial definition}, are such that 
\begin{eqnarray}
H_{h _1}^{F _1}\cdot H_{h _2}^{F _2}&= &H_{h}^{F}, \quad  \mbox{with} \quad  h:= h _1\cdot h _2 \in C^{\infty}(D_{N _1}\times D_{N _2}), \label{pol algebra product}\\
H_{h _1}^{F _1}+ \lambda H_{h _2}^{F _2}&= &H_{h'}^{F}, \quad  \mbox{with} \quad  h':= h _1+ \lambda h _2 \in C^{\infty}(D_{N _1}\times D_{N _2}), \label{pol algebra sum}
\end{eqnarray}
and where $F: (D_{N _1}\times D_{N _2}) \times \overline{B _n({\bf 0}, M)} \longrightarrow (D_{N _1}\times D_{N _2})$ is given by
\begin{equation}
\label{direct filter}
 F(((\mathbf{x} _1) _t, (\mathbf{x} _2) _t), {\bf z}_t):= \left(F _1((\mathbf{x} _1) _t, {\bf z}_t), F _2((\mathbf{x} _2) _t, {\bf z}_t)\right),
\end{equation}
for any $((\mathbf{x} _1) _t, (\mathbf{x} _2) _t) \in D_{N _1} \times D _{N _2} $,  ${\bf z}_t \in \overline{B _n({\bf 0}, M)}$, and $ t \in \mathbb{Z}_{-} $. We emphasize that the functionals $H_{h}^{F} $  and $H_{h'}^{F} $ in \eqref{pol algebra product} and \eqref{pol algebra sum} are well defined because if the reservoir maps $F _1 $ and $F_2$ satisfy the echo state property then so does $F$. Indeed, if $ \mathbf{x} _1\in \left(D_{N _1}\right)^{\mathbb{Z}} $ and $ \mathbf{x} _2\in \left(D_{N _2}\right)^{\mathbb{Z}} $ are the solutions of the reservoir equation~\eqref{reservoir equation} for $F _1 $ and $F _2  $ associated to the input ${\bf z} \in K _M $, then so is $(\mathbf{x} _1, \mathbf{x} _2) \in \left(D_{N _1}\times D_{N _2}\right)^{\mathbb{Z}} $, defined by $(\mathbf{x} _1, \mathbf{x} _2)_t:=((\mathbf{x} _1) _t, (\mathbf{x} _2) _t)$, for $F$ in \eqref{direct filter}.

This observation has as a consequence that the set formed by {\it all} the RC systems that have the echo state property and the FMP with respect to a given weighted norm $\| \cdot \| _w  $ form a polynomial algebra that contains the constant functions (they can be obtained by using as readouts constant elements in $C^{\infty}(D_{N _i}) $) and separates points (take the trivial reservoir map $F(\mathbf{x}, {\bf z})= {\bf z} $ and use the separation property of $C^{\infty}(D_{N _i})$ together with time-invariance). This remark and Theorem~\ref{universality deterministic general} yield the following corollary.

\begin{corollary}
\label{RCs are universal deterministic}
Let $K_M \subset (\mathbb{R}^n) ^{\Bbb Z _-} $ be a subset as defined in~\eqref{Kset} and let 
\begin{equation}
\label{generating family reservoirs all}
\mathcal{R}_w:=\{H_{h }^{F }:K_M \longrightarrow \mathbb{R}\mid h  \in  C^{\infty}(D_N), F : D_N \times \overline{B _n({\bf 0}, M)}  \longrightarrow D_N, N \in  \mathbb{N}\}
\end{equation} 
\nomenclature{$\mathcal{R}_w $}{Set of reservoir filters defined on $K_M$  that have the FMP with respect to a given weighted norm $\| \cdot \| _w  $}
be the set of all reservoir filters with uniformly bounded inputs in $K_M$ and that have the FMP with respect to a given weighted norm $\| \cdot \| _w  $. Then $\mathcal{R}_w$ is universal, that is, it is dense in the set $(C ^0(K_M),\|\cdot \| _w)$ of real-valued continuous functions on $(K_M, \| \cdot \| _w)$. 
\end{corollary}

\begin{remark}
\normalfont
The stability of reservoir filters under products and linear combinations in \eqref{pol algebra product}-\eqref{pol algebra sum} is a feature that allows us, in Corollary \ref{RCs are universal deterministic} and in some of the results that follow later on, to identify families of reservoir filters that are able to approximate any fading memory filter. This fact is a requirement for the application of the Stone-Weierstrass theorem but does not mean that we have to carry  those operations out in the construction of approximating filters, which would indeed be difficult to implement in specific applications.
\end{remark}

According to the previous corollary,  reservoir filters that have the FMP are able to approximate any time-invariant fading memory filter. We now show that actually a much smaller family of reservoirs suffices to do that, namely, certain classes of linear reservoirs with polynomial readouts. Consider the RC system determined by the expressions
\begin{empheq}[left={\empheqlbrace}]{align}
{\bf x } _t & =A\mathbf{x}_{t-1}+ {\bf c}{\bf z} _t,\quad A \in \mathbb{M} _N, {\bf c} \in \mathbb{M} _{N,n},
\label{linear reservoir equation}
\\
y _t & =h (\mathbf{x} _t), \quad h \in \mathbb{R}[\mathbf{x}]. \label{linear readout}
\end{empheq}
If this system has a reservoir filter associated (the next result provides a sufficient condition for this to happen) we  denote by $H^{A, {\bf c}} _h :K_M \longrightarrow \mathbb{R} $ 
\nomenclature{$H^{A, {\bf c}} _h :K_M \longrightarrow \mathbb{R} $}{Linear reservoir functional determined by $A, {\bf c} $, and the polynomial $h$}
\nomenclature{$U^{A, {\bf c}} _h :K_M \longrightarrow \mathbb{R} ^{\mathbb{Z}}$}{Linear reservoir filter determined by $A, {\bf c} $, and the polynomial $h$}
the associated functional and we  refer to it as the {\bfi  linear reservoir functional} determined by $A, {\bf c} $, and the polynomial $h$. 
These filters exhibit the following universality property that is proved in Appendix~\ref{proof of corollary linear universal}.

\begin{corollary}
\label{linear universality deterministic}
Let $K_M \subset (\mathbb{R}^n) ^{\Bbb Z_- } $ be a subset of the type defined in~\eqref{Kset} and let $0<\epsilon<1$. Consider the set $\mathcal{L} _\epsilon $ formed by all the linear reservoir systems as in \eqref{linear reservoir equation}-\eqref{linear readout} determined by matrices $A \in \mathbb{M}_N $ such that $\sigma_{{\rm max}}(A)< 1- \epsilon $. 
\nomenclature{$\mathcal{L} _\epsilon $}{Set of linear reservoir systems determined by matrices $A \in \mathbb{M}_N $ such that $\sigma_{{\rm max}}(A)< 1- \epsilon $}
Then, the elements in $\mathcal{L} _\epsilon $ generate $\lambda  _\rho$-exponential fading memory reservoir functionals, with $\lambda _\rho:=(1- \epsilon) ^\rho $, for any $\rho \in (0, 1)$. Equivalently, $\mathcal{L} _\epsilon \subset \mathcal{R}_{w ^\rho} $, with $w^\rho _t:= \lambda_{\rho} ^{ t} $, and  $\mathcal{R}_{w ^\rho} $   as in  \eqref{generating family reservoirs all}.
These functionals can be explicitly written as:
\begin{equation}
\label{expression linear filter}
H^{A, {\bf c}}_h({\bf z})=h \left(\sum _{i=0}^{\infty} A^i {\bf c} {\bf z}_{-i}\right), \quad \mbox{for any} \quad {\bf z} \in K_M.
\end{equation}
This family is dense in $(C ^0(K_M), \| \cdot \| _{w ^\rho}) $. 

The same universality result can be stated for the following two smaller subfamilies of $\mathcal{L} _\epsilon$:
\begin{description}
\item [(i)]  The family $\mathcal{DL} _\epsilon \subset \mathcal{L} _\epsilon$ formed by the linear reservoir systems in $\mathcal{L} _\epsilon$ determined by diagonal matrices $A \in \mathbb{D}$ such that $\sigma_{{\rm max}}(A)< 1- \epsilon $.
\item [(ii)]  The family $\mathcal{NL} \subset \mathcal{L} _\epsilon$ formed by the linear reservoir systems determined by nilpotent matrices  $A \in \mathbb{N}{\rm il} $.
\nomenclature{$\mathcal{DL} _\epsilon $}{Set of linear reservoir systems determined by diagonal matrices $A \in \mathbb{D}$ such that $\sigma_{{\rm max}}(A)< 1- \epsilon $}
\nomenclature{$\mathcal{NL}$}{Set of linear reservoir systems determined by nilpotent matrices $A \in \mathbb{N}{\rm il} $}
\end{description}
\end{corollary}

\begin{remark}
\normalfont
The elements in the family $\mathcal{NL} $ belong automatically to $\mathcal{L} _\epsilon$ because the eigenvalues of a nilpotent matrix are always zero. This implies that if a linear reservoir system is determined by a nilpotent matrix $A \in \mathbb{N}{\rm il} _N ^k $ of index $k\leq N$, then the reservoir functional $H^{A, {\bf c}}_h $ is automatically well-defined and given by a finite version of \eqref{expression linear filter}, that is,
\begin{equation}
\label{expression linear filter nilpotent}
H^{A, {\bf c}}_h({\bf z})=h \left(\sum _{i=0}^{k-1} A^i {\bf c} {\bf z}_{-i}\right), \quad \mbox{for any} \quad {\bf z} \in K_M.
\end{equation}
\end{remark}

\subsection{State-affine systems and universality for fading memory RCs with linear readouts}
\label{State-affine systems and universality for fading memory RCs with linear readouts}

As it was explained in the introduction, the standard notion of reservoir computing that one finds in the literature concerns architectures with linear readouts. It is is particularly convenient to work with RCs that have this feature in machine learning applications since in that case the training reduces to solving a linear regression problem. That makes training feasible when there is need for a high number of neurons, as it happens in many cases. This point makes relevant the identification of families of reservoirs that are universal when the readout is restricted to be linear, which is the subject of this subsection. In order to simplify the presentation, we restrict ourselves in this case to one-dimensional input signals, that is, all along this  section we set $n=1 $. The extension to the general case is straightforward, even though more convoluted to write down (see Remark \ref{remark multidim}).

\begin{definition}
\label{state-affine definition}
Let $N \in \mathbb{N}  $, ${\bf W} \in \mathbb{R}^N $, and let $p(z)\in \mathbb{M}_{N}[z] $ and $q(z) \in \mathbb{M}_{N,1}[z] $ be two polynomials on the variable $z$  with matrix coefficients, as they were introduced in~\eqref{polynomial matrices}. The {\bfi  non-homogeneous state-affine system (SAS)} associated to $p, q $ and ${\bf W}  $ is the reservoir system determined by the state-space transformation:
\begin{empheq}[left={\empheqlbrace}]{align}
\mathbf{x} _t &=p(z _t)\mathbf{x}_{t-1}+q( {z} _t),\label{sas reservoir equation}\\
y _t &= {\bf W}^\top\mathbf{x} _t. \label{sas readout}
\end{empheq}
\end{definition}

Our next result spells out a sufficient condition that guarantees that the SAS reservoir system~\eqref{sas reservoir equation}-\eqref{sas readout}  has the echo state property. Moreover, it provides an explicit expression for the unique  causal and time-invariant solution associated to a given input. 

\begin{proposition}
\label{integration sas}
Consider a non-homogeneous state-affine system as in~\eqref{sas reservoir equation}-\eqref{sas readout} determined by polynomials $p, q $, and a vector ${\bf W}  $, with inputs defined on $I^{\Bbb Z} $, $I:=[-1,1] $. 
Assume that
\begin{equation}
\label{condition for sas int}
K _1:= \max _{z \in I}\| p (z)\| _2=\max _{z \in I}\sigma_{{\rm max}}(p (z)) <1.
\end{equation}
Then, the reservoir system~\eqref{sas reservoir equation}-\eqref{sas readout} has the echo state property and for each input ${\bf z}\in I ^{\mathbb{Z}}  $ it has a unique causal and time-invariant solution given by 
\begin{empheq}[left={\empheqlbrace}]{align}
\mathbf{x} _t &=\sum_{j=0}^ \infty \left(\prod_{k=0}^{j-1}p(z_{t-k}) \right)q(z _{t-j}),\label{sas integrated reservoir equation}\\
y _t &= {\bf W}^\top\mathbf{x} _t, \label{sas integrated readout}
\end{empheq}
where
\begin{equation*}
\prod_{k=0}^{j-1}p(z_{t-k}):=p(z_{t}) \cdot p(z_{t-1}) \cdots p(z_{t-j+1}).
\end{equation*}
Let now $K _2:= \max _{z \in I}\| q (z)\| _2 $. Then,
\begin{equation}
\label{ineq state sas}
\left\|\mathbf{x} _t\right\| \leq \frac{K _2}{1-K _1}, \quad \mbox{for all } \quad t \in \Bbb Z.
\end{equation}
We will denote by $U_{{\bf W}}^{p,q}:I^{\Bbb Z} \longrightarrow \mathbb{R}^{\Bbb Z}  $ and $H_{{\bf W}}^{p,q}:I^{\Bbb Z_-} \longrightarrow \mathbb{R}  $ the corresponding SAS reservoir filter and SAS functional, respectively.
\end{proposition}
\nomenclature{$H_{{\bf W}}^{p,q}:I^{\Bbb Z_-} \longrightarrow \mathbb{R}$}{SAS reservoir functional}
\nomenclature{$U_{{\bf W}}^{p,q}:I^{\Bbb Z} \longrightarrow \mathbb{R}^{\Bbb Z}  $}{SAS reservoir filter}

The next result presents two alternative conditions that imply the hypothesis $\max _{z \in I}\| p (z)\| _2 <1 $ in the previous proposition and that are easier to verify in practice.

\begin{lemma}
\label{conditions sas}
Let $p(z)\in \mathbb{M}_{N}[z] $ be the polynomial given by
\begin{equation*}
p(z):=A _0+zA _1+z ^2A _2+ \cdots + z ^{n_1} A _{n_1}, \quad {n_1} \in \mathbb{N}.
\end{equation*}
Suppose that $z \in I $ and consider the following three conditions:
\begin{description}
\item [(i)] There exists a constant $0< \lambda<1  $, such that $\|A _i\| _2= \sigma_{{\rm max}}(A _i) < \lambda $, for any $i \in \{0,1, \ldots, {n_1}\} $, and that at the same time satisfies that $\lambda({n_1}+1)<1$.
\item [(ii)] $B _p:=\|A _0\| _2+\|A _1\| _2+ \cdots + \|A _{n_1}\| _2 <1 $.
\item [(iii)]  $M _p:=\max _{z \in I}\| p (z)\| _2 <1 $.
\end{description}
Then, condition {\bf (i)} implies {\bf (ii)} and condition {\bf (ii)} implies {\bf (iii)}.
\end{lemma}

We emphasize that since Proposition~\ref{integration sas} was proved using condition {\bf (iii)} in the previous lemma then, any of the three conditions in that statement imply the echo state property for~\eqref{sas integrated reservoir equation}-\eqref{sas integrated readout} and the time-invariance of the corresponding solutions. The next result shows that the same situation holds in relation with the fading memory property.

\begin{proposition}
\label{fmp sas}
Consider a non-homogeneous state-affine system as in~\eqref{sas reservoir equation}-\eqref{sas readout} determined by polynomials $p, q $, and a vector ${\bf W}  $, with inputs defined on $I^{\Bbb Z} $, $I:=[-1,1] $. If the polynomial $p$ satisfies any of the three conditions  in Lemma~\ref{conditions sas} then the corresponding reservoir filter has the fading memory property. More specifically, if $p$ satisfies condition {\bf (i)} in Lemma~\ref{conditions sas}, then $H_{{\bf W}}^{p,q}:(I^{\Bbb Z_-}, \| \cdot \| _{w^\rho}) \longrightarrow \mathbb{R}  $ 
is continuous with $w^\rho_t:=(n _1+1) ^{ \rho t} \lambda ^{ \rho t} $ and $\rho \in (0,1) $ arbitrary. The same conclusion holds for conditions {\bf (ii)} and {\bf (iii)} with $w^\rho_t:=B _p ^{ \rho t} $ and $w^\rho_t:=M _p ^{ \rho t} $, respectively.
\end{proposition}

The importance of SAS in relation to the universality problem has to do with the fact that they form a polynomial algebra which allows us, under certain conditions, to use the Stone-Weierstrass theorem to prove a density statement. Before we show that, we observe that for any two polynomials $p_1(z)\in \mathbb{M}_{N_1,M _1}[z] $ and $p_2(z)\in \mathbb{M}_{N_2,M _2}[z] $ given by
\begin{eqnarray}
p_1(z)&:=A _0^1+zA _1^1+z ^2A _2^1+ \cdots + z ^{n_1} A _{n_1}^1,\\
p_2(z)&:=A _0^2+zA _1^2+z ^2A _2^2+ \cdots + z ^{n_2} A _{n_2}^2,
\end{eqnarray}
with $n _1, n _2 \in \mathbb{N} $,  their direct sum and their tensor product are also polynomials in $z$  with matrix coefficients. More explicitly, $p_1\oplus p_2(z)\in \mathbb{M}_{N_1+N _2,M _1+ M _2}[z]$ and is written as
\begin{equation}
\label{sum sas}
p_1\oplus p_2(z)=A _0^1\oplus A _0^2+zA _1^1\oplus A _1^2+z ^2A _2^1\oplus A _2^2+ \cdots +z ^{n_2} A _{n_2}^1\oplus A _{n_2}^2 + z ^{n_2+1} A _{n_2+1}^1\oplus {\bf 0}+ \cdots + z ^{n_1} A _{n_1}^1\oplus {\bf 0},
\end{equation}
where we assumed that $n _2\leq n _1$. Analogously, their tensor product $p_1\otimes p_2(z)\in \mathbb{M}_{N_1\cdot N _2,M _1\cdot  M _2}[z]  $ and is written as
\begin{equation}
\label{product sas}
p_1\otimes p_2(z)=\sum_{i=0}^{n _1}\sum_{j=0}^{n _2} z^{i+j}A _i ^1\otimes A _j ^2.
\end{equation}

The next result shows that the products and the linear combinations of  SAS reservoir functionals are SAS reservoir functionals. Additionally, it makes explicit the polynomials that determine the corresponding SAS reservoir systems.

\begin{proposition}
\label{SAS polynomial algebra} 
Let $H_{{\bf W}_1}^{p_1,q_1}, H_{{\bf W}_2}^{p_2,q_2}:I^{\Bbb Z_-} \longrightarrow \mathbb{R}  $ be two SAS reservoir functionals associated to two corresponding time-invariant SAS reservoir systems. Assume that the two polynomials with matrix coefficients $p_1(z)\in \mathbb{M}_{N_1}[z] $ and $p_2(z)\in \mathbb{M}_{N_2}[z] $ satisfy that $\|p _1 (z)\| _2<1- \epsilon $ and $\|p _2 (z)\| _2 <1- \epsilon$ for all $z \in I:=[-1,1] $ and a given $0<\epsilon<1 $. Then, with the notation introduced in the expressions~\eqref{sum sas} and~\eqref{product sas}, we have that:
\begin{description}
\item [(i)] For any $\lambda \in \mathbb{R} $, the linear combination of the SAS reservoir functionals $H_{{\bf W}_1}^{p_1,q_1}+ \lambda H_{{\bf W}_2}^{p_2,q_2} $ is a SAS reservoir functional and: 
\begin{equation}
\label{sum sas filters}
H_{{\bf W}_1}^{p_1,q_1}+ \lambda H_{{\bf W}_2}^{p_2,q_2}=H_{{\bf W}_1\oplus \lambda{\bf W}_2}^{p _1\oplus p_2,q _1\oplus q_2}.
\end{equation}
\item [(ii)] The product of the SAS reservoir functionals $H_{{\bf W}_1}^{p_1,q_1}\cdot  H_{{\bf W}_2}^{p_2,q_2} $ is a SAS reservoir functional and: 
\begin{equation}
\label{product sas filters}
H_{{\bf W}_1}^{p_1,q_1}\cdot  H_{{\bf W}_2}^{p_2,q_2}=H_{{\bf 0}\oplus {\bf 0}\oplus \left({\bf W}_1\otimes {\bf W}_2\right)}^{p,q _1\oplus q_2\oplus \left(q _1\otimes q_2\right)},
\end{equation}
where $p(z)\in \mathbb{M}_{N_{12}} [z]$, $N_{12}:=N _1 +N _2+N _1 \cdot N _2 $, is the polynomial with matrix coefficients in $\mathbb{M}_{N_{12}} $ whose block-matrix expression for the three summands in $\mathbb{R}^{N _1}\oplus \mathbb{R}^{N _2} \oplus \left(\mathbb{R}^{N _1}\otimes \mathbb{R}^{N _2}\right)$ is:
\begin{equation}
\label{matrix for sas}
p(z):=
\left(
\begin{array}{ccc}
p _1(z)& {\bf 0} &{\bf 0}\\
{\bf 0}&p _2 (z)& {\bf 0}\\
p _1\otimes q _2(z) & q _1\otimes p _2 (z)&p _1\otimes p _2(z)
\end{array}
\right).
\end{equation}
The expression $p_1\otimes p_2(z)\in \mathbb{M}_{N_1\cdot N _2}[z]$    denotes the element defined in  \eqref{product sas}. The symbol $p _1\otimes q _2 (z)$ (respectively, $ q _1\otimes p _2(z)$) denotes the matrix of the linear map from $\mathbb{R}^{N _1} $ (respectively, $\mathbb{R}^{N _2} $) to $\mathbb{R}^{N _1}\otimes \mathbb{R}^{N _2} $ that associates to any  ${\bf v} _1\in \mathbb{R}^{N _1} $ the element $(p _1 (z)  {\bf v} _1)\otimes q _2(z)$ (respectively, $q _1 (z) \otimes (p _2(z)  {\bf v} _2) $, with ${\bf v} _2\in \mathbb{R}^{N _2} $). When all the polynomials in \eqref{matrix for sas} are written in terms of monomials using the conventions that we just mentioned and we factor out the different powers of the variable $z$, we obtain a polynomial with matrix coefficients in $\mathbb{M}_{N_{12}} $  and with degree ${\rm deg}(p) $ equal to 
$${\rm deg}(p)=\max \left\{{\rm deg}(p _1) \cdot {\rm deg}(q _2), {\rm deg}(q _1) \cdot {\rm deg}(p _2), {\rm deg}(p _1) \cdot {\rm deg}(p _2)\right\}.$$
\end{description}
The equalities~\eqref{sum sas filters} and~\eqref{product sas filters} show that the SAS family forms a polynomial algebra.
\end{proposition}

\begin{remark}
\normalfont
Notice that the linear reservoir equation \eqref{linear reservoir equation} is a particular case of the SAS reservoir equation \eqref{sas reservoir equation} that is obtained by taking for $p$  and $q$ polynomials of degree zero and one, respectively. Regarding that specific case, Proposition \ref{SAS polynomial algebra} shows that linear reservoirs with linear readouts do not form a polynomial algebra. Indeed, as it can be seen in \eqref{product sas filters}, the product of two SAS filters involves the tensor product $q _1\otimes q _2$ which, when $q _1  $ and $ q _2 $ come from a linear filter, it has degree two and it is hence not compatible with a linear reservoir filter. 
\end{remark}

\begin{theorem}[Universality of SAS reservoir computers]
\label{universality of sas systems}
Let $I ^{\Bbb Z _-} \subset \mathbb{R} ^{\Bbb Z _-} $ be the subset of real uniformly bounded  sequences in $I:=[-1, 1]$ as in ~\eqref{Kset}, that is, 
\begin{equation*}
I ^{\Bbb Z _-} :=\{{\bf z}\in \mathbb{R} ^{\Bbb Z _-}\mid z _t \in  [-1,1], \  \mbox{for all} \quad  t\leq 0 \},
\end{equation*}
and let 
$\mathcal{S}_\epsilon$ be the family of functionals $H_{{\bf W}}^{p,q}:I ^{\Bbb Z _-}  \longrightarrow \mathbb{R}  $ induced by the state-affine systems defined in~\eqref{sas reservoir equation}-\eqref{sas readout} that satisfy that $M _p:=\max _{z \in I}\| p (z)\| _2 <1 - \epsilon$ and $M _q:=\max _{z \in I}\| q (z)\| _2 <1 - \epsilon$. The family $\mathcal{S}_\epsilon$  forms a polynomial subalgebra of $\mathcal{R} _{w ^\rho}$ (as defined in~\eqref{generating family reservoirs all}) with $w ^\rho _t:=(1- \epsilon )^{ \rho t} $ and $\rho \in (0,1) $ arbitrary, made of fading memory reservoir filters that contains the constant functions and separates points.  The subfamily $\mathcal{S} _\epsilon $ is hence dense in the set $(C ^0(I ^{\Bbb Z _-} ),\|\cdot \| _{w ^\rho})$ of real-valued continuous functions on $(I ^{\Bbb Z _-} , \| \cdot \| _{w ^\rho})$.

This statement implies that any causal, time-invariant fading memory filter $H:I ^{\Bbb Z _-}  \longrightarrow \mathbb{R} $ can be uniformly approximated  by elements in $\mathcal{S}_\epsilon$.
More specifically, for any fading memory filter $H$ and any $\epsilon>0 $, there exist a natural number $N \in \mathbb{N} $,  polynomials $p (z) \in \mathbb{M}_{N}[z], q (z) \in \mathbb{M}_{N,1}[z]$ with $M _p, M _q< 1- \epsilon $, and a vector $ \mathbf{W} \in \mathbb{R}^N $ such that  
\begin{equation*}
\| H - H _{\bf W} ^{p,q}\| _{\infty} := \sup_{{z} \in I ^{\Bbb Z _-} } \{|H ({z})-H _{\bf W} ^{p,q} ({z}) |\}< \epsilon.
\end{equation*}

\nomenclature{$\mathcal{S} _\epsilon$}{State affine reservoir systems (SAS) $H_{{\bf W}}^{p,q}:I ^{\Bbb Z _-}  \longrightarrow \mathbb{R}  $  with $M _p<1 - \epsilon$ and $M _q <1 - \epsilon$}
\nomenclature{$\mathcal{NS} _\epsilon$}{Subfamily of $\mathcal{S}_\epsilon$ formed by  SAS reservoir systems determined by nilpotent polynomials $p$}

The same universality result can be stated for the smaller subfamily $\mathcal{NS}_\epsilon\subset \mathcal{S}_\epsilon$ formed by SAS reservoir systems determined by nilpotent polynomials $p(z) \in\mathbb{N}{\rm il} [z]$.
\end{theorem}

\begin{remark}
\normalfont
As it is stated in Theorem~\ref{universality of sas systems}, it is the condition {\bf (iii)} in Lemma~\ref{conditions sas} that yields a universal family of SAS fading memory reservoirs. As it can deduced from its proof (available in the Appendix~\ref{Proof of universality of sas systems}), the families determined by conditions {\bf (i)} or {\bf (ii)} in that lemma contain  SAS fading memory reservoirs but they do not form a polynomial algebra. In such cases, and according to Theorem~\ref{universality deterministic general}, it is the algebras generated by them and not the families themselves that are universal.
\end{remark}

\begin{remark}
\normalfont
A continuous-time analog of the universality result that we just proved can be obtained using the bilinear systems considered in Section 5.3 of~\cite{Boyd1985}. In discrete time, but only when the number of time steps is finite, this universal approximation property is exhibited~\cite{FliessNormand1980} by homogeneous state-affine systems, that is, by setting $q(z)={\bf 0} $ in~\eqref{sas reservoir equation}-\eqref{sas readout}.
\end{remark}

\begin{remark}
\label{remark multidim}
\normalfont
{\bf Generalization to multidimensional signals.} When the input signal is defined in $I _n ^{\Bbb Z} $, with $I _n:=[-1,1] ^n  $, a SAS family with the same universality properties can be defined by replacing the polynomials $p $ and $q$ in Definition \ref{state-affine definition}, by polynomials of degree $r$ and $s$ of the form:
\begin{eqnarray*}
p({\bf z})&=&\sum_{{i _1, \ldots, i _n \in \left\{0, \ldots, r\right\} \above 0 pt i _1+ \cdots + i _n\leq r}}
z _1^{i _1} \cdots z _n^{i _n} A_{{i _1, \ldots, i _n}}, \quad A_{{i _1, \ldots, i _n}} \in \mathbb{M}_{N}, \quad {\bf z} \in I _n\\
q({\bf z})&=&\sum_{{i _1, \ldots, i _n \in \left\{0, \ldots, s\right\} \above 0 pt i _1+ \cdots + i _n\leq s}}
z _1^{i _1} \cdots z _n^{i _n} B_{{i _1, \ldots, i _n}}, \quad B_{{i _1, \ldots, i _n}} \in \mathbb{M}_{N,1}, \quad {\bf z} \in I _n.
\end{eqnarray*}
\end{remark}

\begin{remark}
\normalfont
{\bf SAS with trigonometric polynomials.} An analogous construction can be carried out using trigonometric polynomials of the type:
\begin{eqnarray*}
p({\bf z})&=&\sum_{{i _1, \ldots, i _n \in \left\{0, \ldots, r\right\} \above 0 pt i _1+ \cdots + i _n\leq r}}
\cos \left( i _1\cdot z _1 +\cdots + i _n \cdot z _n \right)A_{{i _1, \ldots, i _n}}, \quad A_{{i _1, \ldots, i _n}} \in \mathbb{M}_{N}, \quad {\bf z} \in I _n\\
q({\bf z})&=&\sum_{{i _1, \ldots, i _n \in \left\{0, \ldots, s\right\} \above 0 pt i _1+ \cdots + i _n\leq s}}
\cos \left( i _1\cdot z _1 +\cdots + i _n \cdot z _n \right) B_{{i _1, \ldots, i _n}}, \quad B_{{i _1, \ldots, i _n}} \in \mathbb{M}_{N,1},\quad {\bf z} \in I _n.
\end{eqnarray*}
In this case, it is easy to establish that the resulting SAS family forms a polynomial algebra by invoking Proposition \ref{SAS polynomial algebra} and by reformulating the expressions \eqref{sum sas} and \eqref{product  sas} using the trigonometric identity
\begin{equation*}
\cos(\theta)\cos(\phi)= \frac{1}{2}\left(\cos(\theta- \phi)+\cos(\theta+ \phi)\right).
\end{equation*}
Additionally, the corresponding SAS family includes the linear family and hence the point separation property can be established as in the proof of Theorem \ref{universality of sas systems} in the Appendix \ref{Proof of universality of sas systems}.
\end{remark}

\section{Reservoir universality results in the stochastic setup}
\label{Reservoir universality results in the stochastic setup}

This section extends the previously stated deterministic universality results to a setup in which the reservoir inputs and outputs are stochastic, that is, the reservoir is not driven anymore by infinite sequences but by discrete-time stochastic processes. We emphasize that we restrict our discussion to reservoirs that are deterministic and hence the only source of randomness in the systems considered is the stochastic nature of the input. 

The results that follow are   mainly based on the observation that if we adopt a uniform approximation criterion and we assume that the random inputs satisfy almost surely the uniform boundedness that we used as hypothesis in Section \ref{Universality results in the deterministic setup}, then important features like the fading memory property or universality are naturally inherited in the stochastic setup from the deterministic  case. This fact is what we call the {\bfi  deterministic-stochastic transfer principle} and it is contained in the statement of Theorem \ref{fmp is inherited} below. In particular, this result can be easily applied to show that all the universal families with deterministic inputs introduced in Section \ref{Universality results in the deterministic setup} are also universal in the stochastic setup when the input processes considered produce paths that, up to a set of measure zero, are uniformly bounded.

\paragraph{The stochastic setup.}
All along this section we work on a probability space $(\Omega , \mathcal{A}, \mathbb{P}) $. If a condition defined on this probability space holds everywhere except for  a set with zero measure, we will say that the relation is true {\bfi almost surely}.  Let ${\bf X}: \Omega \longrightarrow B$ be a random variable with $(B, \left\|\cdot \right\|_B)$ a normed space endowed with a $\sigma$-algebra (for example, but not necessarily,  its Borel $\sigma$-algebra).  Let
%\begin{equation}
%\label{definition Linfi space}
%\left\|{\bf X}\right\|_{L^{\infty}}:= \mathop{{\rm ess\, sup}}_{\omega \in \Omega}  \left\|{\bf X} (\omega)\right\|= \inf \left\{\rho\in \mathbb{R}_+\mid \left\|{\bf X}\right\|< \rho \quad \mbox{almost surely}\right\}.
%\end{equation}

\begin{equation}
\label{definition Linfi space}
\left\|{\bf X}\right\|_{L^{\infty}}:= \mathop{{\rm ess\, sup}}_{\omega \in \Omega}\{  \left\|{\bf X} (\omega)\right\|_B\} = \inf \left\{\rho\in \overline{\mathbb{R}^+}\mid \left\|{\bf X}\right\|_B\le \rho \quad \mbox{almost surely}\right\},
\end{equation}
We denote by  $L ^{\infty}(\Omega, B) $ the classes of $B$-valued almost surely equal random variables whose norms have a finite essential supremum or that, equivalently, have almost surely bounded  norms, that is,
\begin{equation}
\label{def Linfi}
L ^{\infty}(\Omega, B):= S_B/ \sim_B, 
\end{equation}
where
\begin{equation}
S_B:= \left\{{\bf X}: \Omega \longrightarrow B \ \mbox{ random variable}\mid \left\|{\bf X}\right\|_{L^{\infty}}< \infty\right\},
\end{equation}
and $\sim_B$ is the equivalence relation defined on $S_B$ as follows: two random variables ${\bf Y} $ and $  {\bf Z}$ with finite $\left\|\cdot \right\|_{L^{\infty}} $ norm are $\sim_B$-equivalent   if and only if $\mathbb{P}( \left\{\omega \in \Omega: {\bf Y} ( \omega)\neq   {\bf Z}( \omega) \right\} = 0 $. As it is customary in the literature, we will not make a distinction in what follows between the elements in $S_B  $  and the classes in the quotient $L ^{\infty}(\Omega, B) $. Using this identification we recall, for example, that  $L ^{\infty}(\Omega, B) $ is  a vector space with the operations 
\begin{equation}
\label{lincomb}
({\bf X} + \lambda {\bf Y})( \omega) := {\bf X}( \omega) + \lambda {\bf Y}( \omega)
\end{equation}  for any ${\bf X}, {\bf Y} \in L ^{\infty}(\Omega, B)$,  $\lambda\in \mathbb{R}$, $\omega\in \Omega$. Moreover, $(L ^{\infty}(\Omega, B), \left\|\cdot \right\|_{L^{\infty}}) $ is  a normed space.
We emphasize that  $L ^{\infty}(\Omega, B) $ is in general not a Banach space (see~\cite[pages 42 and 46]{ledoux:talagrand}. It can be shown that whenever $B$ is finite dimensional or, more generally, a separable Banach space, then the space $L ^{\infty}(\Omega, B) $  is also a Banach space \cite{Pisier2016}.% The separability assumption for the Banach space $B$ is very convenient in the context of the so called Radon (tight or regular) random variables and their sequences. In particular, in this case the Borel and cylindrical $\sigma$-algebras coincide on $B$ and one does not need to worry about treatment of measurability questions.

Given an element ${\bf X} \in L ^{\infty}(\Omega, B) $, we denote by ${\rm E}\left[{\bf X}\right] $ its expectation. The following lemma collects some elementary results that will be needed later on and shows, in particular, that the expectation ${\rm E}\left[{\bf X}\right] $ as well as that of all the powers $\left\|{\bf X}\right\| _B ^k$ of its norm are finite for all the elements ${\bf X} \in L ^{\infty}(\Omega, B) $.

\begin{lemma}
\label{properties linfinity}
Let ${\bf X} \in L ^{\infty}(\Omega,B) $ and let $C \in \overline{\mathbb{R}^+}$. Then:
\begin{description}
\item [(i)] $\left\| {\bf X}\right\|_B \leq \left\| {\bf X}\right\| _{L^{\infty}} $ almost surely.
\item [(ii)] $ \left\| {\bf X}\right\| _{L^{\infty}} \leq C$ if and only if $\left\| {\bf X}\right\|_B\leq C $ almost surely.
\item [(iii)] $\left\| {\bf X}\right\|_B\leq C $ almost surely if and only if ${\rm E}\left[\|{\bf X}\|_B ^k\right] \leq C ^k$ for any $k \in \mathbb{N} $. 
\item [(iv)] Let $B = \mathbb{R}^n$, then the components $X _i $ of $ {\bf X} $, $i \in \left\{1, \ldots, n\right\} $, are such that ${\rm E}\left[X _i\right]\leq \left\| {\bf X}\right\| _{L^{\infty}} $.
\end{description}
\end{lemma}
The first point in this lemma explains why we will refer to the elements of $L^{\infty}(\Omega,B) $ as {\bfi  almost surely bounded} random variables.

\medskip

\noindent {\bf Stochastic inputs and outputs.} The  filters that we will consider in this section have {\bfi almost surely bounded  stochastic processes} as inputs and outputs. Recall that a discrete-time stochastic process is a map of the type: 
\begin{equation}
\label{stochastic process 1}
\begin{array}{cccc}
{\bf z}: &\Bbb Z\times \Omega & \longrightarrow &\mathbb{R}^n\\
	&(t, \omega)&\longmapsto &{\bf z}_t(\omega),
\end{array}
\end{equation}
such that,  for each $t \in \Bbb Z $, the assignment ${\bf z} _t: \Omega \longrightarrow {\Bbb R}^n  $ is a random variable.
For each $\omega \in  \Omega  $, we will denote by ${\bf z} (\omega):=\{ {\bf z} _t(\omega ) \in \mathbb{R}^n\mid  t \in \Bbb Z  \}$  the {\bfi realization} or the {\bfi sample path} of the process ${\bf z}$. The results that follow are presented for stochastic processes indexed by $\Bbb Z  $ but are equally valid for  $\mathbb{Z}_{+} $ and $\mathbb{Z}_{-} $.

Recall that a  map of the type \eqref{stochastic process 1} is a $\mathbb{R}^n $-valued stochastic process if and only if the corresponding map ${\bf z}: \Omega \longrightarrow \left(\mathbb{R}^n\right)^{\mathbb{Z}}$ into path space (designated with the same symbol) is a random variable when in $\left(\mathbb{R}^n\right)^{\mathbb{Z}}$ we consider the product sigma algebra generated by cylinder sets \cite[Chapter 1]{Comets:Meyre}. Then, the  space of $\mathbb{R}^n$-valued stochastic processes can be made into a vector space with the same operations as  in \eqref{lincomb} and we can define in this space a norm $\| \cdot \| _{L^{\infty}} $ using the same prescription as in \eqref{definition Linfi space} by considering $\left(\mathbb{R}^n\right)^{\mathbb{Z}}  $ as a normed space with the supremum norm $\| \cdot \| _{\infty} $, that is,
\begin{equation}
\label{definition Linfi space processes}
\left\|{\bf z}\right\|_{L^{\infty}}:= 
\mathop{{\rm ess\, sup}}_{\omega \in \Omega}  
\{\left\|{\bf z} (\omega)\right\|_{\infty}\}= 
\mathop{{\rm ess\, sup}}_{\omega \in \Omega}  
\left\{
\mathop{{\rm sup}}_{ t \in \Bbb Z} \left\{\| {\bf z} _t(\omega)
\|\right\}\right\}.
\end{equation}
The following lemma provides an alternative characterization of the norm $\| \cdot \| _{L^{\infty}} $ that will be very useful in the proofs of the results that follow and in which the supremum and the essential supremum have been interchanged. The last statement contained in it complements part {\bf (ii)} of Lemma \ref{properties linfinity} for processes.
\begin{lemma}
\label{swap sups}
Let ${\bf z}: \Omega \longrightarrow \left({\Bbb R}^n\right) ^{\mathbb{Z}} $ be a stochastic process. Then,
\begin{equation}
\label{equality swap sups}
\left\|{\bf z}\right\|_{L^{\infty}}:=  
\mathop{{\rm ess\, sup}}_{\omega \in \Omega}  
\left\{
\mathop{{\rm sup}}_{ t \in \Bbb Z} \left\{\| {\bf z} _t(\omega)
\|\right\}\right\}=
\mathop{{\rm sup}}_{ t \in \Bbb Z}  
\left\{
\mathop{{\rm ess\, sup}}_{\omega \in \Omega} \left\{\| {\bf z} _t(\omega)
\|\right\}\right\}.
\end{equation}
Equivalently, using the notation in \eqref{definition Linfi space},
\begin{equation}
\label{equality swap sups 2}
\left\|{\bf z}\right\|_{L^{\infty}}:=  \left\|
\sup_{ t \in \Bbb Z}\{\| {\bf z} _t(\omega)
\|\}\right\| _{L^{\infty}}=
\sup_{ t \in \Bbb Z} 
\{\left\|
{\bf z} _t
\right\| _{L^{\infty}}\}.
\end{equation}
These equalities imply that for any non-negative real number $C\ge 0 $, the process ${\bf z} $ satisfies that $\| {\bf z}\| _{L^{\infty}}\leq C $ if and only if $\| {\bf z}_t\| _{L^{\infty}}\leq C $ for all $t \in \Bbb Z  $ or, equivalently, if and only if $\sup_{t \in \Bbb Z }\{\| {\bf z}_t\| _{L^{\infty}}\}\leq C $.
\end{lemma}

{Consider now the  space $L ^{\infty}\left(\Omega, (\mathbb{R}^n) ^{\mathbb{Z}}\right) $
\nomenclature{$L ^{\infty}\left(\Omega, (\mathbb{R}^n) ^{\mathbb{Z}} \right) $}{Space of almost surely bounded time series or discrete-time stochastic processes with values in ${\Bbb R}^n$} 
of processes with finite $\left\|\cdot \right\|_{L^{\infty}} $ norm.  We refer to the elements of $L ^{\infty}\left(\Omega, (\mathbb{R}^n) ^{\mathbb{Z}}\right) $ as {\bfi  almost surely bounded time series}. Additionally,   consider the space $L ^{\infty}\left(\Omega, \ell ^{\infty}({\Bbb R}^n) \right)$ of processes whose paths are all uniformly bounded, that is, they lay in the Banach space $(\ell ^{\infty}({\Bbb R}^n),\| \cdot \| _{\infty})$. According to the definition in \eqref{def Linfi}, we have for both these spaces that  
\begin{equation*}
L ^{\infty}\left(\Omega, (\mathbb{R}^n) ^{\mathbb{Z}}\right):= S_{(\mathbb{R}^n) ^{\mathbb{Z}} }/\sim_{(\mathbb{R}^n) ^{\mathbb{Z}}}, \quad L ^{\infty}\left(\Omega, \ell ^{\infty}({\Bbb R}^n)\right):= S_{ \ell ^{\infty}({\Bbb R}^n)}/\sim_{ \ell ^{\infty}({\Bbb R}^n)}
\end{equation*}
with
\begin{equation*}
S_{(\mathbb{R}^n) ^{\mathbb{Z}}} := \left\{{\bf z}: \Bbb Z \times \Omega \longrightarrow \mathbb{R}^n \ \mbox{ stochastic process, }  {\bf z}( \omega) \in (\mathbb{R}^n) ^{\mathbb{Z}}, \  \mbox{for all} \ \omega\in \Omega \mid \left\|{\bf z}\right\|_{L^{\infty}}< \infty\right\},
\end{equation*} 
\begin{equation*}
S_{\ell ^{\infty}({\Bbb R}^n)} := \left\{{\bf z}: \Bbb Z \times \Omega \longrightarrow \mathbb{R}^n \ \mbox{ stochastic process, } {\bf z}( \omega) \in \ell ^{\infty}({\Bbb R}^n),  \  \mbox{for all} \ \omega\in \Omega \mid \left\|{\bf z}\right\|_{L^{{\infty}}}< \infty\right\},
\end{equation*}
and with the almost sure equality equivalence relations  $\sim_{\ell ^{\infty}({\Bbb R}^n)}$ and $\sim_{(\mathbb{R}^n) ^{\mathbb{Z}}}$ between stochastic processes with paths in $\ell ^{\infty}({\Bbb R}^n)$ and $(\mathbb{R}^n) ^{\mathbb{Z}}$, respectively. The following result shows that  the normed spaces $L ^{\infty}\left(\Omega, (\mathbb{R}^n) ^{\mathbb{Z}}\right)$ and  $L ^{\infty}\left(\Omega, \ell ^{\infty}({\Bbb R}^n)\right)$  are  isomorphic.

\begin{lemma}
\label{lemma linfi}
In the setup that we just introduced
 the inclusion $\iota  : S_{\ell ^{\infty}({\Bbb R}^n)}\hookrightarrow S_{(\mathbb{R}^n) ^{\mathbb{Z}}} $ is equivariant with respect to the equivalence relations $\sim_{\ell ^{\infty}({\Bbb R}^n)}$ and $\sim_{(\mathbb{R}^n) ^{\mathbb{Z}}}$ and drops to an isomorphism of normed spaces $\phi:(L ^{\infty}\left(\Omega, (\mathbb{R}^n) ^{\mathbb{Z}}\right),  \left\|\cdot \right\|_{L^{\infty}})\longrightarrow (L ^{\infty}\left(\Omega, \ell ^{\infty}({\Bbb R}^n)\right),  \left\|\cdot \right\|_{L^{\infty}})$. Equivalently, the following diagram commutes
 \begin{diagram}
S_{\ell ^{\infty}({\Bbb R}^n)} &\rInto^{\enspace \iota} &S_{ ({\Bbb R}^n)^{\mathbb{Z}}}\\
\dTo^{\Pi_{\sim_{\ell ^{\infty}({\Bbb R}^n)}}} & &\dTo_{\Pi_{\sim_{({\Bbb R}^n)^{\mathbb{Z}}}}}\\
L ^{\infty}\left(\Omega, \ell ^{\infty}({\Bbb R}^n)\right) &\rTo^{\phi} &L ^{\infty}\left(\Omega, ({\Bbb R}^n)^{\mathbb{Z}}\right),
\end{diagram}
where $\Pi_{\sim_{\ell ^{\infty}({\Bbb R}^n)}} $ and $\Pi_{\sim_{({\Bbb R}^n)^{\mathbb{Z}}}}$ are the canonical projections.
%\begin{equation}
%\label{equality banach}
%L ^{\infty}\left(\Omega, (\mathbb{R}^n) ^{\mathbb{Z}}\right)=
%L ^{\infty}\left(\Omega, \ell ^{\infty}({\Bbb R}^n)\right)
%\end{equation}
%and  $\left(L ^{\infty}\left(\Omega, (\mathbb{R}^n) ^{\mathbb{Z}}\right), \| \cdot \| _{L^{\infty}}\right) $ is a Banach space.
\end{lemma}
}
Let now $w$ be a weighting sequence and let $\| \cdot \| _w $ be the associated weighted norm in $\left(\mathbb{R}^n\right)^{\mathbb{Z}_{-}} $. If we replace in \eqref{definition Linfi space processes} the $\ell ^{\infty}  $ norm $\| \cdot \| _{\infty} $  by the weighted norm $\| \cdot \| _w $, we obtain a weighted norm $\| \cdot \| _{L^{\infty}_w}$ in the space of processes ${\bf z}: \mathbb{Z}_{-} \times \Omega \longrightarrow\mathbb{R}^n $ indexed by $\Bbb Z_-$ as:
\begin{equation}
\label{definition Linfiw space processes}
\left\|{\bf z}\right\|_{L^{\infty}_w}:= 
\mathop{{\rm ess\, sup}}_{\omega \in \Omega}  
\{\left\|{\bf z} (\omega)\right\|_{w}\}= 
\mathop{{\rm ess\, sup}}_{\omega \in \Omega}  
\left\{
\mathop{{\rm sup}}_{ t \in \Bbb Z_{-}} \left\{\| {\bf z} _t(\omega)
\|w_{-t}\right\}\right\}.
\end{equation}
We will denote by $L^{\infty}_{w} \left(\Omega, (\mathbb{R}^n)^{\mathbb{Z}_{-}}\right) $ the space of processes with finite $\left\|\cdot \right\|_{L^{\infty}_w}$ norm. 
\nomenclature{$L^{\infty}_{w} \left(\Omega, (\mathbb{R}^n)^{\mathbb{Z}_{-}}\right)$}{Space of time series or discrete-time stochastic processes with values in ${\Bbb R}^n$ with finite $L^{\infty}_w$-norm}
A result similar to Lemma \ref{lemma linfi} shows that the normed spaces  $(L ^{\infty}_w\left(\Omega, (\mathbb{R}^n) ^{\mathbb{Z}_{-}}\right), \left\|\cdot \right\|_{L^{\infty}_w})$ and $(L ^{\infty}\left(\Omega, \ell ^{\infty}_w({\Bbb R}^n)\right), \left\|\cdot \right\|_{L^{\infty}_w}) $ are isomorphic. Additionally, as in Lemma \ref{swap sups}, we have that for any ${\bf z}\in L ^{\infty}_w\left(\Omega, (\mathbb{R}^n) ^{\mathbb{Z}_{-}}\right) $:
\begin{equation}
\label{definition Linfiw space processes swap}
\left\|{\bf z}\right\|_{L^{\infty}_w}:=  
\mathop{{\rm ess\, sup}}_{\omega \in \Omega}  
\left\{
\mathop{{\rm sup}}_{ t \in \Bbb Z_{-}} \left\{\| {\bf z} _t(\omega)
\|w_{-t}\right\}\right\}=
\mathop{{\rm sup}}_{ t \in \Bbb Z_{-}}  
\left\{
\mathop{{\rm ess\, sup}}_{\omega \in \Omega}
 \left\{\| {\bf z} _t(\omega)
\|w_{-t}\right\}\right\}.
\end{equation}

\noindent {\bf Deterministic filters in a stochastic setup.} As we already pointed out, we consider filters $U$ that have  almost surely bounded processes as  inputs and outputs. The same conventions as in the deterministic setup  are used in the identification of the different signals, namely, ${\bf z} $  denotes the filter input process and the symbol $y $ is reserved for the output process.
Let now $D _n \subset {\Bbb R}^n $ and let $D_n^{L^{\infty}_{\Bbb Z}} \subset L^{\infty}(\Omega, (\mathbb{R}^n) ^{\mathbb{Z}}) $ be a subset  formed by processes whose paths take values in $D_n$ almost surely. In the sequel we will restrict our attention to intrinsically {\bfi  deterministic filters} $U:D_n^{L^{\infty}_{\Bbb Z}}  \longrightarrow L^{\infty}(\Omega, \mathbb{R} ^\mathbb{Z})$
that are obtained by presenting almost surely bounded stochastic inputs ${\bf z}  \in D_n^{L^{\infty}_{\Bbb Z}} \subset L^{\infty}(\Omega, (\mathbb{R}^n) ^{\mathbb{Z}})$ to filters $U: \left(D_n\right)^{\mathbb{Z}} \longrightarrow  \mathbb{R}^{\mathbb{Z}}$ similar to those introduced in the previous section, which explains why we use the same symbol for both. This is explicitly carried out by defining the output process $U({\bf z}) \in L^{\infty}(\Omega, \mathbb{R} ^\mathbb{Z})$ using the convention
\begin{equation}
\label{convention U}
(U({\bf z}))( \omega):=U({\bf z}( \omega)), \quad \omega\in \Omega,
\end{equation}
where on the right hand side it is the filter $U: \left(D_n\right)^{\mathbb{Z}} \longrightarrow  \mathbb{R}^{\mathbb{Z}}$  which is applied to the paths ${\bf z} (\omega):=\{ {\bf z} _t(\omega ) \in \mathbb{R}^n\mid  t \in \Bbb Z  \} \in  \left(D_n\right)^{\mathbb{Z}}$ of the process ${\bf z}$. We call these filters deterministic because, in view of \eqref{convention U} the dependence  of the image process $(U({\bf z}))(  {\omega}) \in \in L^{\infty}(\Omega, \mathbb{R} ^\mathbb{Z}) $ on the probability space  takes place exclusively through the dependence ${\bf z}( {\omega})$ in the input. In this section we reserve the symbol $U$ to denote deterministic filters $U:D_n^{L^{\infty}_{\Bbb Z}}  \longrightarrow L^{\infty}(\Omega, \mathbb{R} ^\mathbb{Z})$. 
We draw attention to the fact that assuming that the filters map into almost surely bounded processes is a genuine hypothesis that needs to be verified in each specific case considered.

The concepts of causality and time-invariance are defined as in the deterministic case by replacing equalities by almost sure equalities in the corresponding identities. More explicitly, we say that the filter $U:D_n^{L^{\infty}_{\Bbb Z}}  \longrightarrow L^{\infty}(\Omega, \mathbb{R} ^{\mathbb{Z}})$ is time-invariant when for any $\tau\in \Bbb Z $ and any $ {\bf z} \in D_n^{L^{\infty}_{\Bbb Z}}$, we have that 
\begin{equation*}
(U _\tau \circ U)({\bf z})=(U  \circ U_\tau) ({\bf z}), \quad \mbox{almost surely.}
\end{equation*}
Analogously, we say that the filter is causal with stochastic inputs when for any two elements ${\bf z} , \mathbf{w} \in D_n^{L^{\infty}_{\Bbb Z}} $  that satisfy that ${\bf z} _\tau = \mathbf{w} _\tau$ almost surely, for any $\tau \leq t  $ and for a given  $t \in \Bbb Z $, we have that $U( {\bf z}) _t= U ({\bf w}) _t $, almost surely. Causal and time-invariant deterministic filters produce almost surely causal and time-invariant filters when stochastic inputs are presented to them.

In this setup, there is also a correspondence between causal and time-invariant filters $U:D_n^{L^{\infty}_{\Bbb Z}} \longrightarrow L^{\infty}(\Omega, \mathbb{R} ^{\mathbb{Z}})$ and functionals $H _U:D_n^{L^{\infty}_{\Bbb Z_-}} \longrightarrow L^{\infty}(\Omega, \mathbb{R}) $, where $D_n^{L^{\infty}_{\Bbb Z_-}}:= \mathbb{P}_{\mathbb{Z}_{-}} \left(D_n^{L^{\infty}_{\Bbb Z}}\right) $. 

Given a weighting sequence $w : \mathbb{N} \longrightarrow (0,1] $ and a time-invariant filter $U:D_n^{L^{\infty}_{\Bbb Z_-}}\longrightarrow L^{\infty}(\Omega, \mathbb{R} ^{\mathbb{Z}})$ with stochastic inputs, we will say that $U$ has the {\bfi  fading memory property} with respect to the weighting sequence $w $ when the corresponding functional $H _U:\left(D_n^{L^{\infty}_{\Bbb Z_-}}, \| \cdot \| _{L^{\infty}_w}\right) \longrightarrow L^{\infty}(\Omega, \mathbb{R}) $ is a continuous map. 

Let $M >0 $ and define, using Lemma \ref{swap sups},
\begin{equation}
\label{Kset stochastic}
K^{L^{\infty}}_{M}:=\left\{ {\bf z} \in L^{\infty}(\Omega, (\mathbb{R}^n) ^{\mathbb{Z}_-}) \mid \| {\bf z}  \|_{L^{\infty}} \leq M\right\}=\left\{ {\bf z} \in L^{\infty}(\Omega, (\mathbb{R}^n) ^{\mathbb{Z}_-}) \mid \| {\bf z} _t \|_{L^{\infty}} \leq M, \  \mbox{for all} \   t \in \Bbb Z _-\right\}.
\end{equation}
The sets $K^{L^{\infty}}_{M} $ are the stochastic counterparts of the sets $K _M  $ in the deterministic setup; we will say that $K^{L^{\infty}}_{M} $ is a set of {\bfi almost surely uniformly bounded processes}. A stochastic analog of Lemma ~\ref{uniformly bounded} can be formulated for them with $K _M $ replaced by $K^{L^{\infty}}_{M} $, the norm $\left\|\cdot \right\| $ by $\left\|\cdot \right\|_{L^{\infty}} $, and the weighted norm $\left\|\cdot \right\|_w $ by $\left\|\cdot \right\|_{L^{\infty}_w} $. Indeed, the following result shows that the fading memory  and the universality properties are naturally inherited by deterministic filters with almost surely uniformly bounded inputs. We call this fact the {\bfi  deterministic-stochastic transfer principle}.

\begin{theorem}[Deterministic-stochastic transfer principle]
\label{fmp is inherited}
Let $M > 0  $ and let $K _M $ and $K _M ^{L^{\infty}} $ be the sets of deterministic and stochastic inputs defined in ~(\ref{Kset}) and ~(\ref{Kset stochastic}), respectively. The following properties hold true:
\begin{description}
\item [(i)] Let $H: (K _M, \left\|\cdot \right\| _w) \longrightarrow \mathbb{R}  $ be a causal and time-invariant filter. Then $H$ has the fading memory property  if and only if the corresponding filter with almost surely uniformly bounded inputs has almost surely bounded outputs, that is, $H :(K^{L^{\infty}}_{M}, \| \cdot \| _{L^{\infty}_w}) \longrightarrow L^{\infty}(\Omega, \mathbb{R}) $, and it has the fading memory property.
\item [(ii)]  Let $\mathcal{T} := \left\{H _i: (K _M, \left\|\cdot \right\| _w) \longrightarrow \mathbb{R} \mid i \in I \right\}$ be a family of causal and time-invariant fading memory filters. Then, $\mathcal{T} $ is dense in the set $(C ^0(K_M),\|\cdot \| _w)$ if and only if the corresponding family with inputs in $K^{L^{\infty}}_{M} $ is universal in the set of continuous maps of the type $H :(K^{L^{\infty}}_{M}, \| \cdot \| _{L^{\infty}_w}) \longrightarrow L^{\infty}(\Omega, \mathbb{R}) $.
\end{description}
\end{theorem}

\paragraph{A first universality result using RC systems.} The following paragraphs contain a stochastic analog of Theorem~\ref{universality deterministic general} which shows that any fading memory filter with almost surely uniformly bounded inputs can be approximated using the elements of a polynomial algebra of reservoir filters with the same kind of inputs, provided that it contains the constant functionals and has the separation property. We note that, as in the deterministic case, the existence of the reservoir  filter associated to a reservoir system like \eqref{reservoir equation}-\eqref{readout} is guaranteed only in the presence of the echo state property. The next lemma shows that this property is inherited by deterministic fading memory reservoir filters with almost surely bounded inputs. 

\begin{lemma}
\label{lemma for reservoir stochastic}
Consider a reservoir system determined by the relations~\eqref{reservoir equation}--\eqref{readout} and the maps $F: D _N\times \overline{B _n({\bf 0}, M)}\longrightarrow  D _N$ and $h: D_N \rightarrow \mathbb{R}$, for some $n, N \in \mathbb{N} $, $M>0 $,   and $D_N \subset \mathbb{R} ^N $. If this reservoir system has the echo state and the fading memory properties then so does the corresponding system with stochastic inputs in $K^{L^{\infty}}_{M}$  which, additionally, has an associated  reservoir functional $H_h^F :(K^{L^{\infty}}_{M}, \| \cdot \| _{L^{\infty}_w}) \longrightarrow L^{\infty}(\Omega, \mathbb{R}) $ with almost surely bounded outputs that satisfies the fading memory property.
\end{lemma}

\begin{theorem}
\label{general universal stochastic}
Let $M > 0  $  and let $K_M^{L^{\infty}}$ be the set of almost surely uniformly bounded processes introduced in~\eqref{Kset stochastic}. Consider the set  $\mathcal{R} $
\begin{equation}
\label{generating family reservoirs stochastic}
\mathcal{R}:=\{H_{h _i}^{F _i}:K_M^{L^{\infty}} \longrightarrow L^{\infty}(\Omega, \mathbb{R})\mid h _i \in  {\rm Pol}(\mathbb{R} ^{N _i} , \mathbb{R}), F _i: \mathbb{R} ^{N _i} \times \mathbb{R} ^n  \longrightarrow \mathbb{R} ^{N _i}, i \in I, N _i\in  \mathbb{N}\}
\end{equation} 
formed by deterministic fading memory reservoir filters with respect to a given weighted norm $\| \cdot \| _w  $ and driven by stochastic inputs in $K_M^{L^{\infty}}$. 
Let $\mathcal{A}(\mathcal{R}) $ be the polynomial algebra generated by $\mathcal{R} $. If the algebra $\mathcal{A}(\mathcal{R})$ has the separation property and contains all the constant functionals, then any deterministic, causal, time-invariant fading memory filter $H:(K_M^{L^{\infty}}, \| \cdot \| _{L^{\infty}_w}) \longrightarrow L^{\infty}(\Omega, \mathbb{R})$ can be uniformly approximated  by elements in $\mathcal{A}(\mathcal{R})$, that is,  for any $\epsilon>0 $, there exist a finite set of indices $\{i _1, \ldots, i _r\} \subset I $ and a polynomial $p: \mathbb{R} ^r \longrightarrow \mathbb{R} $ such that 
\begin{equation*}
\| H - H _h ^F\| _{\infty} := \sup_{{\bf z} \in K_M^{L^{\infty}}} \{\|H ({\bf z})-H_h^F ({\bf z}) \|_{L^{\infty}}\}< \epsilon \quad \mbox{with} \quad h:=p(h_{i _1}, \ldots, h_{i _r})\quad \mbox{and} \quad F:=(F_{i _1}, \ldots, F_{i _r}).
\end{equation*}
\end{theorem}

In the next paragraphs we identify, as in the deterministic case, families of reservoirs that satisfy the hypotheses of this theorem and where we will eventually impose linearity constraints on the readout function.  The following corollary to Theorem~\ref{general universal stochastic} is the stochastic analog of Corollary~\ref{RCs are universal deterministic}.

\begin{corollary}
\label{RCs are universal stochastic}
Let $M > 0  $  and let $K_M^{L^{\infty}}$ be the set of almost surely uniformly bounded processes introduced in~\eqref{Kset stochastic}. Let
\begin{equation}
\label{generating family reservoirs all stochastic}
\mathcal{R}_w:=\{H_{h }^{F }:K_M^{L^{\infty}} \longrightarrow L^{\infty}(\Omega, \mathbb{R})\mid h  \in  {\rm Pol}(\mathbb{R} ^{N } , \mathbb{R}), F : \mathbb{R} ^{N } \times \mathbb{R} ^n  \longrightarrow \mathbb{R} ^{N }, N \in  \mathbb{N}\}
\end{equation} 
be the set of all the  reservoir filters defined on $K_M^{L^{\infty}}$ that  have the FMP with respect to a given weighted norm $\| \cdot \| _{L^{\infty}_w}  $. Then $\mathcal{R}_w$ is universal, that is,  for any time-invariant fading memory filter $H:(K_M^{L^{\infty}}, \| \cdot \| _{L^{\infty}_w}) \longrightarrow L^{\infty}(\Omega, \mathbb{R})$ and any $\epsilon>0 $, there exists  a reservoir filter $H_h^F \in \mathcal{R}_w$ such that 
$
\| H - H _h ^F\| _{\infty} := \sup_{{\bf z} \in K_M^{L^{\infty}}} \{\|H ({\bf z})-H_h^F ({\bf z}) \|_{L^{\infty}}\}< \epsilon.
$
\end{corollary}

\paragraph{Linear reservoir computers with stochastic inputs are universal.}

As it was the case in the deterministic setup, we can prove in the stochastic case that the linear RC family introduced in~(\ref{linear reservoir equation})-(\ref{linear readout}) suffices to achieve universality. The proof of the following statement is a direct consequence of Corollary ~\ref{linear universality deterministic} and Theorem ~\ref{fmp is inherited}. 

\begin{corollary}
\label{universality stochastic linear}
Let $M > 0  $  and let $K_M^{L^{\infty}}$ be the set of almost surely uniformly bounded processes introduced in~\eqref{Kset stochastic}.
Let $\mathcal{L} _\epsilon $ be the family introduced in Corollary~\ref{linear universality deterministic} and formed by all the linear reservoir filters $H^{A, {\bf c}}_p $ determined by matrices $A \in \mathbb{M}_N $ such that $\sigma_{{\rm max}}(A)< 1- \epsilon $. The elements in $\mathcal{L} _\epsilon $ map $K_M^{L^{\infty}} $ into $L^{\infty}(\Omega, \mathbb{R})$ and are time-invariant fading memory filters with respect to the weighted norm $\| \cdot \| _{w ^\rho} ^{L^{\infty}} $ associated to $w _t^\rho:=(1- \epsilon)^{ \rho t} $, for any $\rho \in (0,1)$. Moreover, they are universal, that is, for any time-invariant and causal fading memory filter $H:(K_M^{L^{\infty}}, \| \cdot \| _{L^{\infty}_{w ^\rho}}) \longrightarrow L^{\infty}(\Omega, \mathbb{R})$ and any $\varepsilon>0 $, there exists  $H^{A, {\bf c}}_p \in \mathcal{L} _\epsilon $ such that 
$
\| H - H^{A, {\bf c}}_p\| _{\infty} := \sup_{{\bf z} \in K_M^{L^{\infty}}} \{\|H ({\bf z})-H^{A, {\bf c}}_p({\bf z}) \|_{L^{\infty}}\}< \varepsilon.
$

The same universality result can be stated for the  subfamily  $\mathcal{DL} _\epsilon \subset \mathcal{L} _\epsilon$, formed by the linear reservoir systems in $\mathcal{L} _\epsilon$ determined by diagonal matrices, and for
$\mathcal{NL} \subset \mathcal{L} _\epsilon$, formed by the linear reservoir systems determined by nilpotent matrices.
\end{corollary}

\begin{remark}
\normalfont
The linear reservoir filters in $\mathcal{NL}$ determined by nilpotent matrices have been used in \cite{RC8} to formulate a $L ^p $ version of these universality results.
\end{remark}

\begin{remark}
\normalfont
The previous corollary has interesting consequences in the realm of time series analysis. Indeed, many well-known parametric time series models  consist in autoregressive relations, possibly nonlinear, driven by independent or uncorrelated innovations. The parameter constraints that are imposed on them in order to ensure that they have (second order) stationary solutions imply, in may situations, that the resulting filter has the FMP. In those cases, Corollary~\ref{universality stochastic linear} allows us to conclude that when those models are driven by almost surely uniformly bounded innovations, they can be arbitrarily well approximated by a polynomial function of a vector autoregressive model (VAR) of order 1. This statement applies, for example, to any stationary ARMA~\cite{Box1976, BrocDavisYellowBook} or GARCH~\cite{engle:arch, bollerslev:garch, Francq2010} model driven by almost surely uniformly bounded innovations.
\end{remark}

\paragraph{State-affine reservoir computers with almost surely uniformly bounded inputs are universal.}

As it was the case in the deterministic setup, non-homogeneous SAS are universal time-invariant fading memory filters  in the stochastic framework with almost surely uniformly bounded inputs. Their advantage with respect to the families proposed in the previous corollary is that they use a linear readout which is of major importance in practical implementations. More specifically, the following result holds true as a direct consequence of Theorem~\ref{universality of sas systems} and the equivalence stated in Theorem ~\ref{fmp is inherited}.

\begin{theorem}{\rm {\bf (Universality of SAS reservoir computers with almost surely uniformly bounded inputs)}}
\label{universality of sas systems stochastic}
Let $K ^{L^{\infty}} _I  \subset L^{\infty}(\Omega, \mathbb{R}^{\mathbb{Z}_-}) $ be the set of almost surely and uniformly bounded processes in the interval $I=[-1,1] $, that is,
\begin{equation*}
K ^{L^{\infty}} _I:= \left\{ {z} \in L^{\infty}(\Omega, \mathbb{R}^{\mathbb{Z}_-})  \mid \| {z} _t \|_{L^{\infty}} \leq 1, \quad \mbox{for all} \quad  t \in \Bbb Z _-\right\}.
\end{equation*}
Let $\mathcal{S} _\epsilon$ be the family of functionals $H_{{\bf W}}^{p,q}:K ^{L^{\infty}} _I  \longrightarrow L^{\infty}(\Omega, \mathbb{R})  $ induced by the state-affine systems defined in~\eqref{sas reservoir equation}-\eqref{sas readout} and that satisfy $M _p:=\max _{z \in I}\| p (z)\|  <1 - \epsilon$ and $M _q:=\max _{z \in I}\| q (z)\|  <1 - \epsilon$. The family  $\mathcal{S} _\epsilon$ forms a polynomial subalgebra of $\mathcal{R} _{w ^\rho} $ (as defined in~\eqref{generating family reservoirs all stochastic}) with $w^{\rho} _t:=(1- \epsilon )^{ \rho t} $, made of fading memory reservoir filters that map into $L^{\infty}(\Omega, \mathbb{R}) $.

Moreover, for any time-invariant and causal fading memory filter $H:(K ^{L^{\infty}} _I,\| \cdot \|  _{L^{\infty}_{w ^\rho}}) \longrightarrow L^{\infty}(\Omega, \mathbb{R})$ and any $\epsilon>0 $, there exist a natural number $N \in \mathbb{N} $,  polynomials $p (z) \in \mathbb{M}_{N,N}[z], q (z) \in \mathbb{M}_{N,1}[z]$ with $M _p, M _q< 1- \epsilon $, and a vector $ \mathbf{W} \in \mathbb{R}^N $ such that  
\begin{equation*}
\| H - H _{\bf W} ^{p,q}\| _{\infty} := \sup_{{z} \in K ^{L^{\infty}} _I} \{\|H ({z})-H _{\bf W} ^{p,q} ({z})\|_{L^{\infty}}\}< \epsilon.
\end{equation*}

The same universality result can be stated for the smaller subfamily $\mathcal{NS}_\epsilon\subset \mathcal{S}_\epsilon$ formed by SAS reservoir systems determined by nilpotent polynomials $p(z) \in\mathbb{N}{\rm il} [z]$.
\end{theorem}

\section{Conclusion}

This paper studies and proposes solutions for the universality problem in the approximation of fading memory filters using reservoir computer (RC) systems. RCs are a particular type of recurrent neural networks  that have  important applications both in machine learning and in signal processing where they exhibit superb information processing performances. Their importance is also linked to the possibility of building highly efficient hardware realizations.  RC systems are in general defined  as nonlinear state-space systems  determined by a reservoir and a readout map. In many supervised machine learning applications the readout  is chosen to be linear and the reservoir map is randomly generated, which  reduces the training of a dynamic task to a static regression problem and allows to circumvent well-known difficulties in the training of generic recurrent neural networks.
         
The universality question that we addressed consists in finding families of RCs as simple as possible such that  the set of input/output functionals that can be generated with them is dense in a sufficiently rich class. The work presented here is the dynamic counterpart of a  statement of this type for neural networks in a static and deterministic setup in which they have been proved to be universal approximators. 

The RC universality results stated in the paper correspond to two different situations in which the inputs are either deterministic and uniformly bounded or stochastic and almost surely uniformly bounded. In both cases we proved two different universality statements.  First, we showed that the  family of fading memory RCs is universal in the much larger fading memory filters category. The same applies to the much smaller RC family containing just linear reservoirs with polynomial readouts, when certain spectral restrictions are imposed on the reservoir maps.  The second result concerns exclusively reservoir computers with linear readouts, which are closer to the type of RCs used in applications and hardware implementations. More specifically, we introduced the family of what we called non-homogeneous  state-affine systems  and identified sufficient conditions  that guarantee that the associated reservoir computers with linear readouts are causal, time-invariant, and satisfy the echo state and the fading memory properties. Finally, we stated a universality result for a  subset of this class which  was shown to be universal in the same fading memory filters category as above.  These universality statements are then generalized to the stochastic setup for almost surely uniformly bounded inputs. In particular, we showed that any discrete-time filter that has the fading memory property with almost surely uniformly bounded stochastic inputs can be uniformly approximated by elements in the non-homogeneous state-affine family. All the density statements in the paper are formulated with respect to natural uniform approximation norms that appear in each of the different cases considered.

Despite preexisting work, these universality results are, to our knowledge, the first of their type in the semi-infinite discrete-time inputs setup. In the stochastic case they open the door to new developments in the learning theory of stochastic processes.

\section{Appendices}

\subsection{Proof of Lemma~\ref{uniformly bounded}}
Let $w : \mathbb{N} \longrightarrow (0,1] $ be an arbitrary weighting sequence. Then, for any ${\bf z} \in K_{M} $:
\begin{equation*}
\| {\bf z}\|_{w}:= \sup_{t \in \Bbb Z_-}\{\| {\bf z}_t w_{-t}\|\}=\sup_{t \in \Bbb Z_-}\{\| {\bf z}_t \|w_{-t}\}\leq M \cdot 1=M< \infty. 
\end{equation*}
Regarding the inequalities ~\eqref{ineq1rho} and ~\eqref{ineq2rho}, notice that if $w _t= \lambda ^t  $ then:
\begin{multline*}
\sum_{t=0}^{\infty}\| {\bf z} _{-t}\| w _t=\sum_{t=0}^{\infty}\| {\bf z} _{-t}\| \lambda ^t =
\sum_{t=0}^{\infty}\| {\bf z} _{-t}\| (\lambda^{1 - \rho} \lambda^\rho) ^t=
\sum_{t=0}^{\infty}\| {\bf z} _{-t}\| \lambda^{(1 - \rho)t} \lambda^{\rho t}\\
\leq  \sum_{t=0}^{\infty} \sup_{i \in \mathbb{N}} \left\{\| {\bf z} _{-i}\| \lambda^{(1 - \rho)i} \right\}\lambda^{\rho t}
=\sup_{i \in \mathbb{N}} \left\{\| {\bf z} _{-i}\| \lambda^{(1 - \rho)i} \right\}\sum_{t=0}^{\infty} \lambda^{\rho t}
= \| {\bf z}\|_{w^{1- \rho}}\frac{1}{1- \lambda ^\rho},
\end{multline*}
which proves~\eqref{ineq1rho}. The proof of~\eqref{ineq2rho} is similar and follows from noticing that:
\begin{equation*}
\sum_{t=0}^{\infty}\| {\bf z} _{-t}\| \lambda^{(1 - \rho)t} \lambda^{\rho t}\\
\leq  \sum_{t=0}^{\infty} \sup_{i \in \mathbb{N}} \left\{\| {\bf z} _{-i}\| \lambda^{\rho i} \right\}\lambda^{(1 - \rho) t)}
=\sup_{i \in \mathbb{N}} \left\{\| {\bf z} _{-i}\| \lambda^{\rho i} \right\}\sum_{t=0}^{\infty} \lambda^{(1-\rho) t}
= \| {\bf z}\|_{w^{\rho}}\frac{1}{1- \lambda ^{1-\rho}}. \quad \blacksquare
\end{equation*}

\subsection{Proof of Lemma~\ref{uniformly bounded and equicontinuous}} 

We recall first that by Lemma ~\ref{uniformly bounded} we have that $\| {\bf z}\|_w < \infty $, for any ${\bf z} \in K _M $. Second, since $(\ell ^{\infty}_w({\Bbb R}^n), \left\|\cdot \right\|_w) $  is a Banach space \cite{RC7}, it is hence metrizable and therefore so is $(K_M, \left\|\cdot \right\|_w) $ when endowed with the relative topology (see, for instance,~\cite[Exercise 1, Chapter 2, \textsection{21}]{Munkres:topology}). We will then conclude the compactness of $(K_M, \left\|\cdot \right\|_w) $ by showing that this space is sequentially compact (see, for example~\cite[Theorem 28.2]{Munkres:topology}). We proceed by using the strategy in the proof of Lemma 1 in~\cite{Boyd1985}.

For any $m \in \mathbb{N}  $, let $K _M^m $ be the set obtained by projecting into $\left(\mathbb{R}^n\right)^{\{-m, \ldots, -1,0\}} $ the elements of $K _M\subset ({\Bbb R}^n)^{\Bbb Z_-}$.  Given an element ${\bf z} \in K _M$, we will denote by ${\bf z}^{(m)}:=({\bf z} _{-m}, \ldots, {\bf z}_{0}) $ its projection into $K _M^m $. Additionally, notice that $K _M^m=\overline{B_n({\bf 0}, M)}^{m+1}$ is compact (and hence sequentially compact) with the product topology, since it is a product of  closed balls $\overline{B_n({\bf 0}, M)} \subset \mathbb{R}^n $ which are compact.

Let $\{{\bf z} (n)\}_{n \in \mathbb{N}}\subset K _M $ be a sequence of elements in $K _M $. The argument that we just stated proves that for any $k \in \mathbb{N} $, there is a subset $\mathbb{N}_k \subset \mathbb{N} $ and an element ${\bf z} ^{(k)} \in K _M^k$ such that  
\begin{equation*}
\max_{t \in \left\{-k, \ldots, 0\right\}} \left\| {\bf z} _t (n)- {\bf z} ^{(k)}_t\right\| \longrightarrow 0, \quad \mbox{as $n \rightarrow \infty $,\, $n \in \mathbb{N}_k$}.
\end{equation*}
Moreover, the sets $\mathbb{N} _k$ can be constructed so that $\mathbb{N}\supset \mathbb{N}_1\supset\mathbb{N} _2\supset \cdots $ and so that ${\bf z} ^{(k)} $ extends ${\bf z} ^{(l)}$ when $k\geq l $. This implies the existence of an element ${\bf z}  \in K _M$ such that, for each  $k \in \mathbb{N} $,
\begin{equation*}
\max_{t \in \left\{-k, \ldots, 0\right\}} \left\| {\bf z} _t (n)- {\bf z} _t\right\| \longrightarrow 0, \quad \mbox{as $n \rightarrow \infty $,\, $n \in \mathbb{N}_k$},
\end{equation*}
and hence there exists an increasing subsequence $n _k $ such that $n _k \in \mathbb{N} _k$ and that for each $k _0$,
\begin{equation}
\label{almost there 0}
\max_{t \in \left\{-k_0, \ldots, 0\right\}} \left\| {\bf z} _t (n_k)- {\bf z} _t\right\| \longrightarrow 0, \quad \mbox{as $k \longrightarrow \infty $}.
\end{equation}
We conclude by showing that the sequence $\left\{{\bf z}(n _k)\right\}_{k \in \mathbb{N}} $ converges in $(K_M, \left\|\cdot \right\|_w) $ to the element ${\bf z}\in K _M$. First, given that $w _t \longrightarrow 0 $ as $t \longrightarrow \infty $, then for any  $\varepsilon > 0 $ there exists $k _0 $ such that $w_{k}< \varepsilon/ 2M $, for any $k\geq k _0$. Additionally, since ${\bf z}(n _k), {\bf z}\in K _M$ for any $k \in \mathbb{N} $, we have that
\begin{equation}
\label{almost there 1}
\sup_{t \leq -k_0} \{\left\| {\bf z} _t (n_k)- {\bf z} _t\right\| w_{-t}\} \leq 2M w_{k _0}< \varepsilon.
\end{equation}
Now, by~\eqref{almost there 0} there exists $k _1 $ such that for any $k\geq k _1 $
\begin{equation}
\label{almost there 2}
\sup_{t \in \left\{-k_0, \ldots, 0\right\}} \{\left\| {\bf z} _t (n_k)- {\bf z} _t\right\|w_{-t}\}<
\sup_{t \in \left\{-k_0, \ldots, 0\right\}} \{\left\| {\bf z} _t (n_k)- {\bf z} _t\right\|\} < \varepsilon.
\end{equation}
Consequently,  \eqref{almost there 1} and \eqref{almost there 2} imply that for any $k> \max \{k _0, k _1\} $, $\left\|{\bf z}(n _k)- {\bf z}\right\| _w < \varepsilon $, as required. \quad $\blacksquare$

\subsection{Proof of Lemma~\ref{weighting sequence dependence}}
Let $\delta ^w( \epsilon) $ be the epsilon-delta relation for the FMP associated to the weighting sequence $w$. We now show that $H _U $ has the FMP with respect to $w'  $ via the epsilon-delta relation given by $\delta ^{w'}( \epsilon):= \delta ^{w}( \epsilon)/ \lambda$. Indeed, for any $\epsilon >0 $ and any ${\bf z}, {\bf s} \in K $ such that $\| {\bf z} - {\bf s}\|_{w '} <\delta ^{w'}( \epsilon) $, we have that
\begin{equation*}
\| {\bf z} - {\bf s}\|_w=\sup_{t \in \Bbb Z_-}\{\| {\bf z}_t-{\bf s}_t \|w_{-t}\}= \sup_{t \in \Bbb Z_-}\left\{\| {\bf z}_t-{\bf s}_t \|\frac{w_{-t}}{w'_{-t}}w'_{-t}\right\}< \lambda\sup_{t \in \Bbb Z_-}\{\| {\bf z}_t-{\bf s}_t\|w'_{-t}\}< \lambda \| {\bf z} - {\bf s}\|_{w'}< \lambda \delta ^{w'}( \epsilon)=\delta ^{w}( \epsilon),
\end{equation*}
and consequently, since $H _U  $ has the FMP with respect to the weighting sequence $w $, we can conclude that  $|H _U({\bf z})-H _U({\bf s})|< \epsilon $. This shows that the implication 
\begin{equation*}
\| {\bf z} - {\bf s}\|_{w '} <\delta ^{w'}( \epsilon) \Longrightarrow |H _U({\bf z})-H _U({\bf s})|< \epsilon
\end{equation*}
holds, as required. \quad $\blacksquare$

\subsection{Proof of Theorem~\ref{universality deterministic general}}
\label{proof of universality deterministic general}

Since the elements in $\mathcal{R} $ have the FMP with respect to a given weighted norm $\| \cdot \| _w  $, then so do those in $\mathcal{A}(\mathcal{R}) $ since polynomial combinations of continuous elements of the form $H_{h _i}^{F _i}:(K_M, \| \cdot \| _w) \longrightarrow \mathbb{R} $ are also continuous. Therefore, under that hypothesis,  $\mathcal{A}(\mathcal{R}) $ is a polynomial subalgebra of the algebra $(C ^0(K_M),\|\cdot \| _w)$ of real-valued continuous functions on $(K_M, \| \cdot \| _w)$. Since by hypothesis $\mathcal{A}(\mathcal{R})$ contains the constant functionals and separates the points in $K_M$ and,
by Lemma~\ref{uniformly bounded and equicontinuous}, the set $(K_M, \| \cdot \| _w)$ is compact, the Stone-Weierstrass theorem (Theorem 7.3.1 in~\cite{Dieudonne:analysis}) implies that  $\mathcal{A}(\mathcal{R}) $ is dense in $(C ^0(K_M),\|\cdot \| _w)$, which concludes the proof. \quad $\blacksquare$

\subsection{Proof of Corollary~\ref{linear universality deterministic}}
\label{proof of corollary linear universal}

In order to show that the reservoir systems in $\mathcal{L}_\epsilon $ induce reservoir filters, we first show that they have the echo state property by using the following lemma, whose proof can be found in \cite{RC7}.

\begin{lemma}
\label{contraction lemma in rc7}
Let $D_N \subset \mathbb{R}^N  $ and $D_n\subset {\Bbb R}^n $ and let $F: D _N\times D_n \longrightarrow D_N $ be a continuous reservoir map. Suppose that $F$ is a contraction map with contraction constant $0<r<1 $, that is:
\begin{equation*}
\left\|F(\mathbf{x}, {\bf z})-F(\mathbf{y}, {\bf z})\right\|\leq r \left\|\mathbf{x}- {\bf y}\right\|, \quad \mbox{for all $\mathbf{x}, {\bf y} \in D_N $ and all ${\bf z}\in D_n $},
\end{equation*}
then the corresponding reservoir system has the echo state property.
\end{lemma}

\medskip

\noindent  We start now by noting that the condition $\sigma_{{\rm max}}(A)< 1- \epsilon < 1$ implies that the reservoir map $F(\mathbf{x}, {\bf z}):=A \mathbf{x} + {\bf c} {\bf z} $ associated to \eqref{linear reservoir equation} is a contracting map with constant $\sigma_{{\rm max}}(A) $ which, by hypothesis, is smaller than one. Indeed,
\begin{equation*}
\left\|F(\mathbf{x}, {\bf z})-F(\mathbf{y}, {\bf z})\right\|
= \left\|A(\mathbf{x}- {\bf y})\right\| 
\leq \sigma_{{\rm max}}(A) \left\|\mathbf{x}- {\bf y}\right\|
 \quad \mbox{for all $\mathbf{x}, {\bf y} \in D_N $ and all ${\bf z}\in D_n $}.
\end{equation*}
By Lemma \ref{contraction lemma in rc7} we can conclude that this reservoir system has a reservoir filter associated that we now show is explicitly given by~\eqref{expression linear filter}.
We start by proving that the conditions $\sigma_{{\rm max}}(A)< 1- \epsilon < 1$ and that the elements in $K_M$  are uniformly bounded by a constant $M$ imply that the infinite sum in~\eqref{expression linear filter} is convergent. Let $n, m \in \mathbb{N}  $  be such that  $n<m $ and let $S _n:=  \sum _{i=0}^{n} A^i {\bf c} z_{-i} $. Now:
\begin{multline*}
\left\| S _n- S _m\right\| =\left\| \sum_{j=n+1}^m A^i {\bf c} {\bf z}_{-i}\right\| \leq 
\sum_{j=n+1}^m \left\|A\right\| _2^i \left\|{\bf c}\right\| _2 \|{\bf z}_{-i}\|\leq 
M \left\|{\bf c}\right\| _2\sum_{j=n+1}^m \sigma_{{\rm max}}(A)^i  \\
\leq M \left\|{\bf c}\right\| _2\sum_{j=n+1}^\infty \sigma_{{\rm max}}(A)^i =
M \left\|{\bf c}\right\| _2 \frac{\sigma_{{\rm max}}(A)^{n+1}}{1-\sigma_{{\rm max}}(A)}.
\end{multline*}
The condition $\sigma_{{\rm max}}(A)< 1- \epsilon < 1$ implies that $M \left\|{\bf c}\right\| _2 \frac{\sigma_{{\rm max}}(A)^{n+1}}{1-\sigma_{{\rm max}}(A)}=M  \frac{\sigma_{{\rm max}}({\bf c})\sigma_{{\rm max}}(A)^{n+1}}{1-\sigma_{{\rm max}}(A)} \rightarrow 0 $ as $n \rightarrow \infty $  and hence $\left\{ S _n\right\} _{n \in \mathbb{N} } $ is a Cauchy sequence in $\mathbb{R}^N $ that consequently converges.

The fact that the filter determined by the expression~\eqref{expression linear filter} is a solution of the recursions~\eqref{linear reservoir equation}-\eqref{linear readout} is a straightforward verification. In order to carry it out, it suffices to use that the filter $U^{A, {\bf c}}_h({\bf z}) $ associated to the functional $H^{A, {\bf c}}_h({\bf z}) $ is given by
\begin{equation*}
U^{A, {\bf c}}_h({\bf z})  _t=h \left(\sum _{i=0}^{\infty} A^i {\bf c} {\bf z}_{t-i}\right),
\end{equation*}
and that the time series $\widetilde{{\bf x} _t}  $ defined by $\widetilde{{\bf x} _t}  :=\sum _{i=0}^{\infty} A^i {\bf c} {\bf z}_{t-i}$ satisfies the recursion relation~\eqref{linear reservoir equation}.

We now verify by hand that the filters $U^{A, {\bf c}}_h$ are time-invariant. Let  ${\bf z} \in K_M $ and $t, \tau \in \mathbb{N} $ arbitrary and let $U _\tau $ be the corresponding time delay operator, then:
\begin{equation}
\label{tinvariancelhs}
\left(U^{A, {\bf c}}_h \circ U _\tau\right)({\bf z})_t=
\left(U^{A, {\bf c}}_h \left(U _\tau({\bf z})\right)\right)_t=
h \left(\sum _{i=0}^{\infty} A^i {\bf c} U _\tau({\bf z})_{t-i}\right)=
h \left(\sum _{i=0}^{\infty} A^i {\bf c} {\bf z}_{t-i- \tau}\right)
\end{equation}
At the same time,
\begin{equation*}
\left(U _\tau \circ U^{A, {\bf c}}_h\right)({\bf z})_t=
\left( U _\tau \left(U^{A, {\bf c}}_h({\bf z})\right)\right)_t=
U^{A, {\bf c}}_h({\bf z})_{t- \tau}=
h \left(\sum _{i=0}^{\infty} A^i {\bf c} {\bf z}_{t- \tau- i}\right),
\end{equation*}
which coincides with ~\eqref{tinvariancelhs} and proves the time-invariance of $U^{A, {\bf c}}_h $.

The next step consists in showing that the elements in $\mathcal{L} _\epsilon $ are $\lambda _\rho $-exponential fading memory filters, with $\lambda _\rho:=(1- \epsilon) ^\rho $, for any $\rho \in (0,1) $,  that is, $\mathcal{L} _\epsilon \subset  \mathcal{R}_{w^\rho}$, with $w^\rho: \mathbb{N} \rightarrow (0,1] $ the sequence given by $w _t^\rho:= (1- \epsilon) ^{ \rho t} $. Let $\| \cdot \| _{w ^\rho}$ be the associated weighted norm in $K_M$ and let ${\bf z}\in K_M $ be an arbitrary element. We start by noting that the continuity of the readout map $h: D_N\rightarrow \mathbb{R}$ implies that for any  $\varepsilon >0$ there exists an element $\delta(\varepsilon)>0 $ such that for any $ \mathbf{v} \in D _N $ that satisfies 
\begin{equation}
\label{continuoush}
\left\| \mathbf{v} - \sum _{i=0}^{\infty} A^i {\bf c} {\bf z}_{t-i}\right\|<\delta(\varepsilon),  \quad \mbox{then} \quad\left| h(\mathbf{v}) - h \left(\sum _{i=0}^{\infty} A^i {\bf c} {\bf z}_{t-i}\right)\right|<\varepsilon.
\end{equation}
We now show that for any ${\bf s} \in K_M $ such that 
\begin{equation}
\label{continuoushw}
\| {\bf s}- {\bf z}\| _{w ^\rho} < \frac{\delta(\varepsilon) \left(1- (1- \epsilon)^{1- \rho}\right)}{\sigma_{{\rm max}}({\bf c})}, \quad \mbox{then} \quad \left| H^{A, {\bf c}}_h({\bf s})-H^{A, {\bf c}}_h({\bf z})\right| < \varepsilon.
\end{equation}
Indeed,
\begin{multline*}
\left\|\sum _{i=0}^{\infty} A^i {\bf c} {\bf s}_{t-i} - \sum _{i=0}^{\infty} A^i {\bf c} {\bf z}_{t-i} \right\|=
\left\|\sum _{i=0}^{\infty} A^i {\bf c} ({\bf s}_{t-i} -  {\bf z}_{t-i}) \right\|\leq
\sum _{i=0}^{\infty} \left\|A^i {\bf c} ({\bf s}_{t-i} -  {\bf z}_{t-i}) \right\|\\
\leq \sum _{i=0}^{\infty} \sigma_{{\rm max}}(A^i) \left\|{\bf c} ({\bf s}_{t-i} -  {\bf z}_{t-i}) \right\|\leq
\sum _{i=0}^{\infty} \sigma_{{\rm max}}(A)^i \left\|{\bf c} ({\bf s}_{t-i} -  {\bf z}_{t-i}) \right\|\leq
\sum _{i=0}^{\infty} (1- \epsilon)^i \left\|{\bf c} ({\bf s}_{t-i} -  {\bf z}_{t-i}) \right\|.
\end{multline*}
If we now use~\eqref{ineq2rho} in Lemma~\ref{uniformly bounded} and the hypothesis in~\eqref{continuoushw}, we can conclude that
\begin{equation*}
\sum _{i=0}^{\infty} (1- \epsilon)^i \left\|{\bf c} ({\bf s}_{t-i} -  {\bf z}_{t-i}) \right\|
\leq \sigma_{{\rm max}}({\bf c})\sum _{i=0}^{\infty} (1- \epsilon)^i \left\| ({\bf s}_{t-i} -  {\bf z}_{t-i}) \right\|\leq
\frac{\sigma_{{\rm max}}({\bf c})\| {\bf s}- {\bf z}\| _{w ^\rho}}{1- (1- \epsilon)^{1- \rho}}< \delta(\varepsilon),
\end{equation*}
which proves the continuity of the map $H^{A, {\bf c}} _h:(K_M, \| \cdot \|_{w ^\rho})\longrightarrow \mathbb{R} $ and hence shows that $H^{A, {\bf c}} _h $ is a $\lambda _\rho $-exponential fading memory filter. 

In order to establish the universality statement in the corollary we will proceed, as in the proof of Theorem~\ref{universality deterministic general}, by showing that $\mathcal{L} _\epsilon $ is a polynomial algebra that contains the constant functionals and  separates the points in $K_M$ and then by invoking the Stone-Weierstrass theorem using the compactness of $(K_M, \| \cdot \| _{w ^\rho}) $.

In order to show that $(\mathcal{L} _\epsilon, \| \cdot \|_{w ^\rho}) $ is a polynomial algebra, notice first that if $A _1, A _2 \in \mathbb{M}_{N} $ are such that $\sigma_{{\rm max}}(A _1),\sigma_{{\rm max}}(A _2)< 1 - \epsilon $, then
\begin{equation}
\label{for oplus lin}
\sigma_{{\rm max}}(A _1\oplus A _2)=\max \left(\sigma_{{\rm max}}(A _1), \sigma_{{\rm max}}(A _2)\right)<1- \epsilon.
\end{equation} 
If we now take ${\bf c}_i \in \mathbb{M} _{N_i,n}$, $i \in \{1,2\} $ and $h _1, h _2 $ two real-valued polynomials in $N _1 $ and $N _2$ variables, respectively, we have by the first part of the corollary that we just proved that the filter functionals $H^{A_1, {\bf c}_1} _{h_1}$ and $ H^{A_2, {\bf c}_2} _{h_2}$  are well defined. Additionally, by \eqref{pol algebra product}-\eqref{pol algebra sum} so are the combinations $H^{A_1, {\bf c}_1} _{h_1} \cdot H^{A_2, {\bf c}_2} _{h_2} $ and $H^{A_1, {\bf c}_1} _{h_1} + \lambda H^{A_2, {\bf c}_2} _{h_2} $ that satisfy:
\begin{equation}
\label{algebra for linear}
H^{A_1, {\bf c}_1} _{h_1} \cdot H^{A_2, {\bf c}_2} _{h_2} = H^{A _1\oplus A_2, {\bf c}_1\oplus {\bf c}_2} _{h _1\cdot h_2}, \quad H^{A_1, {\bf c}_1} _{h_1} + \lambda H^{A_2, {\bf c}_2} _{h_2}= H^{A _1\oplus A_2, {\bf c}_1\oplus {\bf c}_2} _{h _1\oplus \lambda h_2}, \quad \lambda \in \mathbb{R}.
\end{equation}
Using the relations~\eqref{algebra for linear}  and   \eqref{for oplus lin}, we can conclude that both  $H^{A_1, {\bf c}_1} _{h_1} \cdot H^{A_2, {\bf c}_2} _{h_2}$ and  $H^{A_1, {\bf c}_1} _{h_1} + \lambda H^{A_2, {\bf c}_2} _{h_2}$ belong to $\mathcal{L} _\epsilon \subset  \mathcal{R}_{w^\rho}$. This implies that $(\mathcal{L} _\epsilon, \| \cdot \|_{w ^\rho}) $ is a polynomial subalgebra of $(\mathcal{R}_{w^\rho}, \| \cdot \|_{w ^\rho}) $

Since $\mathcal{L} _\epsilon $  contains the constant functionals (just take constant readout maps $h $), in order to conclude the proof, it is enough to show that the elements in $\mathcal{L} _\epsilon $  separate points in $K_M$. In the proof of this statement we need the following elementary fact about analytic functions.

\begin{lemma}
\label{analytic elementary}
Let $M > 0 $ and let ${\bf z}\in [-M,M]^{\mathbb{Z}_{-}} $. Define the real valued function $f _{{\bf z}} (x):=\sum_{j=0}^{\infty} z_{-j} x ^j $. This function is real analytic in the interval $(-1,1)$. Moreover, if ${\bf z}\neq {\bf 0} $, then there exists a point $x _0 \in (-1,1) $ such that $f _{{\bf z}} (x_0)\neq 0 $.
\end{lemma}

\noindent\textbf{Proof of the lemma.\ \ } We note first that for any $x \in (-1,1)$ and any $s \in \mathbb{N} $ we have that
\begin{equation*}
\left| \sum_{j=0}^s z_{-j} x ^j \right|\leq  \sum_{j=0}^s \left|z_{-j}\right| \left|x ^j\right|\leq M\sum_{j=0}^s  \left|x \right|^j\leq \frac{M}{1-|x|}.
\end{equation*}
Taking the limit $s \rightarrow \infty $, we obtain that
\begin{equation*}
\left| f _{{\bf z}} (x) \right|\leq \frac{M}{1-|x|}, \quad \mbox{for all $x \in (-1,1)$,} 
\end{equation*}
which proves the first claim in the lemma. Now, by the uniqueness theorem for the representation of analytic functions by power series (see \cite[page 217]{brown:churchill}), the series $\sum_{j=0}^{\infty} z_{-j} x ^j $ is the Taylor expansion around $0$ of $f _{{\bf z}} (x) $. Since ${\bf z} \neq {\bf 0} $
by hypothesis, some of the derivatives of $f _{{\bf z}} (x) $ are non-zero and hence this function cannot be flat, which implies that there exists a point $x _0 \in (-1,1) $ such that $f _{{\bf z}} (x_0)\neq 0 $. $\blacktriangledown $

\medskip

We now show that the elements in $\mathcal{L} _\epsilon $  separate points in $K_M$.
Take ${\bf z} _1, {\bf z} _2 \in K_M \subset  \left(\mathbb{R}^n\right)^{\Bbb Z_-}  $  such that  ${\bf z} _1\neq {\bf z} _2 $ and let $A \in \mathbb{M}(n,n) $, with $\sigma_{{\rm max}}(A)< 1 - \epsilon $, and ${\bf c}:= \mathbb{I} _n $. Let $U^{A, {\bf c}}: K _M \longrightarrow \left(\mathbb{R}^n\right)^{\Bbb Z_-} $ be the linear filter associated to $A $ and ${\bf c}$ via the recursion ~\eqref{linear reservoir equation}. Using the preceding arguments we have that
\begin{equation}
\label{expression linear 1}
U^{A, {\bf c}}({\bf z})_t = \sum_{j=0}^{\infty} A ^j {\bf z}_{t-j}.
\end{equation}
Since ${\bf z} _1\neq {\bf z} _2 $, then there exists and index $i \in \left\{1, \ldots, n\right\} $ and $t \in \mathbb{Z}_{-} $ such that $\left(z _1 ^i\right)_t\neq \left(z _2 ^i\right)_t $. Let now $b \in (-1+ \epsilon,1- \epsilon) $ and let $A _b:= {\rm diag} \left(0, \ldots,0,b,0, \ldots,0\right) \in \mathbb{D} _n$ be the matrix that has the element $b$ in the $i$-th entry. It is easy to see using  \eqref{expression linear 1} that
\begin{equation}
\label{expression linear 2}
U^{A_b, {\bf c}}({\bf z})_t = \left(0, \ldots, 0, \sum_{j=0}^{\infty} b ^j z^i_{t-j}, 0, \ldots,0\right)^{\top}, \quad \mbox{with} \quad \sum_{j=0}^{\infty} b ^j z^i_{t-j}\quad \mbox{in the $i$-th entry.}
\end{equation}
Let ${\bf s}:= {\bf z}_1- {\bf z}_2 \neq {\bf 0}$. Notice that by \eqref{expression linear 2} we have that $U^{A_b, {\bf c}}({\bf s})_0 = \left(0, \ldots, 0, \sum_{j=0}^{\infty} b ^j s^i_{-j}, 0, \ldots,0\right)^{\top}$. Given that the vector ${\bf s} ^i \in \mathbb{R}^{\mathbb{Z}_{-}}  $ is non-zero, Lemma \ref{analytic elementary}, implies the existence of an element $b _0 \in (-1+ \epsilon,1- \epsilon) $ such that $U^{A_{b_0}, {\bf c}}({\bf s})_0 \neq {\bf 0} $, which is equivalent to $U^{A_{b_0}, {\bf c}}({\bf z}_1)_0 \neq U^{A_{b_0}, {\bf c}}({\bf z}_2)_0 $. Using the polynomial $h(\mathbf{x}):=x  _i \in \mathbb{R}$, the previous relation implies that $U^{A_{b_0}, {\bf c}}_h({\bf z}_1)_0 \neq U_h^{A_{b_0}, {\bf c}}({\bf z}_2)_0 $ or, equivalently,
\begin{equation*}
H^{A_{b_0}, {\bf c}} _h \left({\bf z} _1\right)\neq H^{A_{b_0}, {\bf c}} _h \left({\bf z} _2\right), \quad \mbox{as required}.
\end{equation*}

We conclude the proof by establishing the universality the families $\mathcal{DL} _\epsilon $ and $\mathcal{NL} $ formed by the linear reservoir filters generated by  diagonal and nilpotent matrices, respectively. First, in the case of $\mathcal{DL} _\epsilon $, the statement is a consequence of \eqref{algebra for linear} and of the fact that when the matrices $A _1$ and $A_2 $ are diagonal, then the matrix associated to the linear map $A _1\oplus A_2 $ is also diagonal. Additionally, notice that the point separation property for $\mathcal{L} _\epsilon $ has been proved using diagonal matrices in \eqref{expression linear 2} and hence it also holds for $\mathcal{DL} _\epsilon $. The claim follows from the Stone-Weierstrass theorem. 

Finally, in the case of $\mathcal{NL} $, the proof also follows from \eqref{algebra for linear} since it is straightforward to see that when the matrices $A _1$ and $A_2 $ are nilpotent, then the matrix associated to the linear map $A _1\oplus A_2 $ is also nilpotent. It is only the point separation property of $\mathcal{N} $ that requires a separate argument that we provide in the following lines. Let ${\bf z} _1, {\bf z} _2 \in K_M$  such that  ${\bf z} _1\neq {\bf z} _2 $ and let $t_0 \in \mathbb{N}$ be the first  time index for which $\left({\bf z} _1\right)_{-t _0}\neq \left( {\bf z} _2\right)_{-t _0} $, that is, $\left({\bf z} _1\right)_{-t}= \left( {\bf z} _2\right)_{-t } $, for all $t \in \{0,1, \ldots, t _0-1\}$. Let now $i_0 \in \left\{1, \ldots, n\right\} $ be such that $\left(z _1 ^{i_0}\right)_{-t_0}\neq \left(z _2 ^{i_0}\right)_{-t_0} $. Let now $A_{t _0+1}\in \mathbb{N}{\rm il}_{t _0+1}^{t _0+1}$ be the upper shift matrix in dimension $t _0+1$, that is,  $A_{t _0+1}\in \mathbb{M}_{t _0+1}$ is by definition a superdiagonal matrix with a diagonal of ones above the main diagonal, and construct an element ${\bf c} \in \mathbb{M}_{t _0+1,n} $ whose last row is given by a vector of zeros with the exception of a one in the entry $i_0 $. The nilpotency of $A_{t _0+1} $ implies
\begin{equation*}
U^{A_{t _0+1}, {\bf c}}({\bf z}) _0=\sum_{j=0}^{t _0}A_{t _0+1} ^j {\bf c} {\bf z}_{-j}. 
\end{equation*}
When we apply this expression to ${\bf z} _1$ and  ${\bf z} _2 $, since $\left({\bf z} _1\right)_{-t}= \left( {\bf z} _2\right)_{-t } $, for all $t \in \{0,1, \ldots, t _0-1\}$, we obtain that 
\begin{equation*}
U^{A_{t _0+1}, {\bf c}}({\bf z}_1-{\bf z}_2)_0=A_{t _0+1} ^{t _0} {\bf c} ({\bf z_1}-{\bf z_2})_{-t _0}= \left(0, \ldots,0, \left(z _1 ^{i_0}\right)_{-t_0}- \left(z _2 ^{i_0}\right)_{-t_0}\right)^{\top}\neq {\bf 0}. 
\end{equation*}
Using the polynomial $h(\mathbf{x}):=x_{t _0+1} $, this relation implies that $U^{A_{t _0+1}, {\bf c}}_h({\bf z}_1)_0 \neq U_h^{A_{t _0+1}, {\bf c}}({\bf z}_2)_0 $ or, equivalently,
$
H^{A_{t _0+1}, {\bf c}} _h \left({\bf z} _1\right)\neq H^{A_{t _0+1}, {\bf c}} _h \left({\bf z} _2\right), \  \mbox{as required}.
$
\quad $\blacksquare$
\subsection{Proof of Proposition~\ref{integration sas}}
\label{Proof of proposition integration sas}

\noindent We start by noting, as we did in the proof of Corollary~\ref{linear universality deterministic}, that the condition~\eqref{condition for sas int} implies that the reservoir map associated to \eqref{sas reservoir equation} is a contraction and hence, by Lemma \ref{contraction lemma in rc7}, it satisfies the echo state property and has a well-defined associated filter.

We now prove that the condition ~\eqref{condition for sas int} implies the convergence of the series in the expression ~\eqref{sas integrated reservoir equation}. Let $K  _1:=\max _{z \in I}\| p (z)\| _2=\max _{z \in I}\sigma_{{\rm max}}(p (z)) <1 $  and $K _2:=\max _{z \in I}\| q (z)\| _2=\max _{z \in I}\sigma_{{\rm max}}(q (z))$; notice that $K _1$ and $K _2  $ are well-defined due to the compactness of $I $. Let now $n, m \in  \mathbb{N} $ be such that $n<m $  and let $S _n:=\sum_{j=0}^ n \left(\prod_{k=0}^{j-1}p(z_{t-k}) \right)q(z _{t-j}) \in \mathbb{R}^N$. Then,
\begin{eqnarray*}
\left\|S _n-S _m\right\| &=& \left\|\sum_{j=n+1}^ m \left(\prod_{k=0}^{j-1}p(z_{t-k}) \right)q(z _{t-j}) \right\| \leq
\sum_{j=n+1}^ m \left\|\prod_{k=0}^{j-1}p(z_{t-k}) \right\|_2 \left\|q(z _{t-j}) \right\| \\
	&\leq &\sum_{j=n+1}^ m \prod_{k=0}^{j-1}\left\|p(z_{t-k}) \right\|_2 \left\|q(z _{t-j}) \right\| \leq
	K _2 \sum_{j=n+1}^ m K _1^j\leq K _2 \sum_{j=n+1}^ \infty K _1^j= \frac{K _2 K _1^{n+1}}{1-K _1}.
\end{eqnarray*}
The condition $K _1<  1$ implies that $\frac{K _2 K _1^{n+1}}{1-K _1} \rightarrow 0 $ as $n \rightarrow \infty $  and hence $\left\{ S _n\right\} _{n \in \mathbb{N} } $ is a Cauchy sequence in $\mathbb{R}^N $ that consequently converges. This proves the convergence of the infinite series in~\eqref{sas integrated reservoir equation} and the causal character of the filter that it defines. The time-invariance can also be easily established by mimicking the verification that we carried out in the proof of Corollary~\ref{linear universality deterministic}.
We now prove that~\eqref{sas integrated reservoir equation} is indeed a solution of ~\eqref{sas reservoir equation}:
\begin{multline*}
p(z _t)\mathbf{x}_{t-1}+q( {z} _t) = p(z _t)\left(\sum_{j=0}^ \infty \left(\prod_{k=0}^{j-1}p(z_{t-1-k}) \right)q(z _{t-1-j})\right)+q( {z} _t)
=q(z _t)+p(z _t) q(z_{t-1})\\
+ p(z _t) p(z_{t-1})q(z_{t-2})+ p(z _t)p(z_{t-1})p(z_{t-2}) q(z_{t-3})+ \cdots 
=\sum_{j=0}^ \infty \left(\prod_{k=0}^{j-1}p(z_{t-k}) \right)q(z _{t-j})= \mathbf{x} _t. 
\end{multline*} 
We conclude by proving the inequality in \eqref{ineq state sas}. Note first that for any $m \in \mathbb{N}$,
\begin{multline*}
 \left\|\sum_{j=0}^ m \left(\prod_{k=0}^{j-1}p(z_{t-k}) \right)q(z _{t-j}) \right\| \leq
\sum_{j=0}^ m \left\|\prod_{k=0}^{j-1}p(z_{t-k}) \right\|_2 \left\|q(z _{t-j}) \right\| \\
	\leq \sum_{j=0}^ m \prod_{k=0}^{j-1}\left\|p(z_{t-k}) \right\|_2 \left\|q(z _{t-j}) \right\| \leq
	\frac{K _2 \left(1-K _1^{m+1}\right)}{1-K _1}, 
\end{multline*}
and hence, by the continuity of the norm and for any $t \in \Bbb Z$:
\begin{equation*}
\left\|\mathbf{x} _t\right\|= \lim_{m \rightarrow \infty} \left\|\sum_{j=0}^ m \left(\prod_{k=0}^{j-1}p(z_{t-k}) \right)q(z _{t-j}) \right\| \leq
 \lim_{m \rightarrow \infty} \frac{K _2 \left(1-K _1^{m+1}\right)}{1-K _1}= \frac{K _2 }{1-K _1}. \quad \blacksquare
\end{equation*}
\subsection{Proof of Lemma~\ref{conditions sas}}

\noindent {\bf (i)} $\Longrightarrow $ {\bf (ii)}: $\|A _0\| _2+\|A _1\| _2+ \cdots + \|A _{n_1}\| _2< \sum_{i=0}^{n _1}\lambda=\lambda(n_1+1)<1 $.

\noindent {\bf (ii)} $\Longrightarrow $ {\bf (iii)}: $\| p (z)\| _2=\| A _0+zA _1+z ^2A _2+ \cdots + z ^{n_1} A _{n_1}\| _2\leq \|A _0\| _2+|z|\|A _1\| _2+|z ^2|\|A _2\| _2+ \cdots + |z ^{n_1}| \|A _{n_1}\| _2 <\|A _0\| _2+\|A _1\| _2+ \cdots + \|A _{n_1}\| _2 <1$. \quad $\blacksquare$

\subsection{Proof of Proposition~\ref{fmp sas}}
\label{proof of fmp sas}

We start by formulating and proving an elementary result that will be needed later on.

\begin{lemma}
\label{lip for differential}
Let $ {\bf f}: U \subset \mathbb{R}^n \longrightarrow \mathbb{M}_m $ be a differentiable function defined on the convex set $U $. For any ${\bf z} \in U $ denote by $\partial _i {\bf f} ({\bf z}) \in \mathbb{M}_m$ the matrix containing the partial derivatives of the components of $ {\bf f}$ with respect to their ith-entry, $i \in \left\{1, \ldots, n\right\}$. 
Then, for any $\mathbf{x}, {\bf y} \in U$ we have:
\begin{equation}
\label{lip ineq convex}
\left\| {\bf f}({\bf y})- {\bf f}(\mathbf{x})\right\|_2 \leq \sqrt{nm}\max_{i \in \left\{1, \ldots, n\right\}} \left(\sup_{{\bf z} \in U} \{\left\|\partial _i {\bf f} ({\bf z})\right\|_2\}\right)\left\|\mathbf{x}- {\bf y}\right\|.
\end{equation} 
\end{lemma}

\noindent\textbf{Proof.\ \ } Given $A =(A _{i,j})\in \mathbb{M}_{n,m} $, let $\left\|A\right\|_F:= \mbox{tr} \left(A ^{\top}A \right) =\sum_{i=1}^n\sum_{j=1}^m A _{i,j} ^2$ be its Frobenius norm. Recall (see  Theorem 5.6.34 and Exercise 5.6.P24 in~\cite{horn:matrix:analysis}) that 
\begin{equation}
\label{ineq norms matrix}
\left\|A\right\| _2\leq \left\|A\right\|_F\leq \sqrt{r}\left\|A\right\| _2,
\end{equation}
where $r $ is the rank of $A$. Consider now $\mathbf{x}, {\bf y} \in U$ arbitrary and let $D {\bf f} ( {\bf z}): {\Bbb R}^n \longrightarrow \mathbb{M}_m $  be the differential of ${\bf f} $ evaluated at ${\bf z}\in U $. The convexity of $U$ implies that the Mean Value Inequality holds (see Theorem 2.4.8 in~\cite{mta}) and hence:
\begin{equation}
\label{first ineq mta}
\left\| {\bf f}({\bf y})- {\bf f}(\mathbf{x})\right\|_F \leq  \sup_{t \in [0,1]} \{\left\|D {\bf f} ((1-t){\bf x}+ t {\bf y})\right\|_2\} \left\|\mathbf{x}- {\bf y}\right\|.
\end{equation}
The first inequality in ~\eqref{ineq norms matrix} and~\eqref{first ineq mta} imply that
\begin{equation}
\label{second ineq mta}
\left\| {\bf f}({\bf y})- {\bf f}(\mathbf{x})\right\|_2 \leq  \sup_{{\bf z} \in U}\{ \left\|D {\bf f} ({\bf z})\right\|_2\} \left\|\mathbf{x}- {\bf y}\right\|.
\end{equation}
At the same time, notice that by ~\eqref{ineq norms matrix}
\begin{multline*}
 \left\|D {\bf f} ({\bf z})\right\|_2  ^2\leq \left\|D {\bf f} ({\bf z})\right\|_F  ^2 = 
\sum_{i=1}^n\sum_{j=1}^{m}\sum_{k=1}^{m} \partial _i f _{j k} ^2({\bf z})=
\sum_{i=1}^n \left\|\partial _i {\bf f} ({\bf z})\right\|^2_F \\
\leq
m\sum_{i=1}^n \left\|\partial _i {\bf f} ({\bf z})\right\|^2_2\leq
m n\max_{i \in \left\{1, \ldots, n\right\}} \left(\left\|\partial _i {\bf f} ({\bf z})\right\|^2_2\right).
\end{multline*}
This inequality, together with ~\eqref{second ineq mta}, imply the statement ~\eqref{lip ineq convex} since the maximum and the supremum can be trivially exchanged. \quad $\blacktriangledown$

We now carry out the proof of the proposition under the hypothesis {\bf (iii)} in Lemma ~\ref{conditions sas} which is implied by the other two. The modifications necessary to establish the result under the other two hypotheses are straightforward.
Consider two arbitrary elements ${\bf z}, {\bf s} \in I^{\Bbb Z_-} $. Then, by the Cauchy-Schwarz and  Minkowski inequalities:
\begin{multline}
\label{first step prop sas}
\left| H_{{\bf W}}^{p,q} ({\bf z})- H_{{\bf W}}^{p,q}({\bf s}) \right|= \left| {\bf W}^\top \left[\sum_{j=0}^ \infty \left(\left(\prod_{k=0}^{j-1}p(z_{-k}) \right)q(z _{-j}) - \left(\prod_{k=0}^{j-1}p(s_{-k}) \right)q(s _{-j})\right)\right]\right|\\
\leq \left\|{\bf W}\right\| \sum_{j=0}^ \infty  \left\| a _j(\underline{z_{-j+1}})q(z_{-j})- a _j(\underline{s_{-j+1}})q(s_{-j})\right\|, \quad \mbox{where} \quad a _j(\underline{z_{-j+1}}):=\prod_{k=0}^{j-1}p(z_{-k}).
\end{multline} 
We now bound the right hand side of ~\eqref{first step prop sas} as follows:
\begin{multline}
\label{second step prop sas}
\sum_{j=0}^ \infty  \left\| a _j(\underline{z_{-j+1}})q(z_{-j})- a _j(\underline{s_{-j+1}})q(s_{-j})\right\|\\
=\sum_{j=0}^ \infty  \left\| a _j(\underline{z_{-j+1}})q(z_{-j}) +
a _j(\underline{z_{-j+1}})q(s_{-j})-
a _j(\underline{z_{-j+1}})q(s_{-j})
- a _j(\underline{s_{-j+1}})q(s_{-j})\right\|\\
\leq \sum_{j=0}^ \infty  \left\| a _j(\underline{z_{-j+1}})\right\| _2 \left\|q(z_{-j}) -
q(s_{-j})\right\|+\left\|a _j(\underline{z_{-j+1}})
- a _j(\underline{s_{-j+1}})\right\| _2  \left\|q(s_{-j})\right\|
\end{multline}
If $L _q$ is a Lipschitz constant of $q : I \longrightarrow \mathbb{R}^N  $ then
\begin{equation}
\label{first place to bound}
\left\| a _j(\underline{z_{-j+1}})\right\| _2 \left\|q(z_{-j}) -
q(s_{-j})\right\| \leq M _p ^j L _q \left|z_{-j}-s_{-j}\right|,
\end{equation}
which inserted in ~\eqref{second step prop sas} and in ~\eqref{first step prop sas} implies that
\begin{equation}
\label{third step prop sas}
\left| H_{{\bf W}}^{p,q} ({\bf z})- H_{{\bf W}}^{p,q}({\bf s}) \right| \leq \left\|{\bf W}\right\| L _q 
\left[
\sum_{j=0}^ \infty M _p ^j  \left| z_{-j}-s_{-j}\right|
+\sum_{j=0}^ \infty \left\|a _j(\underline{z_{-j+1}})
- a _j(\underline{s_{-j+1}})\right\| _2
\right]
\end{equation}
We now bound above the second summand in \eqref{third step prop sas} using the inequality  \eqref{lip ineq convex} in the statement of Lemma~\ref{lip for differential} as well as the following identity: 
\begin{multline}
\label{relation with as}
a _j(\underline{z_{-j+1}})- a _j(\underline{s_{-j+1}})=
\sum_{l=0}^{j-1}(
p(s _0) \cdots p(s_{-(l-1)})\cdot p(z_{-l}) \cdot p(z_{-(l+1)}) \cdots p(z_{-(j-1)})\\-
p(s _0) \cdots p(s_{-(l-1)})\cdot p(s_{-l}) \cdot p(z_{-(l+1)}) \cdots p(z_{-(j-1)})).
\end{multline}
This equality simply follows from writing:
\begin{multline*}
a _j(\underline{z_{-j+1}})- a _j(\underline{s_{-j+1}})=\prod_{l=0}^{j-1} p(z_{-l}) - \prod_{l=0}^{j-1} p(s_{-l})= p(z_{0}) p(z_{-1}) \cdots p(z_{-(j-1)}) - {p(s_{0}) p(s_{-1}) \cdots p(s_{-(j-1)})}\\
={p(z_{0}) p(z_{-1}) \cdots p(z_{-(j-1)}) }- {p(s_{0}) p(s_{-1}) \cdots p(s_{-(j-1)})} \\
+\Bigg\{{{ p(s_{0}) p(z_{-1}) \cdots p(z_{-(j-1)})}} -{p(s_{0}) p(z_{-1}) \cdots p(z_{-(j-1)}) }\\
+{{ {p(s_{0}) p(s_{-1}) p(z_{-2})\cdots p(z_{-(j-1)})}}} - {{ p(s_{0}) p(s_{-1})  p(z_{-2})\cdots p(z_{-(j-1)}) }}\\
+ \cdots +{p(s_{0})  \cdots p(s_{-(l - 1)}) p(z_{-l}) p(z_{-(l + 1)})\cdots  p(z_{-(j-1)}) }- p(s_{0})  \cdots p(s_{-(l - 1)}) p(z_{-l}) p(z_{-(l + 1)})\cdots  p(z_{-(j-1)}) \\
+ \cdots +{p(s_{0}) \cdots p(s_{-(j-2)}) p(z_{-(j-1)}) }- {{{ {p(s_{0}) \cdots p(s_{-(j-2)}) p(z_{-(j-1)})}}}} \Bigg\} 
\\ =\sum_{l=0}^{j-1}(
p(s _0) \cdots p(s_{-(l-1)})\cdot p(z_{-l}) \cdot p(z_{-(l+1)}) \cdots p(z_{-(j-1)})\\-
p(s _0) \cdots p(s_{-(l-1)})\cdot p(s_{-l}) \cdot p(z_{-(l+1)}) \cdots p(z_{-(j-1)})),
\end{multline*}
where the $2(j-1)$ summands inside the braces are obtained by adding and  subtracting  polynomials recursively constructed out of $a _j(\underline{z_{-j+1}})$ by changing the variables of the first $k$ factors, $k\in \{1, \cdots, j-1\}$.  We then combine all the $(2k-1)$-th with the $(2k+2)-th$ summands of the resulting expression in order to obtain the first $j-1$ terms in the sum in \eqref{relation with as}. Then the last $j$-th term results from combining the second with the one before last   summands, that is, $p(s_{0}) p(s_{-1}) \cdots p(s_{-(j-1)}) $ and $p(s_{0}) \cdots p(s_{-(j-2)}) p(z_{-(j-1)}) $, respectively.

Using the relation \eqref{relation with as} we can write:
\begin{multline*}
\left\|a _j(\underline{z_{-j+1}})- a _j(\underline{s_{-j+1}})\right\| _2\leq 
\sum_{l=0}^{j-1}\left\|p(s _0) \cdots p(s_{-(l-1)})\cdot (p(z_{-l})-p(s_{-l})) \cdot p(z_{-(l+1)}) \cdots p(z_{-(j-1)})\right\| _2 \\
\leq \sum_{l=0}^{j-1}\left\|p(s _0)\right\| _2 \cdots \left\|p(s_{-(l-1)})\right\| _2\cdot \left\|p(z_{-l})-p(s_{-l})\right\| _2 \cdot \left\|p(z_{-(l+1)})\right\| _2 \cdots \left\|p(z_{-(j-1)})\right\| _2\\
\leq M _p^{j-1} \sqrt{N}\sup_{z \in I} \left\{\left\|p'(z)\right\|_2\right\}\sum_{l=1}^{j} \left|z_{-j+l}-s_{-j+l}\right|,
\end{multline*}
where the last inequality is a consequence of~\eqref{lip ineq convex}.
Let $M _{p'}:=\sqrt{N}\sup_{z \in I} \left\{\left\|p'(z)\right\|_2\right\} $, then
\begin{equation*}
\left\|a _j(\underline{z_{-j+1}})- a _j(\underline{s_{-j+1}})\right\| _2\leq \frac{M _{p'}}{M _p}M _p^j\sum_{l=1}^{j} \left|z_{-j+l}-s_{-j+l}\right|=
 \frac{M _{p'}}{M _p}\sum_{l=1}^{j} M _p ^l M _p^{j-l}\left|z_{-(j-l)}-s_{-(j-l)}\right|
\end{equation*}
Since the last term in this inequality is one summand of the Cauchy product of the series with general terms
$M _p ^j$ and $M _p ^j \left|z_{-j}-s_{-j}\right| $ and these two series are absolutely convergent (recall the statement \eqref{ineq1rho}), we can conclude (see for instance  \cite[\textsection{8.24}]{Apostol:analysis}) that 
\begin{multline*}
\label{bound needed 2}
\sum_{j=0}^{\infty}\left\|a _j(\underline{z_{-j+1}})- a _j(\underline{s_{-j+1}})\right\| _2\leq 
 \frac{M _{p'}}{M _p}\sum_{j=0}^{\infty}\sum_{l=1}^{j} M _p ^l M _p^{j-l}\left|z_{-(j-l)}-s_{-(j-l)}\right|\\=
 \frac{M _{p'}}{M _p} \frac{1}{1- M _p}\sum_{j=0}^{\infty}M _p ^j\left|z_{-j}-s_{-j}\right|.
\end{multline*}
 If we now substitute this relation in  \eqref{third step prop sas} and we use Lemma  \ref{uniformly bounded}  with weighting sequences $w_t ^\rho:= M _p ^{ \rho t}$, for any $\rho\in (0,1) $, we obtain that:
\begin{eqnarray*}
\label{fourth step prop sas}
\left| H_{{\bf W}}^{p,q} ({\bf z})- H_{{\bf W}}^{p,q}({\bf s}) \right| &\leq& \left\|{\bf W}\right\| L _q 
 \left(1+  \frac{M _{p'}}{M _p} \frac{1}{1- M _p}\right)
\sum_{j=0}^ \infty M _p ^j  \left| z_{-j}-s_{-j}\right|\\
&\leq &
\left\|{\bf W}\right\| L _q 
 \left(1+  \frac{M _{p'}}{M _p} \frac{1}{1- M _p}\right) \left(\frac{1}{1- M _p^{1 - \rho}}\right)
\left\|{\bf z} - {\bf s}\right\|_{w ^\rho},
\end{eqnarray*}
which proves the continuity of the map $H_{{\bf W}}^{p,q}:(I^{\Bbb Z_-}, \| \cdot \| _{w^\rho}) \longrightarrow \mathbb{R}  $, as required. \quad $\blacksquare$

\subsection{Proof of Proposition~\ref{SAS polynomial algebra} }
  
We first recall that since by hypothesis the reservoir functionals $H_{{\bf W}_1}^{p_1,q_1}, H_{{\bf W}_2}^{p_2,q_2} $  are well-defined then,  by the comments that follow \eqref{direct filter}, so are $H_{{\bf W}_1}^{p_1,q_1}+ \lambda H_{{\bf W}_2}^{p_2,q_2} $ and $H_{{\bf W}_1}^{p_1,q_1}\cdot  H_{{\bf W}_2}^{p_2,q_2} $.

The proof of $ {\bf (i)} $ is a straightforward verification.  As to $ {\bf (ii)} $,  denote first by $y _t ^1, y _t ^2 $ and $\mathbf{x} _t ^1, \mathbf{x} _t ^2 $ the outputs and the state variables, respectively, of the SAS corresponding  to the two functionals that we are considering. We note first that by~\eqref{sas readout}:
\begin{equation*}
\label{readout product}
y _t ^1\cdot  y _t ^2 =  {\bf W}_1^\top\mathbf{x} ^1_t \cdot {\bf W}_2^\top\mathbf{x} ^2_t = \left({\bf W}_1\otimes {\bf W}_2 \right)^\top(\mathbf{x} ^1_t\otimes \mathbf{x} ^2_t).
\end{equation*}
Using~\eqref{sas reservoir equation} it can be readily verified that the time evolution of the tensor product $\mathbf{x} ^1_t\otimes \mathbf{x} ^2_t $ is given by
\begin{align*}
\mathbf{x} ^1_t\otimes \mathbf{x} ^2_t&=(p _1(z _t)\otimes p _2(z _t)) (\mathbf{x} ^1_{t-1}\otimes \mathbf{x} ^2_{t-1}) 
+p _1(z _t)\mathbf{x} ^1_{t-1}\otimes q _2(z _t) +q _1(z _t)\otimes p _2(z _t)\mathbf{x} ^2_{t-1}+q _1(z _t)\otimes q _2(z _t),\\
&=(p _1\otimes p _2)(z _t) (\mathbf{x} ^1_{t-1}\otimes \mathbf{x} ^2_{t-1}) 
+p _1(z _t)\mathbf{x} ^1_{t-1}\otimes q _2(z _t) +q _1(z _t)\otimes p _2(z _t)\mathbf{x} ^2_{t-1}+(q _1 \otimes q _2)(z _t) ,\\
\end{align*}
which proves~\eqref{matrix for sas} and hence~\eqref{product sas filters}. 

In order to show that the reservoir functionals on the right hand side of  \eqref{sum sas filters} and \eqref{product sas filters} are well-defined we prove the following lemma.

\begin{lemma}
\label{lemma for sums and products}
Let $p_1(z)\in \mathbb{M}_{N_1,M _1}[z] $ and $p_2(z)\in \mathbb{M}_{N_2,M _2}[z] $ be two polynomials with matrix coefficients and assume that they satisfy that $\|p _1 (z)\| _2<1- \epsilon $ and $\|p _2 (z)\| _2 <1- \epsilon$ for all $z \in I:=[-1,1] $ and a given $0<\epsilon>1 $. Then:
\begin{description}
\item [(i)] $\|p _1 \oplus p _2(z)\| _2<1- \epsilon $,
\item [(ii)]  $\|p _1 \otimes p _2(z)\| _2<1- \epsilon $, 
\end{description}
for all $z \in I:=[-1,1] $.
\end{lemma}

\noindent\textbf{Proof of the lemma.\ \ } Let $ \mathbf{x}= \mathbf{x} _1\oplus \mathbf{x} _2 \in \mathbb{R}^{M _1} \oplus \mathbb{R}^{M _2}$. Then, in order to prove part {\bf (i)} note that
\begin{multline*}
\|(p _1 \oplus p _2)(z)\cdot \mathbf{x}\|^2=\|(p _1(z)\cdot \mathbf{x}_1,  p _2(z)\cdot \mathbf{x}_2) \| ^2= \|p _1(z)\cdot \mathbf{x}_1\| ^2+\|  p _2(z)\cdot \mathbf{x}_2 \| ^2\\
\leq 
\|p _1(z)\| ^2 _2\| \mathbf{x}_1\| ^2+\|  p _2(z)\| ^2 _2\|\mathbf{x}_2 \| ^2\leq
(1- \epsilon)^2 \left(\|\mathbf{x}_1 \| ^2+ \|\mathbf{x}_2 \| ^2\right)=(1- \epsilon)^2 \| \mathbf{x}\| ^2.
\end{multline*}
This inequality implies that
\begin{equation*}
\|p _1 \oplus p _2(z)\| _2= \sup_{\mathbf{x}\neq {\bf 0}} \left\{\frac{\|(p _1 \oplus p _2)(z)\cdot \mathbf{x}\|}{\| \mathbf{x}\|}\right\} \leq  \sup_{\mathbf{x}\neq {\bf 0}} \left\{\frac{(1- \epsilon)\| \mathbf{x}\|}{\| \mathbf{x}\|}\right\}=1- \epsilon, \quad \mbox{as required.}
\end{equation*}
As to the statement in part {\bf (ii)}:
\begin{equation*}
\|p _1 \otimes p _2(z)\| _2= \sigma_{{\rm max}} (p _1 \otimes p _2(z))=\sigma_{{\rm max}} (p _1(z)) \sigma_{{\rm max}} (p _2(z))= \|p _1 (z) \| _2 \| p _2(z)\| _2<(1- \epsilon) ^2<(1- \epsilon). \  
\blacktriangledown
\end{equation*}

Now, the first part of this lemma and Proposition \ref{integration sas} guarantee that $H_{{\bf W}_1\oplus \lambda{\bf W}_2}^{p _1\oplus p_2,q _1\oplus q_2} $ is well-defined. The same conclusion holds for $H_{{\bf 0}\oplus {\bf 0}\oplus \left({\bf W}_1\otimes {\bf W}_2\right)}^{p,q _1\oplus q_2\oplus \left(q _1\otimes q_2\right)} $ because due to the block diagonal character of~\eqref{matrix for sas} then $\sigma_{{\rm max}}(p(z))=\sigma_{{\rm max}}((p _1(z)\oplus p _2(z)\oplus \left(p _1\otimes p _2\right)(z))=\|  p _1(z)\oplus p _2(z)\oplus \left(p _1\otimes p _2\right)(z)\| _2$. By parts {\bf (i)} and {\bf (ii)} in Lemma~\ref{lemma for sums and products} we can conclude that $\|p(z)\| _2< 1- \epsilon $ for all $z \in [-1,1]$ and, again by Proposition \ref{integration sas}, the reservoir functional $H_{{\bf 0}\oplus {\bf 0}\oplus \left({\bf W}_1\otimes {\bf W}_2\right)}^{p,q _1\oplus q_2\oplus \left(q _1\otimes q_2\right)} $ is well-defined. \quad $\blacksquare$

\subsection{Proof of Theorem~\ref{universality of sas systems}}
\label{Proof of universality of sas systems}

Note first that the hypothesis $M _p<1- \epsilon <1$ on  the polynomials $p$ associated to the elements in $\mathcal{S} _\epsilon $ implies, by Propositions~\ref{integration sas} and~\ref{fmp sas}, that this family is made of time-invariant reservoir filters that have the FMP with respect to weighting sequences of the form $w ^p _t:=M _p ^{ \rho t} $, $\rho \in (0,1) $. Additionally, using Lemma~\ref{weighting sequence dependence} and the hypothesis $M _p< 1- \epsilon$, for a fixed given  $\epsilon\in (0,1) $, we can conclude that all the reservoir filters in $\mathcal{S} _\epsilon $ have the FMP with the common weighting sequence $w  _t ^\rho:= (1- \epsilon )^{ \rho t}  $, $\rho \in (0,1) $.

The elements in $\mathcal{S} _\epsilon$ form a polynomial algebra as a consequence  of Lemma~\ref{lemma for sums and products} and Proposition~\ref{SAS polynomial algebra}. Moreover,
the family $\mathcal{S} _\epsilon$ has the point separation property and contains all the constant functionals. Indeed, since $\mathcal{S} _\epsilon$ includes the linear family $\mathcal{L} _\epsilon $, we recall that  in Appendix \ref{proof of corollary linear universal} we proved that given ${\bf z} _1, {\bf z} _2 \in K_M \subset  \left(\mathbb{R}^n\right)^{\Bbb Z_-}  $  such that  ${\bf z} _1\neq {\bf z} _2 $, there exists $A \in \mathbb{M}(n,n) $, with $\sigma_{{\rm max}}(A)< 1 - \epsilon $ and ${\bf c}:= \mathbb{I} _n $ such that $U^{A, {\bf c}} ({\bf z}_1)_0\neq U^{A, {\bf c}} ({\bf z}_2)_0$. The point separation property follows from choosing any vector ${\bf W} \in \mathbb{R}^N$ such that  ${\bf W}^\top (U^{A, {\bf c}} ({\bf z}_1))_0\neq {\bf W}^\top (U^{A, {\bf c}} ({\bf z}_2))_0$, which implies that $U^{A, {\bf c}}_{{\bf W}} ({\bf z}_1)_0\neq U^{A, {\bf c}}_{{\bf W}} ({\bf z}_2)_0 $ and hence $H_{U^{A, {\bf c}}_{{\bf W}} }({\bf z}_1)\neq H_{U^{A, {\bf c}}_{{\bf W}} }({\bf z}_2) $, as required. 

All the constant functionals can be obtained by taking for $p$ the zero polynomial and for $q$ the constant polynomials ($q$ has degree zero). In that case, the state variables are a constant sequence $\mathbf{x} _t= q $ and the associated functional is the constant map $H^{0,q}_{{\bf W}}({\bf z})= {\bf W}^{\top}q $, for all ${\bf z} \in K _M $.

The universality result follows hence from the 
Stone-Weierstrass Theorem and the compactness of $(I ^{\Bbb Z _-} , \| \cdot \|_{w^\rho}) $ established in Lemma~\ref{uniformly bounded and equicontinuous}. 

Finally, we prove the statement regarding the family $\mathcal{NS}_\epsilon$ determined by nilpotent polynomials $p$. First, by expressions \eqref{sum sas filters}, \eqref{product sas filters}, and \eqref{matrix for sas}, it is easy to show that this family is a polynomial algebra. The only point that requires some detail is the fact that the $k $-th power of the polynomial $p$ in \eqref{matrix for sas} that is obtained in the product of the two SAS reservoir functionals $H_{{\bf W}_1}^{p_1,q_1}$ and $  H_{{\bf W}_2}^{p_2,q_2} $ is given by 
\begin{equation*}
p^k(z):=
\left(
\begin{array}{ccc}
p _1^k(z)& {\bf 0} &{\bf 0}\\
{\bf 0}&p _2 ^k(z)& {\bf 0}\\
p _1^k\otimes q _2^{k-1}(z) & q _1^{k-1}\otimes p _2^k (z)&p _1^k\otimes p _2^k(z)
\end{array}
\right),
\end{equation*}
which shows that if $p _1 $  and $p _2 $ are nilpotent then so is the associated polynomial $p$. The point separation property is, again, inherited from the proof of linear case provided in the Appendix \ref{proof of corollary linear universal}.
\quad $\blacksquare$

\subsection{Proof of Lemma~\ref{properties linfinity}}

\noindent\textbf{(i) } Let $A:=\left\{\rho\in \overline{\mathbb{R}_+}\mid \left\|{\bf X}\right\|_B\le \rho \quad \mbox{almost surely}\right\}$. It suffices to show that $\left\|{\bf X}\right\|_{L^{\infty}}:=\inf A  \in A $, which implies that $\left\| {\bf X}\right\|_B \leq \left\| {\bf X}\right\| _{L^{\infty}} $ almost surely. Indeed, consider the sequence $ \left\| {\bf X}\right\| _{L^{\infty}}+ 1/j $, $j \in \mathbb{N} $. By the approximation property of the infimum, there exists a decreasing sequence of numbers $ \{\rho _j\} _{ j \in \mathbb{N}}  \subset A$ in $A$ satisfying $\left\| {\bf X}\right\| _{L^{\infty}}\leq \rho_j<\left\| {\bf X}\right\| _{L^{\infty}}+ 1/j $ for all $j \in \mathbb{N} $. Define $F := \left\{\omega \in \Omega \mid \left\|{\bf X} (\omega)\right\| _B> \left\| {\bf X}\right\| _{L^{\infty}}\right\} $ and $F_j := \left\{\omega \in \Omega \mid \left\|{\bf X} (\omega)\right\|_B > \rho _j\right\} $. It is easy to see that  $F _j \subset F _{j+1} $, $j \in \mathbb{N} $ and that $\lim_{j \rightarrow \infty}F_j=F $ and, consequently, (see ~\cite[Lemma 5, page 7]{Grimmett2001}) $\lim_{j \rightarrow \infty}\mathbb{P}(F_j)=\mathbb{P}(F)$. Since by construction $\mathbb{P}(F_j) =0$ for all $j \in \mathbb{N} $ then  $\mathbb{P}(F) =0$ necessarily, which shows that $\left\|{\bf X}\right\|_{L^{\infty}} \in A $, as required.

\noindent {\bf (ii)} If $ \left\| {\bf X}\right\| _{L^{\infty}} \leq C$ then by part {\bf (i)}, $\left\| {\bf X}\right\| _B\leq \left\| {\bf X}\right\| _{L^{\infty}}\leq C $ almost surely. Conversely, if $\left\| {\bf X}\right\|_B\leq C $ almost surely, then $C \in A=\left\{\rho\in\overline{\mathbb{R}^+}\mid \left\|{\bf X}\right\|_B\le \rho \quad \mbox{almost surely}\right\}$. Consequently, $\left\|{\bf X}\right\|_{L^{\infty}}=\inf A  \leq C \in A$, as required. 

\noindent {\bf (iii)} Suppose first that $\left\|{\bf X}\right\| _B\leq C  $ almost surely and define $F := \left\{\omega \in \Omega \mid \left\|{\bf X} (\omega)\right\| _B> C\right\} $. By hypothesis, we have that $\mathbb{P} (F)=0 $ and $\mathbb{P}(\Omega\setminus F)=1 $. Then,
\begin{eqnarray*}
{\rm E}\left[\left\|{\bf X}\right\|_B^k\right]&=& \int _\Omega \left\|{\bf X}\right\|_B^k d\mathbb{P}= \int _{\Omega\setminus F} \left\|{\bf X}\right\|_B^k d\mathbb{P} +  \int _{F} \left\|{\bf X}\right\|_B^k d\mathbb{P}\\
	&=&\int _{\Omega\setminus F} \left\|{\bf X}\right\|_B^k d\mathbb{P}\leq \int _{\Omega\setminus F} C ^k d\mathbb{P}=C ^k \mathbb{P}(\Omega\setminus F)=C ^k,
\end{eqnarray*} 
as required. 
Conversely, assume that ${\rm E}\left[\left\|{\bf X}\right\|_B^k\right] \leq C ^k $, for any $k \in \mathbb{N}   $, and define $$F_n := \left\{\omega \in \Omega \mid \left\|{\bf X} (\omega)\right\| _B> C+ \frac{1}{n}\right\}, $$ for all $n \geq 1 $. It is easy to see that  $F _n \subset F _{n+1} $ and that $\lim_{n \rightarrow \infty}F_n=F $ and, consequently, (see ~\cite[Lemma 5, page 7]{Grimmett2001}) $\lim_{n \rightarrow \infty}\mathbb{P}(F_n)=\mathbb{P}(F)$. Now, 
\begin{eqnarray*}
C ^k &\geq &{\rm E}\left[\left\|{\bf X}\right\|_B^k\right]= \int _\Omega \left\|{\bf X}\right\|_B^k d\mathbb{P}= \int _{\Omega\setminus F _n} \left\|{\bf X}\right\|_B^k d\mathbb{P} +  \int _{F _n} \left\|{\bf X}\right\|_B^k d\mathbb{P}\\
	&\geq&\int _{F_n} \left\|{\bf X}\right\|_B^k d\mathbb{P}\geq \int _{ F_n} \left(C+ \frac{1}{n}\right) ^k d\mathbb{P}=\left(C+ \frac{1}{n}\right) ^k \mathbb{P}(  F_n),
\end{eqnarray*} 
which implies that $\mathbb{P}( F_n) \leq C ^k/\left(C+ \frac{1}{n}\right) ^k $ for any $ k \in \mathbb{N} $ and hence, by taking the limit $k \rightarrow \infty   $, we can conclude that $\mathbb{P}( F_n)=0 $. Consequently,  $\mathbb{P}(F)=\lim_{n \rightarrow \infty}\mathbb{P}(F_n)=0$, which shows that $\left\|{\bf X}\right\| _B\leq C  $ almost surely.

\noindent {\bf (iv)} Let $||\cdot||$ denote the Euclidean norm on $\mathbb{R}^n$. Since $\left|X _i\right| \leq \left\|{\bf X}\right\| $ always and by part {\bf (i)} $\left\| {\bf X}\right\| \leq \left\| {\bf X}\right\| _{L^{\infty}} $ almost surely, we can conclude that $\left|X _i\right| \leq \left\| {\bf X}\right\| _{L^{\infty}} $ almost surely. This implies that $X _i\in L^{\infty}(\Omega, \mathbb{R}) $ and hence the statement follows from part {\bf (iii)}. \quad $\blacksquare$

\subsection{Proof of Lemma \ref{swap sups}}

We start by proving by contradiction that 
\begin{equation}
\label{first ineq c}
\mathop{{\rm ess\, sup}}_{\omega \in \Omega}  
\left\{
\mathop{{\rm sup}}_{ t \in \Bbb Z} \left\{\| {\bf z} _t(\omega)
\|\right\}\right\}\geq 
\mathop{{\rm sup}}_{ t \in \Bbb Z}  
\left\{
\mathop{{\rm ess\, sup}}_{\omega \in \Omega} \left\{\| {\bf z} _t(\omega)
\|\right\}\right\}.
\end{equation}
Indeed, suppose  that 
\begin{equation}
\label{contradiction1}
\mathop{{\rm ess\, sup}}_{\omega \in \Omega}  
\left\{
\mathop{{\rm sup}}_{ t \in \Bbb Z} \left\{\| {\bf z} _t(\omega)
\|\right\}\right\}<
\mathop{{\rm sup}}_{ t \in \Bbb Z}  
\left\{
\mathop{{\rm ess\, sup}}_{\omega \in \Omega} \left\{\| {\bf z} _t(\omega)
\|\right\}\right\}.
\end{equation}
By the approximation property of the supremum \cite[Theorem 1.14]{Apostol:analysis}, there exists $t _0\in \Bbb Z $ such that 
\begin{equation}
\label{first ineq 1}
\mathop{{\rm ess\, sup}}_{\omega \in \Omega}  
\left\{
\mathop{{\rm sup}}_{ t \in \Bbb Z} \left\{\| {\bf z} _t(\omega)
\|\right\}\right\}<
\mathop{{\rm ess\, sup}}_{\omega \in \Omega} \left\{\| {\bf z} _{t _0}(\omega)
\|\right\}
\leq
\mathop{{\rm sup}}_{ t \in \Bbb Z}  
\left\{
\mathop{{\rm ess\, sup}}_{\omega \in \Omega} \left\{\| {\bf z} _t(\omega)
\|\right\}\right\}.
\end{equation}
However, $\| {\bf z} _{t _0}(\omega)
\|\leq \sup_{t \in \Bbb Z}\{\| {\bf z} _{t }(\omega)
\| \}$ for all $\omega\in \Omega  $ and hence by part {\bf (i)} in Lemma \ref{properties linfinity}
\begin{equation*}
\| {\bf z} _{t _0}(\omega)
\|\leq \sup_{t \in \Bbb Z}\{\| {\bf z} _{t }(\omega)
\|\}\leq \mathop{{\rm ess\, sup}}_{\omega \in \Omega}  
\left\{
\mathop{{\rm sup}}_{ t \in \Bbb Z} \left\{\| {\bf z} _t(\omega)
\|\right\}\right\}, \quad \mbox{almost surely.}
\end{equation*}
Now, by part {\bf (ii)} in Lemma \ref{properties linfinity}, this implies that 
\begin{equation*}
\mathop{{\rm ess\, sup}}_{\omega \in \Omega}  
\left\{\| {\bf z} _{t _0}(\omega)
\|\right\}
\leq  \mathop{{\rm ess\, sup}}_{\omega \in \Omega}  
\left\{
\mathop{{\rm sup}}_{ t \in \Bbb Z} \left\{\| {\bf z} _t(\omega)
\|\right\}\right\}.
\end{equation*}
However, this expression is in contradiction with the first inequality in \eqref{first ineq 1} and hence the assumption \eqref{contradiction1} cannot be correct. This argument implies that the inequality \eqref{first ineq c} holds.

We now prove the reverse inequality, that is,
\begin{equation}
\label{reverse ineq1}
\mathop{{\rm ess\, sup}}_{\omega \in \Omega}  
\left\{
\mathop{{\rm sup}}_{ t \in \Bbb Z} \left\{\| {\bf z} _t(\omega)
\|\right\}\right\}\leq 
\mathop{{\rm sup}}_{ t \in \Bbb Z}  
\left\{
\mathop{{\rm ess\, sup}}_{\omega \in \Omega} \left\{\| {\bf z} _t(\omega)
\|\right\}\right\}.
\end{equation}
By part {\bf (ii)} of Lemma  \ref{properties linfinity}, this inequality holds if and only if
\begin{equation}
\label{reverse ineq2}
\mathop{{\rm sup}}_{ t \in \Bbb Z} \left\{\| {\bf z} _t(\omega)
\|\right\}\leq 
\mathop{{\rm sup}}_{ t \in \Bbb Z} 
\left\{
\mathop{{\rm ess\, sup}}_{\omega \in \Omega} \left\{\| {\bf z} _t(\omega)
\|\right\}\right\}, \quad \mbox{almost surely.}
\end{equation}
Now, by part {\bf (i)} in Lemma  \ref{properties linfinity},  we have that $\| {\bf z} _t(\omega)\|\leq \mathop{{\rm ess\, sup}}_{\omega \in \Omega} \left\{\| {\bf z} _t(\omega)
\|\right\} $, almost surely and for each fixed $t \in \Bbb Z$. Let $A _t \subset \Omega  $ be the zero-measure set such that $\| {\bf z} _t(\omega)\|> \mathop{{\rm ess\, sup}}_{\omega \in \Omega} \left\{\| {\bf z} _t(\omega)
\|\right\} $ for all $\omega \in A _t  $. Let $A:=\bigcup_{t \in \Bbb Z}A _t $. Notice that $\mathbb{P}(A)=\mathbb{P}\left(\bigcup_{t \in \Bbb Z}A _t\right) \leq \sum_{t \in \Bbb Z}\mathbb{P}(A _t)=0$ and hence $B:=A^\mathsf{c} $ has measure one and 
\[
\| {\bf z} _t(\omega)\|\leq \mathop{{\rm ess\, sup}}_{\omega \in \Omega} \left\{\| {\bf z} _t(\omega)
\|\right\}, \quad \mbox{for all $\omega\in B $ and all $t \in \Bbb Z$.}
\]
Since $B$ has measure one, this inequality is equivalent to   \eqref{reverse ineq2}, which guarantees that  \eqref{reverse ineq1} holds. The inequalities \eqref{first ineq c} and \eqref{reverse ineq1} that we just proved imply that the equality \eqref{equality swap sups} holds true.
 \quad $\blacksquare$

\subsection{Proof of Lemma \ref{lemma linfi}}

It is obvious that $S_{ \ell ^{\infty}({\Bbb R}^n) } \subset S_{ ({\Bbb R}^n)^{\mathbb{Z}}}$ and hence the inclusion map 
\begin{align}
\label{iota}
\iota  &: S_{\ell ^{\infty}({\Bbb R}^n)}\hookrightarrow S_{(\mathbb{R}^n) ^{\mathbb{Z}}}, 
\end{align}
is well-defined. The equivariance with respect to the equivalence relations $\sim_{\ell ^{\infty}({\Bbb R}^n)}$ and $\sim_{(\mathbb{R}^n) ^{\mathbb{Z}}}$ follows trivially from noticing that if ${\bf z}_1, {\bf z}_2 \in  S_{\ell ^{\infty}({\Bbb R}^n)}$ are such that  ${\bf z}_1\sim_{\ell ^{\infty}({\Bbb R}^n) }{\bf z}_2$ one obviously have that $\iota({\bf z})_1\sim_{(\mathbb{R}^n) ^{\mathbb{Z}} }\iota({\bf z}_2)$. This shows the existence of the projected map $\phi$ that makes the diagram 
\begin{diagram}
S_{\ell ^{\infty}({\Bbb R}^n)} &\rInto^{\enspace \iota} &S_{ ({\Bbb R}^n)^{\mathbb{Z}}}\\
\dTo^{\Pi_{\sim_{\ell ^{\infty}({\Bbb R}^n)}}} & &\dTo_{\Pi_{\sim_{({\Bbb R}^n)^{\mathbb{Z}}}}}\\
L ^{\infty}\left(\Omega, \ell ^{\infty}({\Bbb R}^n)\right) &\rTo^{\phi} &L ^{\infty}\left(\Omega, ({\Bbb R}^n)^{\mathbb{Z}}\right),
\end{diagram}
commutative
where $\Pi_{\sim_{\ell ^{\infty}({\Bbb R}^n)}}$ and $\Pi_{\sim_{({\Bbb R}^n)^{\mathbb{Z}}}}$  map the elements in $S_{\ell ^{\infty}({\Bbb R}^n)}$ and $S_{(\mathbb{R}^n) ^{\mathbb{Z}}}$ onto their corresponding equivalence classes with respect to the associated equivalence relations. One can easily prove that the norm preservation following the diagram. It is a straightforward exercise to verify that $\phi  $  is injective and preserves the norm $\left\|\cdot \right\|_{L^{\infty}} $. 
In order to show that $\phi$ is surjective, let $ {\bf z} \in L ^{\infty}\left(\Omega, ({\Bbb R}^n)^{\mathbb{Z}}\right)$. Given that $\left\|{\bf z}\right\|_{L^{\infty}}< \infty $ or, equivalently, $\mathop{{\rm ess\, sup}}_{\omega \in \Omega}  
\left\{
\mathop{{\rm sup}}_{ t \in \Bbb Z} \left\{\| {\bf z} _t(\omega)
\|\right\}\right\}< \infty $, by part {\bf (i)} in Lemma \ref{properties linfinity}, this implies that 
\begin{equation}
\label{inter sup}
\sup_{ t \in \Bbb Z}\{\| {\bf z} _t(\omega)
\|\}< \infty, \quad \mbox{almost surely.}  
\end{equation}
Since the elements in the spaces in $L ^{\infty}\left(\Omega, \ell ^{\infty}({\Bbb R}^n)\right) $ and $L ^{\infty}\left(\Omega, ({\Bbb R}^n)^{\mathbb{Z}}\right) $ are equivalence classes containing almost surely equal random variables, we can take another representative ${\bf z} ^\ast  : \Omega \longrightarrow ({\Bbb R}^n) ^{\mathbb{Z}}$ for the class containing ${\bf z} \in L ^{\infty}\left(\Omega, ({\Bbb R}^n)^{\mathbb{Z}}\right)$ defined as
\begin{equation*}
{\bf z} ^\ast (\omega):=
\left\{
\begin{array}{cc}
{\bf z}(\omega),&\quad \mbox{when} \quad \sup_{ t \in \Bbb Z}\{\| {\bf z} _t(\omega)
\|\}< \infty,\\
0, & \quad \mbox{otherwise}.
\end{array}
\right.
\end{equation*} 
%$$\minCDarrowwidth55pt
%\begin{CD}
%S_{\ell ^{\infty}({\Bbb R}^n)}    @>\iota>> S_{ ({\Bbb R}^n)^{\mathbb{Z}}}\\
%@V\Pi_{\sim_{\ell ^{\infty}({\Bbb R}^n)}}VV        @VV\Pi_{\sim_{({\Bbb R}^n)^{\mathbb{Z}}}}V\\
%L ^{\infty}\left(\Omega, \ell ^{\infty}({\Bbb R}^n)\right)     @>{\phi}>>  L ^{\infty}\left(\Omega, ({\Bbb R}^n)^{\mathbb{Z}}\right).
%\end{CD}$$
Since the processes ${\bf z}  $ and ${\bf z}^\ast  $ differ by \eqref{inter sup} only in a set of zero measure, they are equal in $L ^{\infty}\left(\Omega, (\mathbb{R}^n) ^{\mathbb{Z}}\right)  $ but, this time, ${\bf z} ^\ast \in L ^{\infty}\left(\Omega, \ell ^{\infty}({\Bbb R}^n)\right) $ and $\phi({\bf z} ^\ast)= {\bf z} $, as required. \quad $\blacksquare$

\subsection{Proof of Theorem~\ref{fmp is inherited}}

\noindent {\bf Proof of part {\bf (i)}}. All along this proof we will denote the elements in $K _M $ with a lower bold case (${\bf z} \in K _M $) and those in $K _M ^{L^{\infty}} $ with an upper bold case (${\bf Z} \in K _M^{L^{\infty}}$). 

We first assume that the functional $H: (K _M, \left\|\cdot \right\| _w) \longrightarrow \mathbb{R}  $ has the fading memory property. This means that $H$ is a continuous map and since by Lemma ~\ref{uniformly bounded and equicontinuous} the space $(K _M, \left\|\cdot \right\| _w) $ is compact, then so is the image $H(K _M) $ as a subset of the real line. This implies that there exists a finite real number $L>0 $ such that $H(K _M) \subset [-L,L]$. Let now $ {\bf Z} \in K^{L^{\infty}}_{M} $; the condition $ \| {\bf Z}  \|_{L^{\infty}} \leq M$ is equivalent to $\left\|{\bf Z} _t\right\| \leq M $, for all $t \in \mathbb{Z}_{-}  $, almost surely, and hence implies that $H \left( {\bf Z}\right) \in [-L, L] $, almost surely or, equivalently, that $\left\|H \left({\bf Z}\right)\right\|_{L^{\infty}}\leq L $. This, in turn, implies that $H ({\bf Z}) \in L^{\infty}(\Omega, \mathbb{R}) $ for any ${\bf Z} \in K^{L^{\infty}}_{M} $, as required. 

We now show that $H :(K^{L^{\infty}}_{M}, \| \cdot \| _{L^{\infty}_w}) \longrightarrow L^{\infty}(\Omega, \mathbb{R}) $ has the FMP.
The FMP hypothesis on $H: (K _M, \left\|\cdot \right\| _w) \longrightarrow \mathbb{R}  $ implies that for any ${\bf z} \in K _M $ and any $\epsilon>0  $ there exists a $\delta(\epsilon)> 0 $ such that for any ${\bf s} \in K _M $ that satisfies that
\begin{equation}
\label{fmp for later}
\| {\bf z} - {\bf s}\|_w=\sup_{t \in \Bbb Z_-}\{\| ({\bf z}_t-{\bf s}_t) w_{-t}\|\}< \delta(\epsilon), \quad \mbox{then} \quad |H  ({\bf z})-H  ({\bf s})|< \epsilon.
\end{equation}
Moreover, since by Lemma ~\ref{uniformly bounded and equicontinuous} the space $(K _M, \left\|\cdot \right\| _w) $ is compact, the Uniform Continuity Theorem ~\cite[Theorem 7.3]{Munkres:topology} guarantees that the relation $\delta(\epsilon) $ does not depend on the point ${\bf z} \in K _M $.

We  now prove the statement by showing that for any $\epsilon>0 $ and ${\bf Z} \in K _M^{L^{\infty}} $ then $\|H  ({\bf Z})-H  ({\bf S})\| _{L^{\infty}}< \epsilon $, for all ${\bf S} \in K _M^{L^{\infty}} $ such that $\left\|{\bf Z}- {\bf S}\right\| _{L^{\infty}_{w}} < \delta (\epsilon)$. Indeed, the inequality $\left\|{\bf Z}- {\bf S}\right\| _{L^{\infty}_{w}} < \delta (\epsilon)$ holds  if and only if $\sup_{t \in \Bbb Z_-}\{\left\|{\bf Z}_t- {\bf S}_t\right\| _{L^{\infty}} w _{-t}\}< \delta (\epsilon)$. Given that for any $l\in \Bbb Z_- $ we have that $\left\|{\bf Z}_l- {\bf S}_l\right\| _{L^{\infty}} w _{-l}\leq\sup_{t \in \Bbb Z_-}\{\left\|{\bf Z}_t- {\bf S}_t\right\| _{L^{\infty}} w _{-t}\}< \delta (\epsilon)$, part {\bf (ii)} in Lemma ~\ref{properties linfinity} implies that $\left\|{\bf Z}_l- {\bf S}_l\right\| w _{-l}< \delta (\epsilon)$ almost surely for any $l\in \Bbb Z_- $ and hence $\sup_{t \in \Bbb Z_-}\{\left\|{\bf Z}_t- {\bf S}_t\right\|  w _{-t}\}= \left\|{\bf Z}- {\bf S}\right\| _w< \delta (\epsilon)$, almost surely. This implies, using ~\eqref{fmp for later}, that $|H  ({\bf Z})-H  ({\bf S})|< \epsilon $, almost surely,  which by part {\bf (ii)} in Lemma ~\ref{properties linfinity} implies that $\|H  ({\bf Z})-H  ({\bf S})\|_{L^{\infty}}< \epsilon $, as required.

Conversely, if $H :(K^{L^{\infty}}_{M}, \| \cdot \| _{L^{\infty}_w}) \longrightarrow L^{\infty}(\Omega, \mathbb{R}) $ has the fading memory property then so does $H: (K _M, \left\|\cdot \right\| _w) \longrightarrow \mathbb{R}  $ because $K _M \subset K^{L^{\infty}}_{M} $ and 
$
\left\|{\bf z}\right\|= \left\|{\bf z}\right\| _{L^{\infty}}$   for the elements ${\bf z}\in K _M $.

\medskip

\noindent {\bf Proof of part {\bf (ii)}}. We suppose first that $\mathcal{T}$ is dense in the set $(C ^0(K_M),\|\cdot \| _w)$ and show that the corresponding family with intputs in $K^{L^{\infty}}_{M} $ is universal. Let $H :(K^{L^{\infty}}_{M}, \| \cdot \| _{L^{\infty}_w}) \longrightarrow L^{\infty}(\Omega, \mathbb{R}) $ be an arbitrary causal and time-invariant FMP filter and let $H_S \in \mathcal{T}$ be such that  $\sup_{{\bf z} \in K_M} \{\|H ({\bf z})-H_S ({\bf z}) \|_{L^{\infty}}\}< \epsilon $. The existence of $H_S $ is ensured by the density hypothesis on $\mathcal{T} $. We   show that this ensures that $\sup_{{\bf Z} \in K_M^{L^{\infty}}}\{ \|H ({\bf Z})-H_S ({\bf Z}) \|_{L^{\infty}}\}< \epsilon $. Indeed, this conclusion is true if $\|H ({\bf Z})-H_S ({\bf Z}) \|_{L^{\infty}}< \epsilon $  for any ${\bf Z} \in K_M^{L^{\infty}} $ which, by part {\bf (ii)} in Lemma ~\ref{properties linfinity} is equivalent to  $|H ({\bf Z})-H_S ({\bf Z}) |< \epsilon $ almost surely,  for any ${\bf Z} \in K_M^{L^{\infty}} $. This condition is in turn true because as ${\bf Z} \in K_M^{L^{\infty}} $, then $\left\| {\bf Z} _t\right\| \leq M $ almost surely for all $t \in \Bbb Z_- $ and hence ${\bf Z} \in K_M $ almost surely. Since $H_S$ approximates $H $ for deterministic inputs, we have that $|H ({\bf Z})-H_S ({\bf Z}) |< \epsilon $ almost surely, as required. 

Conversely, if the family $\mathcal{T} $ with intputs in $K^{L^{\infty}}_{M} $ is universal in the set of continuous maps of the type $H :(K^{L^{\infty}}_{M}, \| \cdot \| _{L^{\infty}_w}) \longrightarrow L^{\infty}(\Omega, \mathbb{R}) $ we can easily show that $\mathcal{T}$ is dense in  $(C ^0(K_M),\|\cdot \| _w)$. Let $H \in (C ^0(K_M),\|\cdot \| _w)$ and let $H _S :(K^{L^{\infty}}_{M}, \| \cdot \| _{L^{\infty}_w}) \longrightarrow L^{\infty}(\Omega, \mathbb{R}) $ be the element that, for a given $\epsilon>0 $, satisfies  $\|H  -H_S   \|_{L^{\infty}}=\sup_{{\bf Z} \in K_M^{L^{\infty}}} \{\|H ({\bf Z})-H_S ({\bf Z}) \|_{L^{\infty}}\}< \epsilon $. Given that, as we pointed out, $K _M \subset K^{L^{\infty}}_{M} $ and 
$
\left\|{\bf z}\right\|= \left\|{\bf z}\right\| _{L^{\infty}}$,   for the elements ${\bf z}\in K _M $, we have
\begin{equation*}
\left\|H-H _S\right\|=\sup_{{\bf z} \in K_M} \{\|H ({\bf z})-H_S ({\bf z}) \|\}=\sup_{{\bf z} \in K_M} \{\|H ({\bf z})-H_S ({\bf z}) \|_{L^{\infty}}\}\leq\sup_{{\bf Z} \in K_M^{L^{\infty}}} \{\|H ({\bf Z})-H_S ({\bf Z}) \|_{L^{\infty}}\}< \epsilon.\quad\blacksquare
\end{equation*} 

\subsection{Proof of Lemma \ref{lemma for reservoir stochastic}}

As we pointed out in Section \ref{Notation, definitions, and preliminary discussions}, if the reservoir system determined by $F: D _N\times \overline{B _n({\bf 0}, M)}\longrightarrow  D _N$ and $h: D_N \rightarrow \mathbb{R}$ has the echo state property, a result in \cite{RC7} guarantees that the associated filter is automatically causal and time-invariant. This implies the existence of a functional $H _h^{F}: \left(\mathbb{R}^n\right)^{\mathbb{Z}_{-}} \longrightarrow \mathbb{R}$ that, by hypothesis, has the fading memory property. The rest of the statement is a consequence of part {\bf (i)} in Theorem \ref{fmp is inherited}. \quad $\blacksquare$

\subsection{Proof of Theorem~\ref{general universal stochastic}}

We first notice that the polynomial algebra  $\mathcal{A}(\mathcal{R}) $ is, by Theorem \ref{universality deterministic general} and the first part of Theorem \ref{fmp is inherited}, made of fading memory reservoir filters that map into  $L^{\infty}(\Omega, \mathbb{R}) $. 
Using the other hypotheses in the statement we can easily conclude that  the family $\mathcal{A}(\mathcal{R}) $ satisfies the thesis of Theorem~\ref{universality deterministic general} and it is hence universal in the deterministic setup. The result follows from the second part of Theorem~\ref{fmp is inherited}. \quad $\blacksquare$

\medskip

\noindent {\bf Acknowledgments:} We thank Philipp Harms and Herbert Jaeger  for carefully looking at early versions of this work and for making suggestions that have significantly improved some of our results. We thank Josef Teichmann for fruitful discussions. We also thank the editor and two remarkable anonymous referees whose input has significantly improved the presentation and the contents of the paper. The authors acknowledge partial financial support of the French ANR ``BIPHOPROC" project (ANR-14-OHRI-0002-02) as well as the hospitality of the Centre Interfacultaire Bernoulli of the Ecole Polytechnique F\'ed\'erale de Lausanne during the program ``Stochastic Dynamical Models in Mathematical Finance, Econometrics, and Actuarial Sciences" that made possible the collaboration that led to some of the results included in this paper. LG acknowledges partial financial support of the Graduate School of Decision Sciences and the Young Scholar Fund AFF of the Universit\"at Konstanz. JPO acknowledges partial financial support  coming from the Research Commission of the Universit\"at Sankt Gallen and the Swiss National Science Foundation (grant number 200021\_175801/1).

\footnotesize
\printnomenclature[15em]
\normalsize
 
\noindent
\addcontentsline{toc}{section}{Bibliography}
\bibliographystyle{wmaainf}
%\bibliography{/Users/Lyudmila/Mendeley1/GOLibrary}
%\bibliography{/Volumes/Staff/Mila/Library/BibTex/GOLibrary}
\bibliography{/Users/JPO/Dropbox/Public/GOLibrary}
\end{document}